%% file: main.tex
\documentclass[]{fairmeta}
\usepackage{hyperref}
\usepackage{url}
\usepackage{makecell}
\usepackage{paralist, tabularx}
\usepackage{times}
\usepackage{epsfig}
\usepackage{graphicx}
\usepackage{wrapfig}
\usepackage{amsmath}
\usepackage{amssymb}
\usepackage{fontawesome5}
\usepackage[ruled,vlined]{algorithm2e}
\usepackage{algorithmic}
\usepackage{colortbl} 
\usepackage{multirow, multicol, graphicx, xcolor, colortbl, arydshln, tikz}
\usetikzlibrary{tikzmark}
\usepackage{siunitx} 
\usepackage{adjustbox}

\title{\ours{}: {Improving Multimodal Large Language Models with Composite Captions}}

\author[1,2,*]{Xiaohui Chen}
\author[1]{Satya Narayan Shukla}
\author[1]{Mahmoud Azab}
\author[1]{Aashu Singh}
\author[1]{Qifan Wang}
\author[1]{David Yang}
\author[1,3,*]{ShengYun Peng}
\author[1]{Hanchao Yu}
\author[1]{Shen Yan}
\author[1]{Xuewen Zhang}
\author[1]{Baosheng He}

\affiliation[1]{Meta}
\affiliation[2]{Tufts University}
\affiliation[3]{Georgia Tech}

\contribution[*]{Work done during internship at Meta}

\newcommand{\thinparagraph}[1]{\vspace{0.1em}\par\noindent\textbf{#1}\enspace}
\newcommand{\ours}[1]{CompCap}
\abstract{
\input{sections/0_abstract}}

\date{\today}

\definecolor{userbg}{RGB}{245, 245, 245}
\definecolor{userborder}{RGB}{210, 229, 255}
\definecolor{userfont}{RGB}{0, 0, 0}

\definecolor{Gray}{gray}{0.93}
\definecolor{uclagold}{rgb}{1.0, 0.7, 0.0}
\definecolor{airforceblue}{rgb}{0.36, 0.54, 0.66}
\definecolor{rosegold}{rgb}{0.72, 0.43, 0.47}
\definecolor{pastelbrown}{rgb}{0.51, 0.41, 0.33}
\definecolor{isabelline}{rgb}{0.85, 0.89, 0.92}
\definecolor{macaroniandcheese}{rgb}{0.98, 0.89, 0.83}
\definecolor{wildblueyonder}{rgb}{0.85, 0.89, 0.95}
\definecolor{mediumtaupe}{rgb}{0.4, 0.3, 0.28}
\definecolor{bluegray}{rgb}{0.4, 0.6, 0.8}
\definecolor{celestialblue}{rgb}{0.29, 0.59, 0.82}
\definecolor{forestgreen}{rgb}{0.13, 0.54, 0.13}
\definecolor{darkorange}{rgb}{1.0, 0.55, 0.0}
\definecolor{cadmiumred}{rgb}{0.89, 0.0, 0.13}
\definecolor{magnolia}{rgb}{0.97, 0.96, 1.0}
\definecolor{pastelblue}{rgb}{0.68, 0.78, 0.81}
\definecolor{persiangreen}{rgb}{0.0, 0.65, 0.58}
\definecolor{steelblue}{rgb}{0.27, 0.51, 0.71}
\definecolor{bluebell}{rgb}{0.64, 0.64, 0.82}
\definecolor{dimgray}{rgb}{0.41, 0.41, 0.41}
\definecolor{splashedwhite}{rgb}{1.0, 0.99, 1.0}
\definecolor{lavendergray}{rgb}{0.77, 0.76, 0.82}
\definecolor{lightgray}{rgb}{0.83, 0.83, 0.83}
\definecolor{lavendermist}{rgb}{0.9, 0.9, 0.98}
\definecolor{lightgreen}{HTML}{f8fcf4}
\definecolor{lightblue}{rgb}{0.90, 0.95, 0.90}
\definecolor{zeroshot}{rgb}{0.9, 0.9, 0.9}
\definecolor{fourshot}{rgb}{0.8, 0.9, 0.8}
\definecolor{eightshot}{rgb}{0.8, 0.8, 0.9}
\definecolor{sixteenshot}{rgb}{0.9, 0.8, 0.8}
\definecolor{cvprblue}{rgb}{0.21,0.49,0.74}
\begin{document}
\maketitle
\vspace{-1em}
\begin{figure}[t]
\centering
\includegraphics[width=\textwidth]{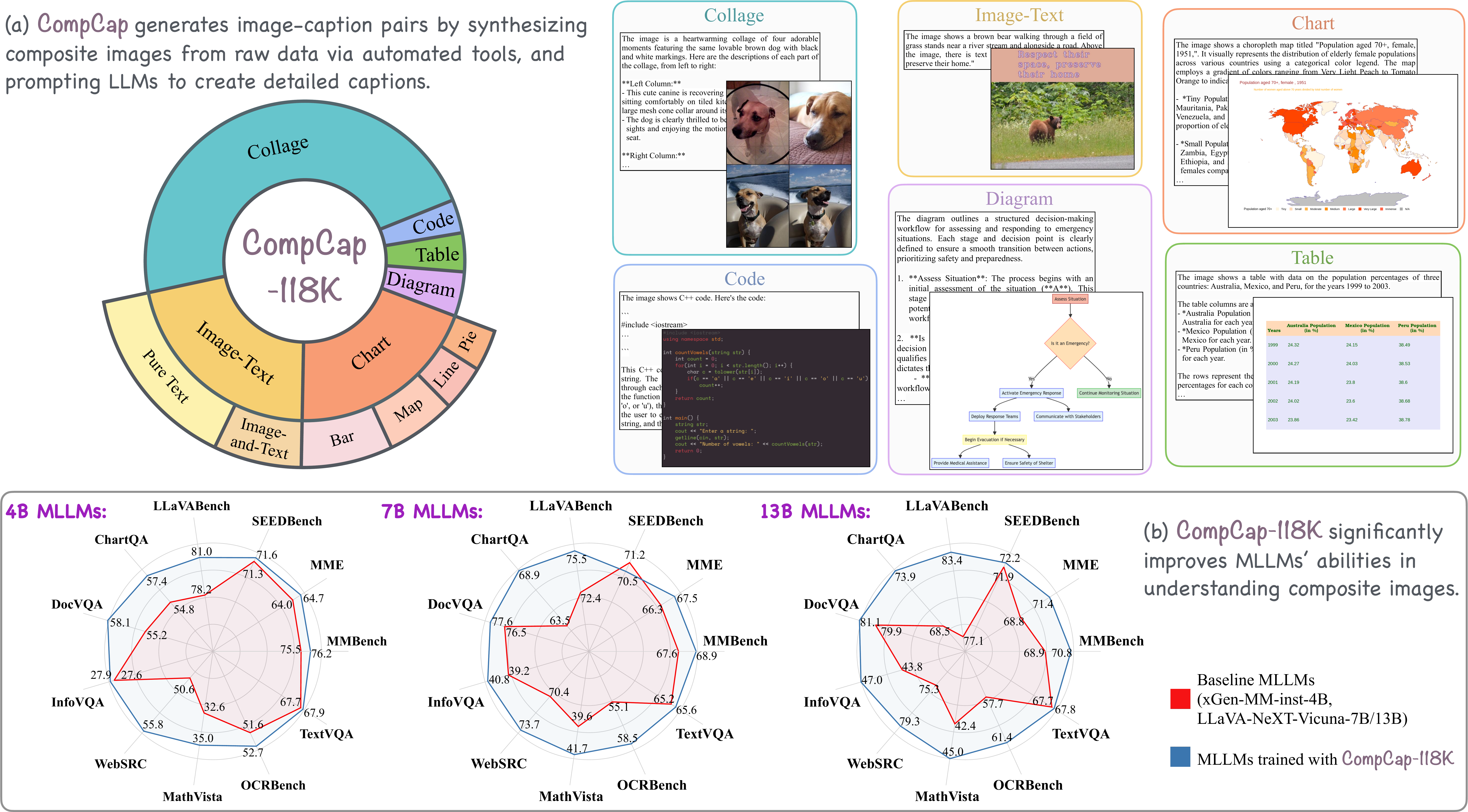}
\caption{\textbf{(a)} \ours{} implements image-caption synthesis pipelines for six composite image types. The composition of the curated \ours{}-118K dataset are 42.3\% Collage, 31.4\% Image-Text, 18.7\% Chart, 3.4\% Table, 2.5\% Diagram, and 1.7\% Code. \textbf{(b)} Introducing \ours{}-118K into the training data significantly improves MLLMs' performance on benchmarks comprising of composite images.} 
\label{fig:teaser}
\end{figure}
\input{sections/1_intro}
\input{sections/2_comp_vs_natu}
\input{sections/3_method}
\input{sections/4_experiments}

\input{sections/5_related_works}
\input{sections/6_discussion}
\bibliographystyle{assets/plainnat}
\bibliography{main}

\clearpage
\newpage
\beginappendix
\input{sections/7_appendix}
\end{document}

%% file: sections/0_abstract.tex
How well can Multimodal Large Language Models (MLLMs) understand composite images?
Composite images (CIs) are synthetic visuals created by merging multiple visual elements, such as charts, posters, or screenshots, rather than being captured directly by a camera. While CIs are prevalent in real-world applications, recent MLLM developments have primarily focused on interpreting natural images (NIs).
Our research reveals that current MLLMs face significant challenges in accurately understanding CIs, often struggling to extract information or perform complex reasoning based on these images. We find that existing training data for CIs are mostly formatted for question-answer tasks (e.g., in datasets like ChartQA and ScienceQA), while high-quality image-caption datasets, critical for robust vision-language alignment, are only available for NIs.
To bridge this gap, we introduce Composite Captions (\ours{}), a flexible framework that leverages Large Language Models (LLMs) and automation tools to synthesize CIs with accurate and detailed captions. 
Using \ours{}, we curate \ours{}-118K, a dataset containing 118K image-caption pairs across six CI types.
We validate the effectiveness of \ours{}-118K by supervised fine-tuning MLLMs of three sizes: xGen-MM-inst.-4B and LLaVA-NeXT-Vicuna-7B/13B.
Empirical results show that \ours{}-118K significantly enhances MLLMs’ understanding of CIs, yielding average gains of 1.7\%, 2.0\%, and 2.9\% across eleven benchmarks, respectively.

%% file: sections/1_intro.tex

\section{Introduction}
Recently, significant advancements have been made in
Multimodal Large Language Models (MLLMs)~\citep{alayrac2022flamingo, Li2023BLIP2BL,mckinzie2024mm1,Liu2023ImprovedBW}.
These models combine images with large language models (LLMs)~\citep{openai2023gpt,team2023gemini,llama3} to harness the powerful capabilities of LLMs, demonstrating exceptional powers in visual and language understanding and achieving remarkable conversational ability. 
However, despite these advances, a notable limitation remains: MLLMs often struggle with comprehensive understanding of composite images (CIs), extracting only partially accurate information. A composite image (CI) is a visual creation that combines various elements, such as photos, graphics, text, or other media, into a single cohesive image. It includes diverse types such as collages, posters, and charts. This raises an important question: {\em Why do these limitations persist?} Our hypothesis is that the observed shortcomings in MLLMs may stem from a lack of CI-caption pairs in the training data.



In essence, the training procedure for MLLMs generally involves two stages: first, pre-training (PT) on image-caption datasets to align the vision encoder with the LLM, and second, supervised fine-tuning (SFT) on instruction or visual question answering (VQA) datasets to enhance the MLLMs' instruction-following abilities~\citep{Li2023BLIP2BL,Liu2023ImprovedBW, mckinzie2024mm1}. Research has shown that using high-quality image captions enhances the alignment between vision and language modalities, thereby improving MLLMs' image understanding~\citep{chen2023sharegpt4v,mckinzie2024mm1}. However, current training data primarily includes high-quality captions for natural images (NIs), while such captions for CIs are often missing.
In this work, we find that MLLMs' captioning abilities are strongly correlated with their VQA performance, suggesting that instruction data is insufficient for MLLMs to fully comprehend CIs.


We introduce \ours{}, a framework that automatically synthesizes high-quality CI-caption pairs, to bridge the data shortage in training MLLMs. 
\ours{} functions as a flexible framework that utilizes various metadata to construct CIs along with their corresponding captions. 
This metadata could include a range of sources such as pre-existing image-caption pairs, layout information, text, or tabular data.  For example, one implementation of \ours{} could be leveraging metadata from individual image-caption pairs and layout specifications to create a collage image with a generated caption. The images can be arranged according to predefined layouts, and LLMs can then generate captions based on the individual image captions and their positional arrangement. Given the diverse nature and unique characteristics of CIs, in this work, we develop six data generation pipelines from \ours{} to comprehensively cover a broad spectrum of CI types. Each pipeline employs distinct types of raw data and various automated tools to facilitate data generation. In total, 
we produce 118K CI-caption pairs, dubbed \ours{}-118K, 
significantly enhancing the diversity and volume of training data available for MLLMs. We highlight the composition of \ours{}-118K in Figure~\ref{fig:teaser}a.

\begin{figure*}[t]
  \includegraphics[width=\textwidth]{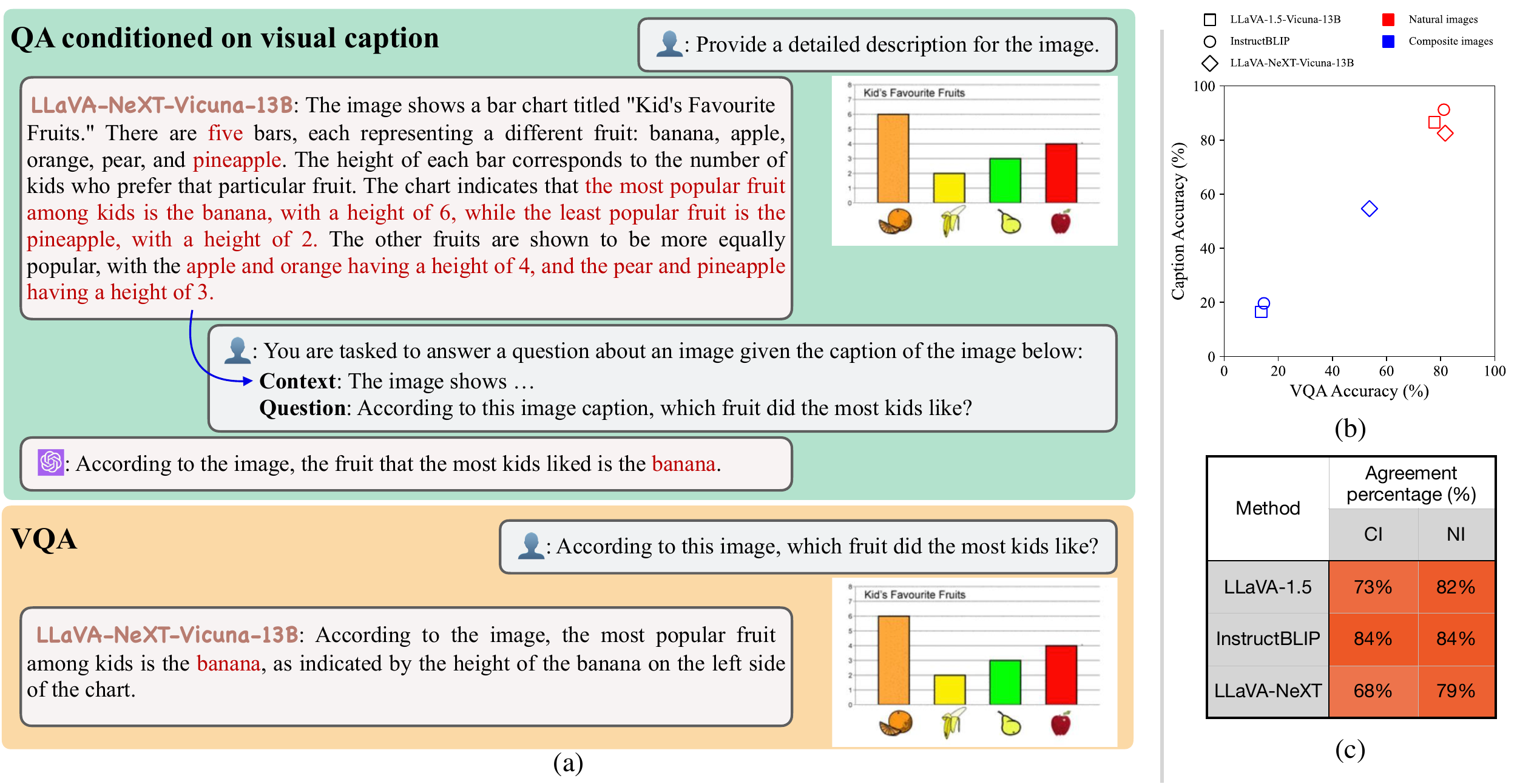}
  \caption{\textbf{MLLMs demonstrate poorer understanding on CIs compared to NIs}. \textbf{(a)} Example of assessing caption accuracy of MLLMs on CI with the help of LLMs. \textbf{(b)} MLLMs generally understand NIs much better than CIs. 
  \textbf{(c)} Errors generated during captioning are consistent with errors generated in VQA,  highlighting the necessity of caption data for better vision-language alignment}
  \label{fig:cap-vqa-demo}
\end{figure*}
We verify the effectiveness of our proposed method by fine-tuning xGen-MM~\citep{xue2024xgen} and LLaVA-NeXT~\citep{liu2024llavanext} on their datasets combined with \ours{}-118K.
Experimental result shows that the integration of \ours{}-118K leads to substantial performance improvements across eleven MLLM benchmarks (see Figure~\ref{fig:teaser}b). To sum up, our contributions are as follows:

\begin{compactitem}
\item\!\! We demonstrate that current MLLMs exhibit limited proficiency in understanding CIs compared to NIs, which could be attributed to the lack of diverse and high-quality CI captions in their training data. 
\item\!\! We introduce \ours{}, a universal framework that generates CIs with detailed and precise captions.
\item\!\! We curate \ours{}-118K, a dataset containing 118K synthetically generated CI-caption pairs.
\item\!\! Our empirical study shows that \ours{}-118K enhances MLLMs' comprehension of CIs, leading to improved performance across various benchmarks.
\end{compactitem}

%% file: sections/2_comp_vs_natu.tex
\section{MLLMs Need Good CI Caption Data} \label{sec: motivation}
In this section, we discuss the necessity of introducing high-quality CI captions for training MLLMs. We starts with giving the definition of composite image in \S~\ref{sec:defn}. \S~\ref{sec:pitfall} illustrates the limitations of MLLMs in accurately understanding CI, which often leads to generating incorrect information during captioning. This observation motivates us to curate a CI-caption dataset.


\subsection{Composite Images}\label{sec:defn}
A CI is a combination of various visual elements including photos, text, and graphics. They are usually designed to deliver rich information for many purposes. In this work, we consider the following six CI categories:

\begin{tcolorbox}[
title=\center{Composite~Images}]
\centering
\begin{tabular}{cccccccccccc}
\faThLarge~~Collage  && \faNewspaper~~Image-Text &&
\faChartBar~~Chart && \faSitemap~~Diagram &&
\faCode~~Code && \faTable~~Table
\end{tabular}
\end{tcolorbox}
While many CIs fall under the first four categories, we specifically include code and table images because they contain structured visual elements. Code images require MLLMs to understand precise syntax, indentation, and programming-specific structures, which differentiates them from other OCR text images. Table images, on the other hand, often contain diverse content types within cells such as images, text, or numerical data, and require MLLMs to accurately extract values, interpret headers, and understand relationships across rows and columns. Additionally, more complex CIs, like posters or infographics, can be created by combining elements from these different CI categories.

%











\subsection{MLLMs Generate Erroneous Caption for CI}\label{sec:pitfall}
MLLMs show inferior performance when it comes to captioning CIs compared to NIs. To illustrate this, we design a demonstrative experiment where MLLMs are tasked with generating detailed captions for both NIs and CIs (See Appendix~\ref{app:exp_demo} for experiment setup). Based on these generated captions, we used an LLM to perform VQA conditioned on the caption and evaluated the performance quantitatively. We also instruct MLLMs to carry out the same VQA by directly referencing the images. Figure~\ref{fig:cap-vqa-demo}a shows an example of how the two pipelines work. We benchmark the caption and VQA accuracy on CIs and NIs using LLaVA-1.5-Vicuna-13B~\citep{Liu2023ImprovedBW}, InstructBLIP~\citep{Dai2023InstructBLIPTG}, and LLaVA-NeXT-Vicuna-13B~\citep{Liu2023ImprovedBW}. We made the following observations:

\thinparagraph{MLLMs exhibit inferior CI understanding:} 
Results from Figure~\ref{fig:cap-vqa-demo}b shows that all three MLLMs report much lower accuracy in CIs compared to NIs, regardless of the task (captioning or VQA). This indicates that MLLMs generally struggle more with understanding CIs.
\thinparagraph{MLLMs make similar errors in captioning and VQA:} 
We also observed that the captioning and VQA accuracy are similar for each image type and model. This raises the question of whether the mistakes made by MLLMs in captions are consistent with those in VQA tasks. To verify this, we counted how often the answers from the two pipelines agreed with each other, regardless right or wrong. As shown in Figure~\ref{fig:cap-vqa-demo}c, the information derived from caption and VQA task tends to be consistent.

The data used to fine-tune MLLMs generally includes high-quality caption data for NIs and instruction data for both NIs and CIs.  However, high-quality caption data for CIs is often missing. Our analysis indicates that MLLMs face challenges in accurately extracting information from CIs when trained only on instruction data, which emphasizes instruction-following over detailed content. The inclusion of high-quality caption data has been shown to improve visual-language alignment of MLLMs~\citep{chen2023sharegpt4v,mckinzie2024mm1}. Additionally, scaling instruction data requires extra human effort to customize diverse and accurate QA pairs for different CI types, making it less economical than generating caption data. This motivates us to propose a curation pipeline that generates high-quality CI-captions.

%% file: sections/3_method.tex
\section{\ours{}} \label{sec: compcap}

This section elaborates on the proposed method. \S~\ref{sec:what is good compcap} explains the characteristics that define a high-quality CI caption. \S~\ref{sec:compcap} illustrates the \ours{} framework. Lastly, \S~\ref{sec:collage-pipe} provides a detailed implementation of \ours{} to generate collage image-caption pairs.

\subsection{What Makes a Good CI Caption}\label{sec:what is good compcap} 
Unlike NIs, CIs contain a diverse array of visual elements that convey complex information. Effective CI captions must be meticulously crafted to enhance MLLMs' ability to accurately interpret CIs. Specifically, we focus on two principles: accuracy and detailedness.

\thinparagraph{Accuracy:} An accurate caption faithfully represents the content of the image without introducing any false or misleading information. This is crucial for ensuring MLLMs generate reliable responses.

\thinparagraph{Detailedness:} A detailed caption provides specific insights into all visual elements and their relationships, going beyond a basic description. This is essential because CIs often have multiple layers of meaning, combining text, graphics, and data. For instance, a comprehensive caption for an infographic should not only describe the overall topic but also explain each section, data point, and visual cue. Including such details helps MLLMs align textual descriptions with all aspects of the visual content, thereby improving the model's ability to fully understand and interpret CIs.

\subsection{The \ours{} Framework}\label{sec:compcap}
\begin{wrapfigure}{r}{0.44\columnwidth} 
\vspace{-2em}
\centering
\includegraphics[width=\linewidth]{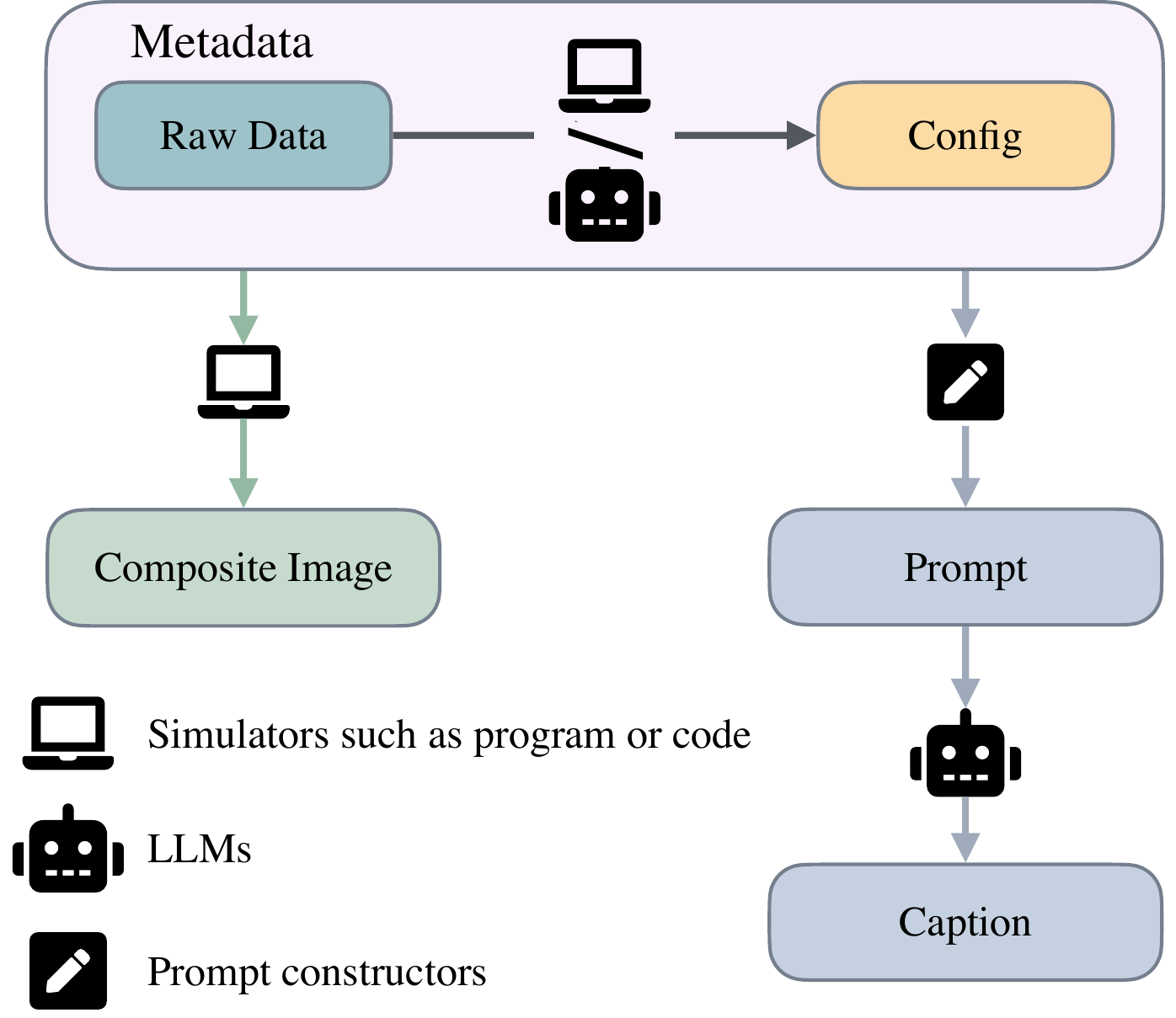} 
\caption{\textbf{The \ours{} Framework.} The synthesis pipeline for different CI types implements \ours{} differently. }
\label{fig:compcap-demo}
\end{wrapfigure}
\ours{} is a general framework that synthesizes CI-caption pairs for training MLLMs. Figure~\ref{fig:compcap-demo} shows an illustration of \ours{}. It leverages metadata to generate both a CI and its corresponding caption. Below, we explain what constitutes the metadata and how it is utilized in the creation of the composite image and caption generation.


\begin{figure}[t]
    \centering
  \includegraphics[width=\textwidth]{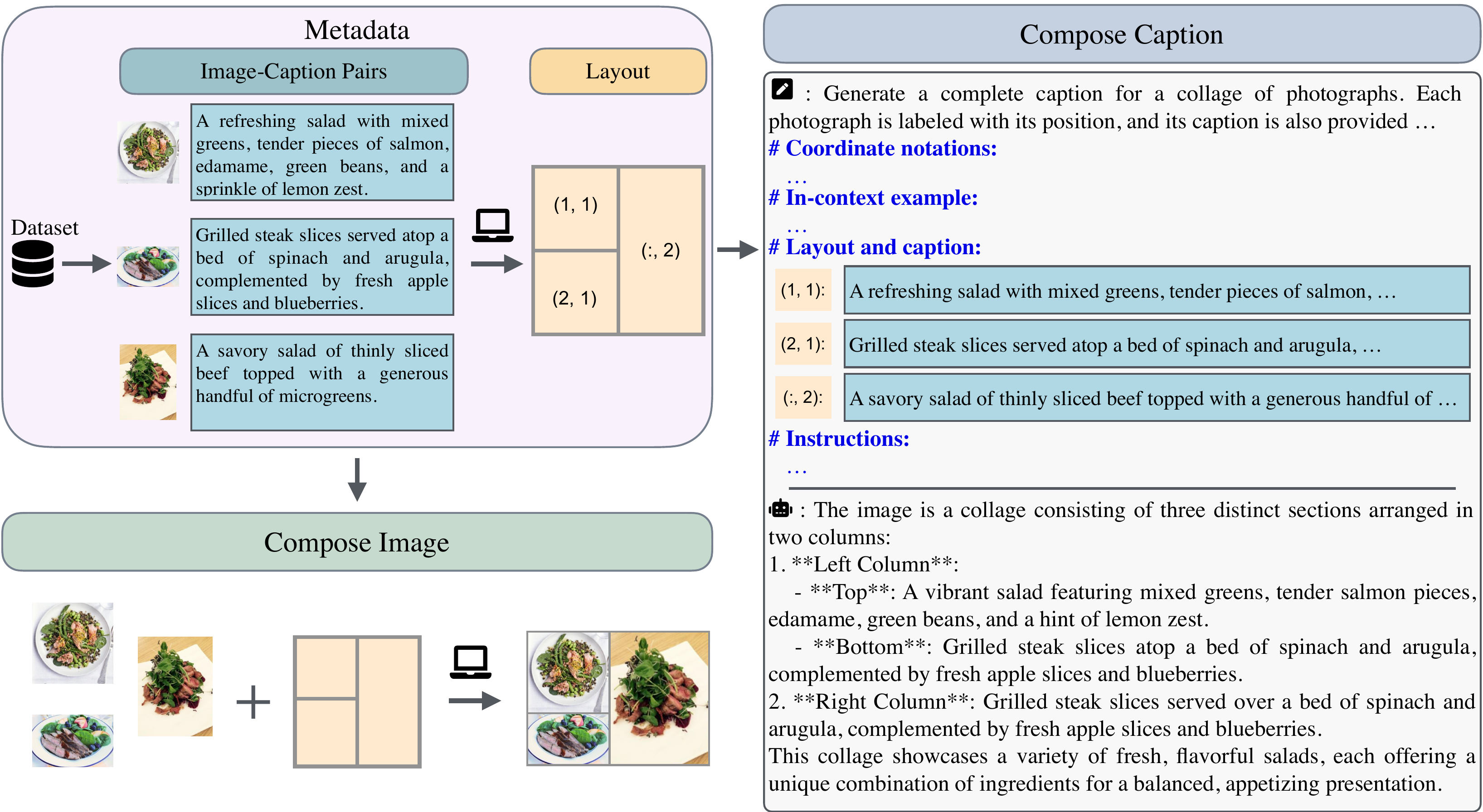}\vspace{-0.5em}
  \caption{\textbf{The Collage implementation.} We sample raw data from image-caption datasets and randomly generate a layout for the selected images. The images are then arranged into a collage following this layout, while an LLM generates a caption for the collage given both the layout details and captions of the individual images.}
  \label{fig:collage-demo}
  \vspace{-1em}
\end{figure}
\thinparagraph{Metadata:} Metadata comprises both raw data and configuration details necessary for generating a CI. The nature of the raw data varies depending on the CI class and could include a collection of image-caption pairs, tabular data, or textual data such as code or math expressions. Raw data serves as the fundamental content that the CI will represent. Configuration, or customization, on the other hand, is generated through a random process based on the raw data. It determines how this raw data is visually represented in the CI. For example, when generating a chart, the raw data might contain tabular information, while the configuration specifies details like the chart type (such as bar or line charts), the data columns to visualize, titles, and other visual parameters such as color and style.

\thinparagraph{Creation of Composite Image:} The image pipeline reads the configuration and composes the raw data accordingly to produces the CI. For each CI class, the pipeline is implemented using rendering tools~\citep{mermaid, carbon_now} or Python libraries~\citep{plotly,opencv_library,clark2015pillow,matplotlib}.  
For instance, in the case of chart visualization, the configuration might provide visualization parameters like color, legend, and font size. These parameters are then applied using Matplotlib~\citep{matplotlib} or Plotly~\citep{plotly} to create the final chart image.

\thinparagraph{Caption Generation:} We use LLMs to produce caption for the CIs. The key challenge here is designing prompts that effectively guide LLMs to generate accurate and detailed caption. A base prompt is crafted for each CI type to provide specific instructions on which aspects of the image the LLMs should focus on when generating the caption. For instance, captions for collage should recognize the content of subimages, identify their layout, and understand the possible interplay among them. Captions for chart should extract and analyze the data, with less emphasis on aesthetics.
To enhance the quality and diversity of the generated captions, we also include in-context examples when constructing prompts for some of the CI types.
\vspace{-0.1em}
\subsection{An Instantiation of \ours{} for \faThLarge~Collage}\label{sec:collage-pipe}
\vspace{-0.1em}
This section outlines the \ours{} framework applied to create a collage. We highlight the implemented pipeline in Figure~\ref{fig:collage-demo}. Detailed implementations for all CI pipelines are provided in the Appendix (Collage:~\ref{app:collage}; Image-Text:~\ref{app:image-text}; Chart:~\ref{app:chart}; Diagram:~\ref{app:diagram}; Code:~\ref{app:code}; Table:~\ref{app:table}).

\thinparagraph{Raw data:}
The pipeline begins with retrieving a set of image-caption pairs from the database.
To simulate diverse, real-world scenarios, we employ three retrieval methods:
\begin{compactenum}
\item \textit{Random retrieval:} Sampling image-caption pairs uniformly from the datasets to create unrelated image. 
\item \textit{Similarity-based retrieval:} Sampling image-caption pairs with similar visual and textual features. We calculate the similarity between any two image-caption pairs by summing the cosine similarity of their image embeddings and that of their caption embeddings. We extract the image embeddings using Dino-v2~\citep{radford2021learning} and caption embedding using CLIP~\citep{radford2021learning}.


\item \textit{Entity-based retrieval:} Retrieving image-caption pairs that depict the same entity (e.g., a public figure or landmark). Beginning with entity randomly sampled from a predefined entity list, we filter pairs to include those whose captions contain the chosen keyword, and sample from the filtered group.
\end{compactenum}
Images retrieved by similarity-based or entity-based methods are related, generating collage that resembles real-world data. On the other hand, random retrieval composes unrelated images to form a collage. Such data are counterfactual and can help in model debiasing~\citep{yu2024hallucidoctor}, thus mitigating hallucination caused by LLMs' parametric knowledge~\citep{bai2024hallucination}.

\thinparagraph{Layout:} We consider layouts where at least one dimension (row or column) is aligned. Since the layout depends on the dimensions (width and height) of the sampled images, we implement two types of layout, each organically combined with image sampling:
\begin{compactenum}
    \item \textit{Grid layout:} We first generate grid layout that specifies the width/height ratios for each image, then samples images that meet these constraints.
    \item \textit{Auto layout:} We first sample images, and find a layout that seamlessly composes them into a collage. 
\end{compactenum}
We discuss more details and shows some examples of the two employed layouts in Appendix~\ref{app:collage-data-layout}.

\thinparagraph{Composing collage:} We use image processing tools (OpenCV~\citep{opencv_library} and PIL~\citep{clark2015pillow}) to assemble the retrieved images into a collage based on the sampled layout. To enhance the diversity of the generated collage, we introduce three types of randomization: (1) the margin between images within the collage, (2) the padding around the border of the collage, and (3) the collage background using predefined background patterns.

\thinparagraph{Prompt and caption design:} To generate consistent and accurate captions from LLMs, the prompt is structured to include the following components:

\begin{compactenum}
\item \textit{Coordinate system:} A coordinate system that clarifies the spatial arrangement of images.

\item \textit{Layout and caption data:} Metadata that describes the generated collage in bullet point style, where each bullet point contains the image location represented using the coordinate system and the caption of the image.

\item \textit{In-context example:} An input-output example that helps LLMs understand the expected format and style for captions. The designed caption describes the collage by listing caption for each image in a bullet-point format. When images are related, an inference highlighting their interplay is added. We use active in-context example selection to improve the accuracy and diversity of captions~\citep{zhang2022active}.

\end{compactenum}
We further post-process the generated collage-caption pairs to improve the data quality, which includes: (1) filter out collages that contain duplicate images, and (2) reformat and filter captions from LLMs' responses to ensure clarity and relevance.

%% file: sections/4_experiments.tex
\section{Training MLLMs with \ours{} Data}

We train MLLMs with CI-caption dataset to validate the effectiveness of the \ours{} framework. We first describe the curated dataset in  \S~\ref{sec:dataset}, then the training details in \S~\ref{sec:train_mllm}.
\input{tables/dataset}

\subsection{The \ours{}-118K Dataset}\label{sec:dataset}
The \ours{}-118K dataset, generated via \ours{}, is a synthetic collection of 117,879 image-caption pairs spanning six CI categories. Each category uses different types of metadata and simulation tools to generate the images, and the captions are created using various LLMs, depending on the complexity of the captioning task. 
Table~\ref{tab:data_stats} provides a summary of the dataset's statistics.

\subsection{The \ours{}-4B/7B/13B MLLMs}\label{sec:train_mllm}
We develop the \ours{} series MLLMs using two recently introduced MLLM architectures: LLaVA-NeXT~\citep{liu2024llavanext} and xGen-MM~\citep{xue2024xgen}. For LLaVA-NeXT, we use the 2024-01 release (7B and 13B Vicuna versions), while for xGen-MM, we use version 1.5 (4B model). 

The MLLMs are trained in two stages: a PT stage and a SFT stage. We incorporate \ours{}-118K dataset into the SFT stage. To ensure a fair comparison, we uniformly downsample the original SFT dataset and add \ours{}-118K such that the total number of training samples remained equivalent. 
Since the SFT dataset for xGen-MM is not released, we curate a SFT dataset comprising 782K image-text pairs and 221K pure text samples, closely following the data recipe reported in \citet{xue2024xgen}. We refer to the resulting MLLMs as \textbf{\ours{}-4B/7B/13B}. We validate the effectiveness of the proposed framework through the experiments outlined in the following section. 


\section{Experiments} 

\subsection{Evaluation Benchmarks}
We evaluate the MLLMs across multiple benchmarks, particularly with the focus on their ability to comprehend CIs. We adopt NI-focused benchmarks like SEEDBench~\citep{li2024seed}, TextVQA~\citep{textvqa}, MMBench~\citep{liu2025mmbench}, MME~\citep{yin2023survey}, and LLaVABench~\citep{Liu2023VisualIT} to test conversational, reasoning, perception, and text recognition abilities.

We also use CI-focused benchmarks, including ChartQA~\citep{chartqa}, DocVQA~\citep{docvqa}, InfoVQA~\citep{infovqa}, WebSRC~\citep{chen2021websrc}, MathVista~\citep{lu2023mathvista}, and OCRBench~\citep{liu2023hidden}. Specifically, ChartQA, DocVQA, and InfoVQA measure the ability to interpret visually rich chart, document, or diagram images, while WebSRC focuses on web-based reading comprehension. MathVista and OCRBench contain both NIs and CIs, testing OCR abilities across various formats. 
For evaluation, we use VLMEvalKit~\citep{duan2024vlmevalkit} and LMMs-EVAL~\citep{zhang2024lmms}.

\input{tables/result_sidebyside}

\subsection{Main Results}\label{sec:result}
We present the quantitative results in Table~\ref{tab:main_result_s2s}, comparing MLLMs of three different sizes against similarly scaled models (3B-4B, 7B-8B, and 13B).

From Table~\ref{tab:main_result_s2s}a, we can see that \ours{}-4B/7B/13B consistently outperform the other MLLMs (xGen-MM-inst.-4B$^*$ and LLaVA-NeXT-Vicuna-7B/13B) that share the same architectures and similar size of training data  by 1.7\%, 2.0\%, and 2.9\%, respectively. 
The performance gains are particularly noticeable on 

\begin{minipage}{0.49\textwidth}
\begin{center}
\centering
\captionsetup{type=table}

\input{tables/ablation}
\end{center}
\end{minipage}
\hspace{0.5em}
\begin{minipage}{0.49\textwidth}
\begin{center}
\centering
\captionsetup{type=figure}
\includegraphics[width=\linewidth]{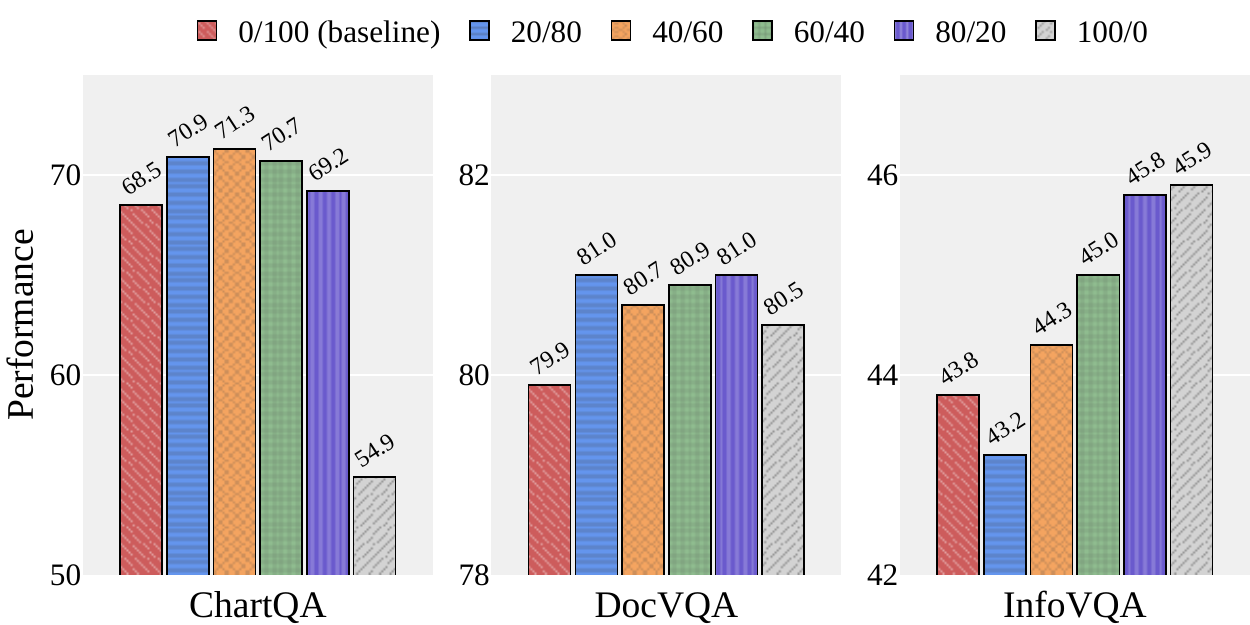}
\caption{\textbf{\#Captions/\#Instructions Ablation on ChartQA.} We replace $\text{k:~k}\in\{\text{0,20,40,60,80,100}\}$ percents of the instructions in ChartQA training set with captions and train the LLaVA-NeXT-Vicuna-13B. We evaluate on ChartQA, DocVQA and InfoVQA benchmarks.} 
\label{fig:chartqa_ablation}
\end{center}
\end{minipage}
\vspace{0.5em}

benchmarks containing CIs. For benchmarks containing NIs, \ours{}-4B/7B/13B also achieve competitive performance.
Interestingly, even though the \ours{}-118K dataset does not include math-specific data, our models still show substantial improvements on the MathVista benchmark. This is likely due to the significant presence of CIs in mathematical tasks, where accurate CI understanding is crucial for mathematical reasoning.
Additionally, we report comparison with SoTA MLLMs in Table~\ref{tab:main_result_s2s}b. Although there remains a performance gap between the \ours{} series and SoTA MLLMs within each size category, it’s important to note that \ours{} series are trained on significantly smaller datasets (10x-100x fewer samples). However, when compared to MLLMs trained on datasets of similar scale (less than 10x difference), the \ours{} series demonstrate competitive performance across the board. 

\begin{figure}[t]
    \centering
    \includegraphics[width=\linewidth]{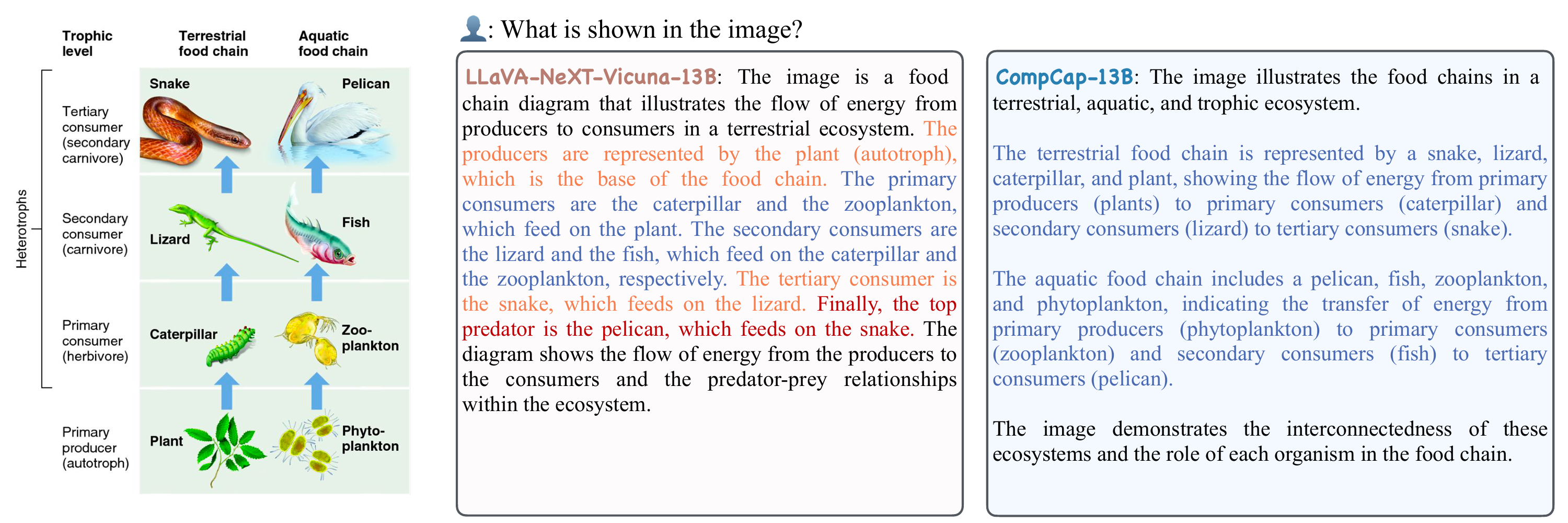}
    \caption{\textbf{Example of MLLMs on CI captioning:} 
    Informative content is highlighted in red if incorrect, in orange if correct but incomplete, and in blue if correct and complete.
    }
    \label{fig:qualitative-1}
    \vspace{-1em}
\end{figure}

We also show a qualitative example in Figure~\ref{fig:qualitative-1}. Compared to LLaVA-NeXT-Vicuna-13B, \ours{}-13B generates a more informative caption that comprehensively summarizes the statistics presented in the image. This showcases how a better visual-language alignment is achieved by introducing caption data for CIs. 

These results indicate that high-quality captions significantly enhance MLLMs' understanding of CIs. Our \ours{} framework and its curated dataset address a critical gap in the dataset blueprint for training MLLMs.

\subsection{Ablations}\label{sec:ablation}
\thinparagraph{CI component ablation:} We incrementally include the curated image-caption pairs from each CI type into the \ours{} dataset to investigate the effectiveness of each type. We train LLaVA-NeXT-Vicuna-13B on the dataset variants and assess how the inclusions affect MLLMs' performance on NI-dominated benchmarks and CI-dominated benchmarks. Results are summarized in Table~\ref{tab:ablation}. With the introduction of each CI-caption component, the MLLM consistently achieves better performance, suggesting the effectiveness of each CI caption design.

\thinparagraph{Caption-Instruction ablation on ChartQA:} 
We design a controlled experiment utilizing the ChartQA dataset
to illustrate how high-quality caption data can enhance performance more effectively than instruction data. We first use the advanced MLLM to generate detailed caption for each chart image in ChartQA training set, resulting in 18,317 image-caption pairs. Then we randomly select k\% of the image-instruction pairs from ChartQA training set, and replace the instruction text with the generated captions, where k$\in$\{0,~20,~40,~60,~80,~100\}. Note that we only modify the ChartQA training set, and all other components in LLaVA-NeXT SFT dataset remain unchanged. Such setting ensures MLLMs are trained on the same amount of data. 

Figure~\ref{fig:chartqa_ablation} reports the ChartQA/DocVQA/InfoVQA performance of MLLMs trained with varing Caption-Instruction ratio.
First, results on the ChartQA test set show that caption data is more effective in boosting performance. 
Second, MLLMs' instruction-following abilities on ChartQA dramatically decrease when only caption data are used, indicating the necessity of including instruction data for training. 
Third, caption data significantly improve MLLMs' performance in other domains, such as DocVQA and InfoVQA, suggesting that knowledge gained from captions is more transferable. 
These findings emphasize the importance of incorporating high-quality caption data for CI, further supporting the motivation for our research.

%% file: tables/dataset.tex
\begin{table}[t]
    \centering
        \resizebox{\linewidth}{!}{

    \begin{tabular}{lccccc}\toprule
Category & Metadata & Image Simulator(s) & Caption Composition & \#Samples & Avg. Char.\\\midrule
\faThLarge~Collage & Image-Caption \& Layout  & OpenCV~\citep{opencv_library} / PIL~\citep{clark2015pillow}& ~~LGC$^*$ & 50K & 913\\
\faNewspaper~Image-Text & Image-Caption \& Text \& Layout& OpenCV / PIL  / Augraphy~\citep{augraphy_paper} & Text + LGC / Text & 37K & 221\\
\faChartBar~Chart & (Geo) Tabular data & Plotly~\citep{plotly} & LGC & 22K & 1468\\
\faSitemap~Diagram & Mermaid diagram code & Mermaid~\citep{mermaid} \& Selenium & LGC & 3K &2044\\
\faCode~Code & Code snippet & Carbon~\citep{carbon_now} \& Selenium  & Code snippet + LGC & 2K& 1106\\
\faTable~Table & Tabular data & Matplotlib~\citep{matplotlib}  & Markdown table + LGC & 4K &928\\
        \bottomrule
    \end{tabular}}
    \caption{\textbf{Statistics of \ours{}-118K dataset.} LGC$^*$ stands for \underline{L}LM-\underline{G}enerated \underline{C}ontent, which encodes the detail breakdown and analysis of the information carried by a CI.}
    \label{tab:data_stats}
\end{table}

%% file: tables/result_sidebyside.tex
\begin{table}[t]
    \centering
    \begin{minipage}{0.69\textwidth}
  \resizebox{\textwidth}{!}{
\begin{tabular}{lccccccccccccccc}
    \toprule
Model &
    \rotatebox{90}{\small{SEEDBench} 
    {$\mathbin{\Diamond}$}} &
    \rotatebox{90}{\small{TextVQA} 
    {$\mathbin{\Diamond}$}} &
    \rotatebox{90}{\small{MMBench} $\mathbin{\Diamond}$\tiny{$\mathbin{\blacklozenge}$}} &
    \rotatebox{90}{\small{MME (norm)} $\mathbin{\Diamond}$\tiny{$\mathbin{\blacklozenge}$}} &
    \rotatebox{90}{\small{LLaVABench} $\mathbin{\Diamond}$\tiny{$\mathbin{\blacklozenge}$}} &
    \rotatebox{90}{\small{MathVista}
    $\mathbin{\blacklozenge}$\tiny{$\mathbin{\Diamond}$}} &
    \rotatebox{90}{\small{OCRBench}
    $\mathbin{\blacklozenge}$\tiny{$\mathbin{\Diamond}$}} &
    \rotatebox{90}{\small{ChartQA} 
    {$\mathbin{\blacklozenge}$}} &
    \rotatebox{90}{\small{DocVQA} 
    {$\mathbin{\blacklozenge}$}} &
    \rotatebox{90}{\small{InfoVQA} 
    {$\mathbin{\blacklozenge}$}} &
    \rotatebox{90}{\small{WebSRC} 
    {$\mathbin{\blacklozenge}$}} &
    \rotatebox{90}{\small{Avg.}} \\
\midrule
\rowcolor{lightgray}
\multicolumn{13}{l}{\emph{4B MLLMs}}\\
\midrule
xGen-MM-inst.-4B$^*$~\citep{xue2024xgen} 
& 71.3 & 67.7 & 75.5 & 64.0 & 78.2 & 32.6 & 51.6 & 54.8 & 55.2 & 27.6  & 50.6 & 57.2 \\
CompCap-4B 
& \textbf{71.6} & \textbf{67.9} & \textbf{76.2} & \textbf{64.7 }& \textbf{81.0} & \textbf{35.0} & \textbf{52.7} & \textbf{57.4} & \textbf{58.1} & \textbf{27.9} & \textbf{55.8} & \textbf{58.9}\\

\midrule
\rowcolor{lightgray}
\multicolumn{13}{l}{\emph{7B MLLMs}}\\
\midrule
LLaVA-NeXT-Vicuna-7B~\citep{liu2024llavanext}  
& \textbf{71.2} & 65.2 & 67.6  & 66.3 & 72.4 & 39.6  & 55.1 & 63.5  & 76.5  & 39.2  & 70.4 & 62.5 \\
CompCap-7B 
& 70.5 & \textbf{65.6} & \textbf{68.9} & \textbf{67.5} & \textbf{75.5}  & \textbf{41.7} & \textbf{58.5} & \textbf{68.9} & \textbf{77.6} & \textbf{40.8} & \textbf{73.7} & \textbf{64.5}\\
\midrule
\rowcolor{lightgray}
\multicolumn{13}{l}{\emph{13B MLLMs}}\\
\midrule
LLaVA-NeXT-Vicuna-13B~\citep{liu2024llavanext} 
& 71.9 & 67.6 & 68.9  & 68.8 & 77.1 & 42.4 & 57.7 & 68.5  & 79.9 & 43.8  & 75.3 & 65.6 \\
CompCap-13B 
&  \textbf{72.2} &   \textbf{67.8} & \textbf{70.8} & \textbf{71.4} & \textbf{83.4} & \textbf{45.0} & \textbf{61.4} & \textbf{73.9} & \textbf{81.1}    & \textbf{47.0} & \textbf{79.3} & \textbf{68.5}
\\
\bottomrule
\end{tabular}}
\center{\small{(a) Comparison with MLLMs with same architectures and same amount of training data.}}
\end{minipage}
\hfill
\begin{minipage}{0.301\textwidth}

\resizebox{\linewidth}{!}{
\begin{tabular}{lcc}
    \toprule
Model &
    PT/SFT \#Data& Avg. \\
\midrule
\rowcolor{lightgray}
\multicolumn{3}{l}{\emph{3B - 4B MLLMs}}\\
\midrule
Phi-3-vision~\citep{abdin2024phi} & 5B/$>$8.3M & \textbf{66.9}\\
xGen-MM-inst.-4B~\citep{xue2024xgen} & $>$25M/UNK. & \underline{60.2}\\
\rowcolor{isabelline}
xGen-MM-inst.-4B$^*$~\citep{xue2024xgen} & \cellcolor{lightblue}$>$25M/1M &  57.2\\
\rowcolor{isabelline}
CompCap-4B & \cellcolor{lightblue}$>$25M/1M  & {58.9}\\

\midrule
\rowcolor{lightgray}
\multicolumn{3}{l}{\emph{7B - 8B MLLMs}}\\
\midrule
ShareGPT4V-7B~\citep{chen2023sharegpt4v} & 1.2M/665K & 43.8\\
Qwen-VL-chat-7B~\citep{wang2024qwen2} & UNK./UNK. & 54.5\\
Cambrian-8B~\cite{tong2024cambrian} & 1.2M/7M & \textbf{65.9}\\
\rowcolor{isabelline}
LLaVA-NeXT-Vicuna-7B~\cite{liu2024llavanext} & \cellcolor{lightblue}558K/779K & 62.5\\
\rowcolor{isabelline}
CompCap-7B & \cellcolor{lightblue}558K/779K & \underline{64.5}\\
\midrule
\rowcolor{lightgray}
\multicolumn{3}{l}{\emph{13B MLLMs}}\\
\midrule
ShareGPT4V-13B~\citep{chen2023sharegpt4v} & 1.2M/665K  & 44.8\\
OmChat-v2.0-13B~\citep{zhao2024omchat} & $>$6.5B/20M & \textbf{75.0}\\
Cambrian-13B~\citep{tong2024cambrian} & 1.2M/7M & 67.2\\
\rowcolor{isabelline}
LLaVA-NeXT-Vicuna-13B~\citep{liu2024llavanext} & \cellcolor{lightblue}558K/779K & 65.6\\
\rowcolor{isabelline}
CompCap-13B & \cellcolor{lightblue}558K/779K & \underline{68.5}
\\
\bottomrule
\end{tabular}
}
\center{\small{(b) Comparison with SoTA MLLMs.}}
\end{minipage}
\caption{\textbf{Evaluation on MLLM benchmarks.} \textbf{(a)} $\mathbin{\Diamond}$ indicates benchmarks with NIs, $\mathbin{\blacklozenge}$ indicates CIs, and combined symbols represent benchmarks containing both, with symbol size reflecting the dominant type. For each model size, the two rows share the same MLLM architecture and PT dataset and are only different in SFT data mixture (We retrain the SFT stage for xGen-MM-inst.-4B$^*$ for a fair comparison). Better performance are marked in bold in each model size. \textbf{(b)} We report the number of samples used for training (PT+SFT) to demonstrate SoTA MLLMs. The greater sign ``$>$'' indicates a lower bound of the number of samples, which is obtained by dividing the number of used tokens by the context window size 4K. We bold the best performance and underline the second best in each model size.} 
\label{tab:main_result_s2s}
\end{table}

%% file: tables/ablation.tex
    \centering

\resizebox{\linewidth}{!}{
\begin{tabular}{lccc}
\toprule
Component ~~~~&~~~~ $\mathbin{\Diamond}$\scriptsize{($\mathbin{\blacklozenge}$)} ~~~~~~~&~~~~ $\mathbin{\blacklozenge}$\scriptsize{($\mathbin{\Diamond}$)} ~~~~~~~&~~~~ Avg.~~~~\\
\midrule
Baseline$^*$ & 70.9 & 61.3 & 65.6$_\text{\textbf{\textcolor{forestgreen}{~~~~~~~~~~}}}$\\
+ \faThLarge~Collage & 71.5 & 62.4 & 66.4$_\text{\textbf{\textcolor{forestgreen}{(+0.8)}}}$\\
+ \faCode~Code & 71.3 & 62.8 & 66.6$_\text{\textbf{\textcolor{forestgreen}{(+1.0)}}}$\\
+ \faTable~Table & 71.7 & 63.0 & 67.0$_\text{\textbf{\textcolor{forestgreen}{(+1.4)}}}$\\
+ \faSitemap~Diagram & 71.5 & 63.1 & 67.4$_\text{\textbf{\textcolor{forestgreen}{(+1.8)}}}$\\
+ \faChartBar~Chart & 72.2 & 63.9 & 68.0$_\text{\textbf{\textcolor{forestgreen}{(+2.4)}}}$\\
+ \faNewspaper~Image-Text & \multirow{2}{*}{73.1} & \multirow{2}{*}{64.6} & \multirow{2}{*}{\textbf{68.5}$_\text{\textbf{\textcolor{forestgreen}{(+2.9)}}}$}\\
(\ours{}-118K)\\
\bottomrule
\end{tabular}}

\caption{\textbf{Ablation study of each CI category} on LLaVA-NeXT-Vicuna-13B. We report the average scores over NI-dominated benchmarks {$\mathbin{\Diamond}$\scriptsize{($\mathbin{\blacklozenge}$)}} (SEEDBench, TextVQA, MMBench, MME, LLaVABench), CI-dominated benchmarks {$\mathbin{\blacklozenge}$\scriptsize{($\mathbin{\Diamond}$)}} (MathVista, OCRBench, ChartQA, DocVQA, InfoVQA, WebSRC), and all benchmarks. Baseline$^*$ stands for the original SFT data recipe in LLaVA-NeXT.}
\vspace{-0.5em}

\label{tab:ablation}

%% file: sections/5_related_works.tex
\section{Related Works} 
\textbf{MLLMs}~\citep{openai20234v, liu2024llavanext, Dai2023InstructBLIPTG, mckinzie2024mm1} are designed to enhance LLMs~\citep{openai2023gpt,touvron2023llama, llama3, yang2024qwen2,bai2023qwen} with multimodal understanding, particularly for visual information. These models typically connect a pre-trained vision encoder~\citep{radford2021learning, zhai2023sigmoid} to a powerful LLM, using a vision-language connector like MLPs~\citep{Liu2023VisualIT} or Q-former~\citep{Dai2023InstructBLIPTG} to align the visual with text modalities. Recent advancement on MLLMs majorly focus on leveraging and curating extensive, diverse, and high-quality training datasets~\citep{chen2023sharegpt4v, xue2024xgen, tong2024cambrian, mckinzie2024mm1} to improve the MLLMs' abilities. Particularly, \citet{chen2023sharegpt4v, mckinzie2024mm1} highlight the importance of high-quality caption data. While image-caption pairs being fundamental for aligning visual and textual representations, there is a lack of such dataset for CIs, which is an important gap our work aims to address.

\noindent\textbf{Multimodal synthetic datasets}~\citep{johnson2017clevr,kafle2018dvqa,methani2020plotqa,kim2022ocr,li2022blip,chang2022mapqa,lindstrom2022clevr,Liu2023VisualIT} have emerged as a scalable and cost-effective solution for training MLLMs. These datasets are either generated by producing synthetic captions or instructions for real images using AI tools or by creating synthetic images paired with template-based instructional text. For instance, LLaVA~\citep{Liu2023VisualIT} prompts GPT to generate instruction data for COCO images~\citep{lin2014microsoft}, while BLIP~\citep{li2022blip} employs CapFilt to generate more refined caption data. For complete image-text pair synthesis pipelines, DVQA~\citep{kafle2018dvqa} and PlotQA~\citep{methani2020plotqa} focus on synthetically generated charts, aiming to develop question-answering pairs that test the ability to interpret, retrieve data from, and reason about the information presented in these charts. Similarly, MapQA~\citep{chang2022mapqa} emphasizes choropleth maps of the United States, where color variations depict data values across geographic regions, offering a range of map styles and question types to evaluate map interpretation and information extraction. Donut~\citep{kim2022ocr} presents SynthDoG, which generates synthetic document image from given text, targeting the ability of document understanding. In contrast to previous methods that generate synthetic images, our proposed \ours{} framework covers a wider spectrum of image types (Collage, Image-Text, Chart, Diagram, Table, and Code). Moreover, \ours{} targets at curating detailed caption for the generated images, which benefits more on the vision-language alignment rather than instruction-following ability.

%% file: sections/6_discussion.tex
\section{Conclusion}
\vspace{-0.5em}
In this work, we propose \ours{}, a versatile framework designed to generate high-quality, detailed captions for composite images (CIs) such as charts, diagrams, and tables. The resulting dataset, \ours{}-118K, comprises 118K captions across six CI categories, significantly enhancing MLLMs' capabilities in CI understanding. Experimental results demonstrate that incorporating CompCap-118K notably improves MLLMs' performance across eleven benchmarks, particularly in CI-specific tasks, emphasizing the critical role of caption data in achieving robust vision-language alignment. Additionally, by expanding \ours{} with more CI pipeline implementation and raw data sources, we can further scale and enhance the generated dataset. 

%% file: sections/7_appendix.tex
\appendix

\section{CI Implementations of \ours{} }\label{app:all-pipe}
This section elaborates the implementation of each CI type. We show the detailed pipeline implementation of collage in \ref{app:collage}; image-text in \ref{app:image-text}; chart in \ref{app:chart}; diagram  in \ref{app:diagram}; Code in \ref{app:code}; and table in \ref{app:table}.
\subsection{\faThLarge~Collage Implementation}\label{app:collage}
We summarize the workflow in Figure~\ref{fig:collage-demo}. In this section, we first show how the image-captions used for composing collage are retrieved, then we elaborate on the design of collage caption and LLM prompting.

\subsubsection{Data Sources, Layout, and Retrieval Engines}~\label{app:collage-data-layout} 
We retrieve image-caption pairs from existing image-caption datasets. And we maintain a curated entity list of public figures, artworks, landmarks, and brands sourced from web data.

\thinparagraph{Pre-processing:} We first process the datasets for better retrieval quality and efficiency:
\begin{compactenum}
\item \textit{Construct entity-sample lookup table.} For each image, we identify entities in the original caption that match entries in our maintained entity list and create an entity-image lookup table specifically for entity-based retrieval.
\item \textit{Pre-compute embeddings.} For each sample in both datasets, we pre-compute Dino-v2 image embeddings and CLIP caption embeddings.
\end{compactenum}

\begin{wrapfigure}{r}{0.46\columnwidth}
\centering
\includegraphics[width=\linewidth]{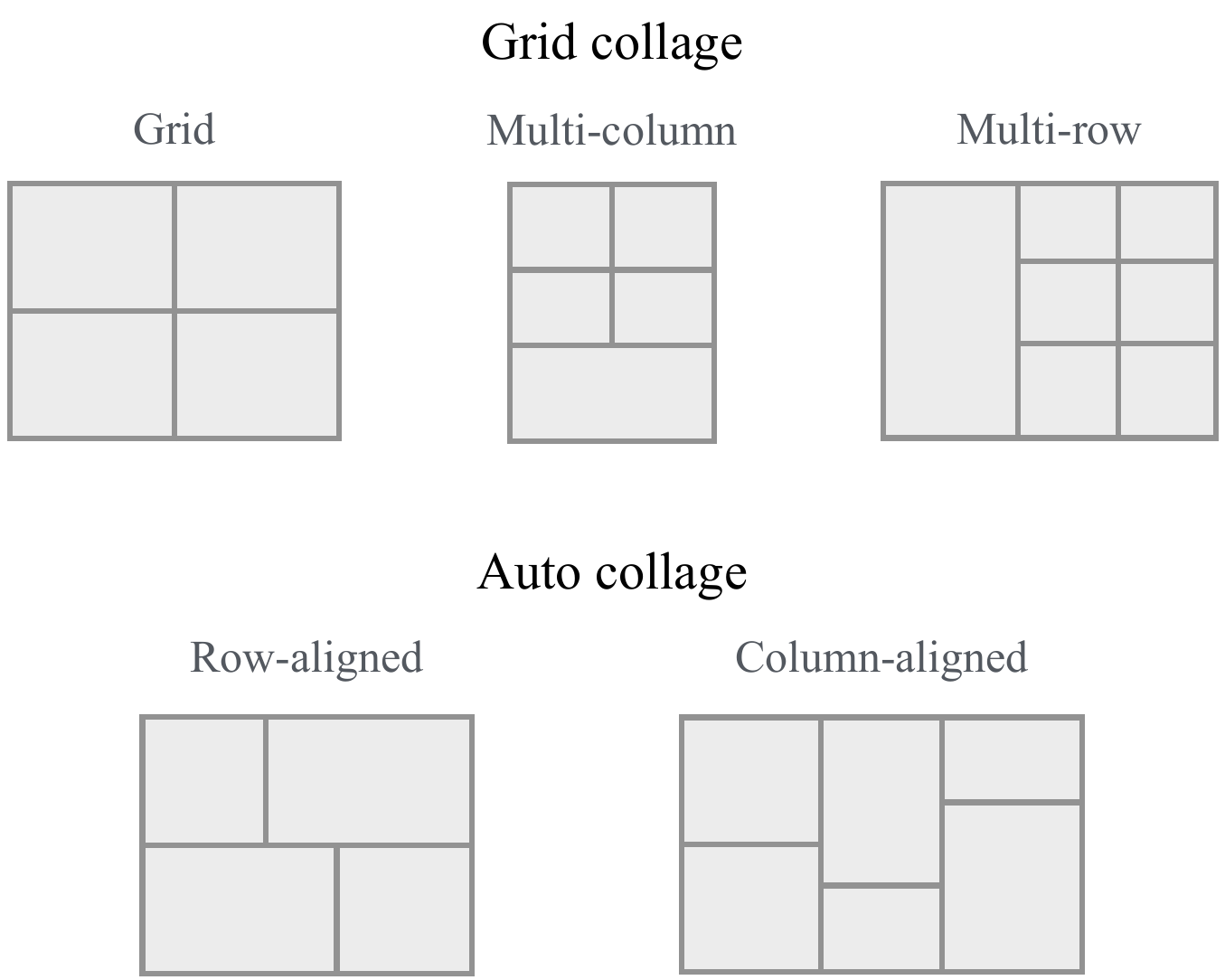}
\caption{Layout examples of grid collage and auto collage.}\vspace{-2em}
\label{fig:layout-template}
\end{wrapfigure}

\thinparagraph{Layout:} We define two collage layouts: grid collage and auto collage. In the grid collage layout, images are arranged in an $n\times m$ grid, where $n,m\in\{1,2,3,4\}$.  To increase layout diversity, cells within the grid can merge to form larger cells. Since rows and columns in the grid layout are aligned, the layout will specify the width/height ratio for each image within a cell, posing constraints for the retrieval process. To further enhance diversity, we introduce the auto collage layout, where only rows or columns are aligned. This enable composing images of arbitrary width/height ratio into a collage image. We demonstrate some examples of the layout in Figure~\ref{fig:layout-template}. 

\thinparagraph{Similarity-based retrieval:} We start by uniformly sampling an image-caption pair as the anchor data $x_\alpha=(I_\alpha,T_\alpha)$ and then retrieve the top 20 most similar image-caption pairs from the database $\mathcal{D}$. Let $I_\alpha^\text{dino}$ and $T_\alpha^\text{clip}$ represent the Dino-v2 image embedding and CLIP text embedding for the anchor data, while $I^\text{dino}$ and $T^\text{clip}$ are the embeddings of an data $x=(I,T)\in\mathcal{D}$. The similarity score between $x_\alpha$ and $x$ is computed as follow:
\begin{align}
    \text{sim}(x_\alpha,x)= \text{cos}(I_\alpha^\text{dino},I^\text{dino})+ \text{cos}(T_\alpha^\text{clip},T^\text{clip}).\nonumber
\end{align}
From the top 20 candidates, we randomly select samples to construct the collage. Where width/height ratios are specified for candidate images, a filter is applied to the database prior to calculating similarity.

\thinparagraph{Entity-based retrieval:}To optimize retrieval, we narrow down the entity list to include only entities that appear more than twice in the dataset. We randomly sample a keyword from this list and apply rule-based matching in the database to select related data. Since such data are sparse, we only use the auto layout to compile collages to avoid the width/height restrictions. In post-processing, we further de-duplicate collages to ensure variety.

In both similarity-based and entity-based retrieval, there may be cases where retrieved images are only loosely related to the anchor image. For instance, when retrieving images based on an anchor image of a cricket game, some results might instead depict baseball due to their visual similarities. However, as long as the corresponding caption accurately describes the image, the final generated caption for the collage will maintain accuracy. This introduces counterfactual samples into the \ours{}-118K dataset, contributing to model debiasing by providing varied contexts.

\begin{figure}[!h]
    \centering
    \captionsetup{justification=centering}
    \vspace{2em}
    \includegraphics[width=\linewidth]{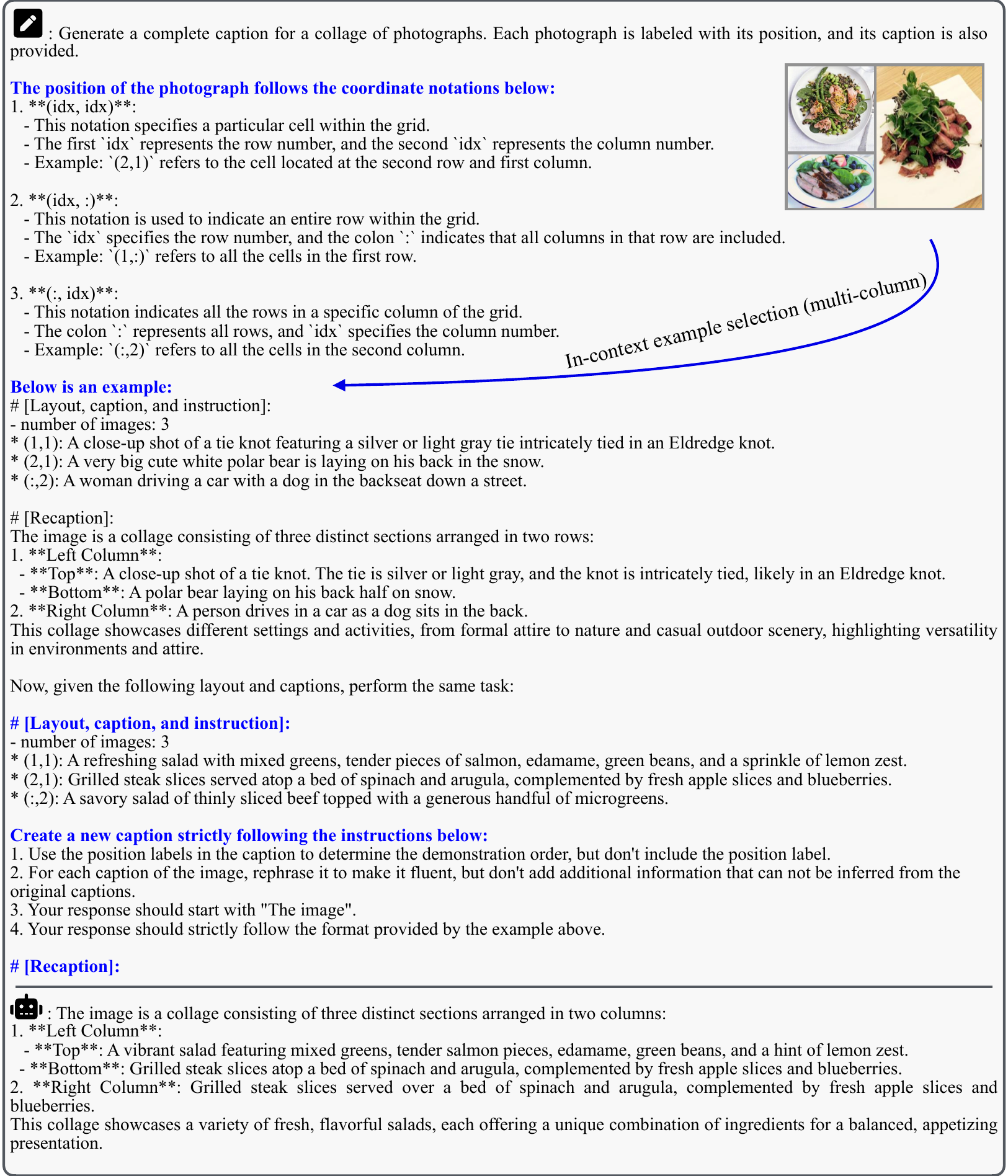}
    \caption{Prompt design and response example for grid layout.}
    \label{fig:collage-grid-prompt-caption}
\end{figure}

\begin{figure}[!h]
    \centering
    \captionsetup{justification=centering}
    \includegraphics[width=\linewidth]{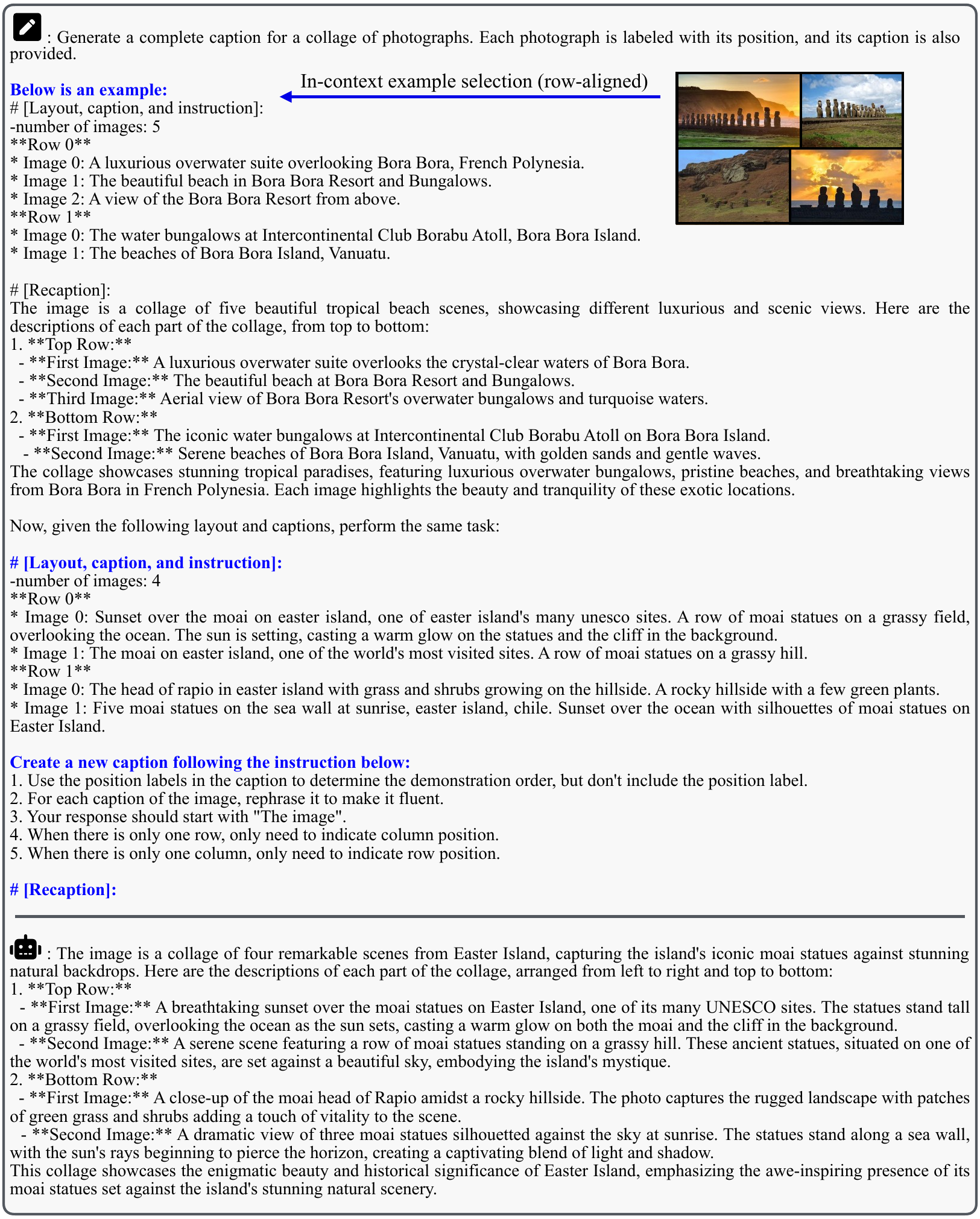}
    \caption{Prompt design and response example for auto layout.}
    \label{fig:collage-auto-prompt-caption}
\end{figure}

\begin{figure}[t]
    \centering
    \captionsetup{justification=centering}
    \includegraphics[width=\linewidth]{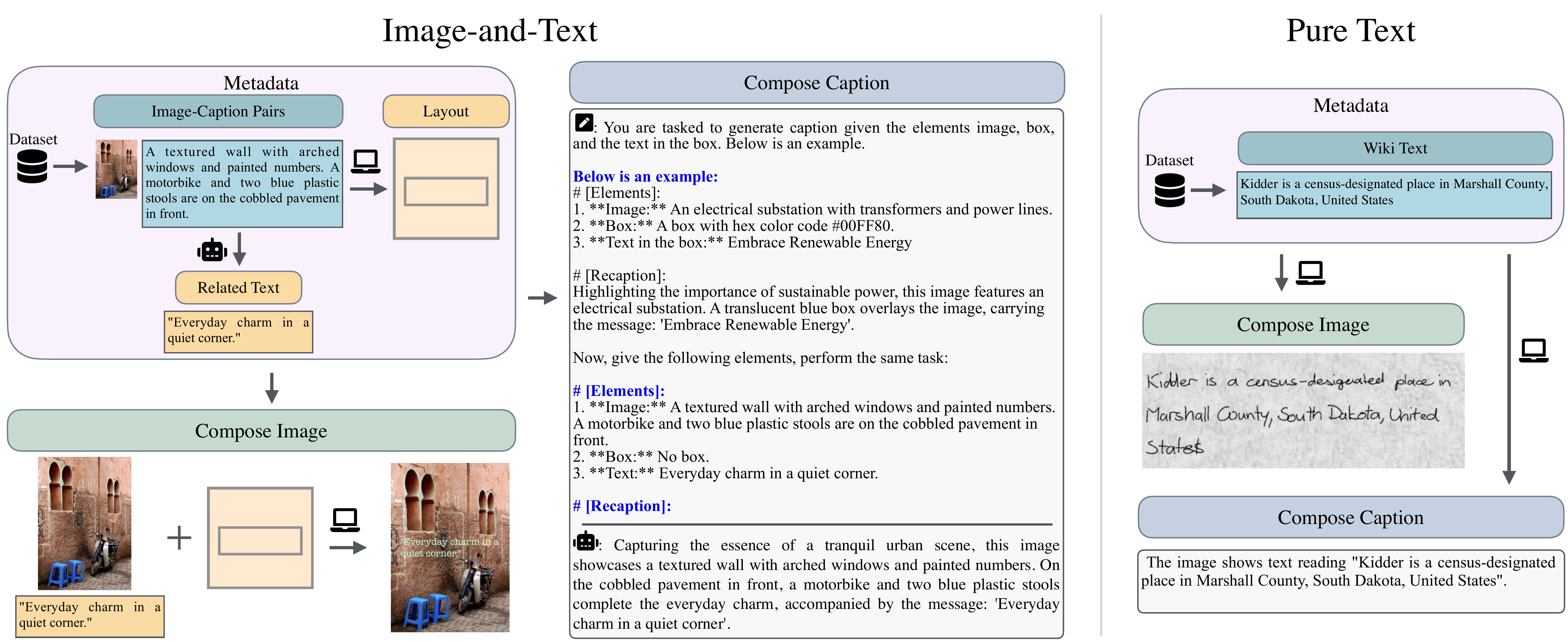}
    \vspace{-1em}
    \caption{The Image-Text implementation.}
    \vspace{-1em}
    \label{fig:image-text-pipe}
\end{figure}

\newpage
\subsubsection{Caption and Prompt Design}\label{sec:collage-caption-detail}
\vspace{-0.5em}
We design the caption such that it provides a detailed walk-through on the images in the collage. Particularly, it either goes over the images by rows or by columns. For each row or column, we specify the demonstration order to be from left to right or from top to bottom, which comes first depends on the generated layout. For instance, for grid collage that comes with a multi-column cell, the caption is designed to go over images by row. And for auto collage whose columns are aligned, the caption is designed to go over images by column. 

In order for LLMs to generate desired caption, different layout uses prompt that are slightly different in terms of the coordinate system notation and the in-context examples. We demonstrate the designed prompts and the example outputs from LLMs in Figures~\ref{fig:collage-grid-prompt-caption} and \ref{fig:collage-auto-prompt-caption}. We use Llama-3.1-70B~\citep{llama3} for all caption generation.

\thinparagraph{Caption post-processing:} We find that some of the responses from LLMs do not completely follow the given instructions. For example, the response may start with ``\#Recaption:''or ``Here is the generated caption:'' before the actual image caption, or contain a follow-up question such as ``Let me know if you have further instruction'' after the caption. To address this, we perform a manual check on a batch of generated responses and summarize all possible patterns, and implement a post-processing pipeline to replace and delete undesired text automatically.

\vspace{-0.5em}
\subsection{\faNewspaper~Image-Text Implementation}\label{app:image-text}
\vspace{-0.5em}
The curation pipeline for the Image-Text CI class is illustrated in Figure~\ref{fig:image-text-pipe}. This class is designed to assess MLLMs' capabilities in extracting text from images and understanding the relationship between text and visuals. We divide this into two subcategories: image-and-text and pure-text. The image-and-text category tests the model's ability to infer the relationship between text and the image, expecting MLLMs to interpret how text interacts with visual elements. The pure-text category focuses solely on text extraction. We present text in various styles against different backgrounds to strengthen the MLLMs' robustness in text recognition.

\begin{figure}[h]
    \centering
    \captionsetup{justification=centering}
    \includegraphics[width=\linewidth]{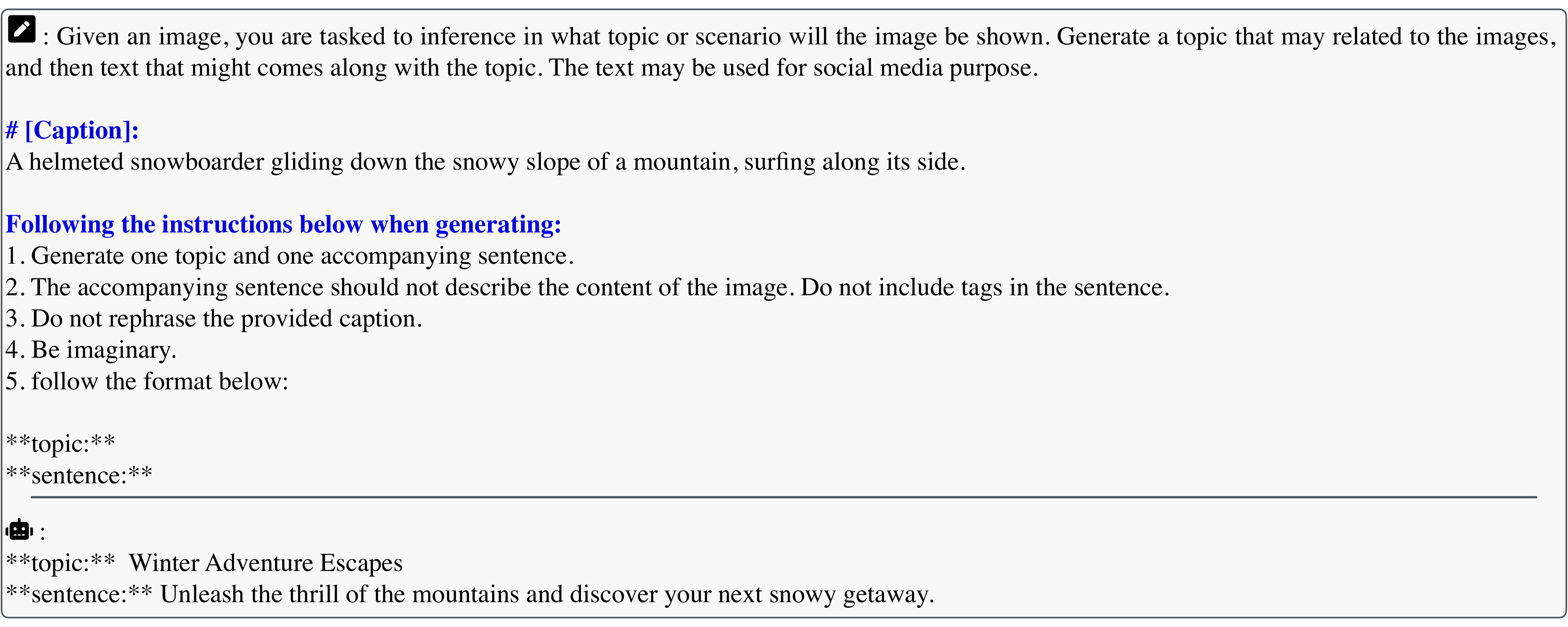}
    \caption{Prompt for generating related text from given image.}
    \label{fig:image-and-text-topic-prompt}
\end{figure}

\vspace{-0.5em}
\subsubsection{Image-and-text Pipeline}
\vspace{-0.5em}
Similar to collage, we sample image-caption pair as the background. We only consider random retrieval as we only retrieve one image at a time. We then prompt an LLM to generate relevant text content to the image based on the caption. Note that we specifically instruct the LLM not to rephrase the caption. The prompt used for text content generation is shown in Figure~\ref{fig:image-and-text-topic-prompt}. We first ask Llama-3.1-70B to infer the topic related to the image, then generate a sentence within the topic.
To enhance visual diversity, we control two primary configurations:

\thinparagraph{Box layout:} We position the text within a bounding box, arranging the box alongside or overlaying the image. The box's size, color, and opacity are randomized to increase variety.
\thinparagraph{Text style:} We customize the text’s appearance by adjusting its size, color, font, and spacing, ensuring it contrasts well with the background for clear visibility.

\subsubsection{Pure Text Pipeline}
This pipeline synthesizes both digital and handwritten text images. The text corpus is sourced from Wikipedia, with digital text generated as in the image-and-text pipeline. Additional details for handwritten text and background generation are as follows:
\input{tables/hw_effect}

\thinparagraph{Handwritten text:} We generate handwritten text images in SVG format following \citet{graves2013generating}, offering 12 distinct writing styles. Similar to digital text, color, stroke width, line spacing, and alignment are customized to increase diversity.
\thinparagraph{Background:} We choose two types of background image:
\begin{compactenum}
\item \textit{Natural image.} We sample images from COCO dataset and apply a blurring effect to highlight the foreground text.
\item \textit{Synthetic paper image.} We use Augraphy~\citep{augraphy_paper} to render
realistic document effect, which sequentially modifies the ink style and the background paper style to create an authentic appearance. We summarize the used pipelines in Table~\ref{table:augmentors}.
\end{compactenum}




\subsection{\faChartBar~Chart Implementation}\label{app:chart}

For chart visualizations, we consider bar charts, line plots, pie charts, and choropleth maps. In this section, we first explain how the bar, line and pie charts and their captions are generated  generated as they share similar data sources. Then we illustrate the map visualization and caption design. Finally, we provide a comparison of our curate dataset against previous synthetic chart-text datasets. For all chart visualizations, we use Plotly~\citep{plotly}. And for all caption generations, we use Llama-3.1-405B.

\subsubsection{Bar, Line, and Pie Charts}

\thinparagraph{Data source:} The tabular data for visualization come from DVQA~\citep{kafle2018dvqa} and UniChart~\citep{masry2023unichart}. DVQA provides canonical tabular data suitable for bar and pie chart visualizations, while UniChart contains time-series tabular data for line charts.

\thinparagraph{Bar chart:} The bar chart generation pipeline supports three bar types: single, grouped, and stacked bar charts. Single bar charts visualize one row of data, whereas grouped and stacked bar charts incorporate multiple rows. To enhance variety, each bar type includes the following customizations:

\begin{compactenum}
\item \textit{Bar pattern.} Adjustments include bar texture, color, border, width, spacing, and the presence or absence of text on the bars.
\item \textit{Orientation.} Bars can be arranged horizontally or vertically.
\item \textit{Axes.} Customizations cover the range and tick intervals on both x-axis and y-axis.
\item \textit{Annotations and layouts.} Variations include font styles, colors, and layout adjustments for titles, axis labels, and legends.
\end{compactenum}

\thinparagraph{Line chart:} For line charts, we use both single-row and multiple-row time-series data, where each line corresponds to one line in a chart. Many customizations from bar charts apply here, including axes, annotations, and layouts. The primary distinction for line charts is in line pattern customization, such as line style, color, and marker pattern.

\thinparagraph{Pie chart:} Pie charts use single-row data for visualization. Customizations include pie appearance adjustments, such as color, size, and display text placement. Text can be displayed inside or outside the pie; when segments are too small for text, pointers are used to indicate the designated region. Other customizations align with those used in bar and line charts.

\thinparagraph{Prompt and caption design:} For generating captions, the input to LLMs includes both data details and visualization parameters. Specifically, it incorporates axes details (type, range, and label) and additional elements like background patterns, titles, and style specifications. We instruct LLMs to create captions that summarize the chart’s data, identify trends, and compare groups. Figures~\ref{fig:bar_prompt},~\ref{fig:line_prompt},~\ref{fig:pie_prompt} illustrate the designed prompts and sample outputs.

\subsubsection{Choropleth Maps} 
Choropleth maps are created for four regions: European countries, global countries, the United States, and Chinese provinces.

\thinparagraph{Data source:} For European and global countries, data is visualized at the country level, with each country assigned a data value. Data is sourced from Eurostat~\citep{eurostat}  and Gapminder~\citep{gapminder}, or generated randomly. For Chinese provinces, we use randomly generated data, while for the United States, randomly generated state data from MapQA ~\citep{chang2022mapqa} is used.

\thinparagraph{Visualization:} Depending on the data type, choropleth maps can represent values using either a color bar for numerical data or a discrete color legend for categorical data. Each region is colored based on its value in the legend or color bar. Various visualization customizations include:

\begin{compactenum}
\item \textit{Color pattern.} Varying color schemes for regions, titles, and legends.
\item \textit{Projection.}  Different projection methods for map rendering.
\item \textit{Value dropout.} Randomly omitting values for certain regions and marking these with a distinct color.
\item \textit{Layouts.} Randomized layout of titles, map entries, and legends.
\item \textit{Region annotation.} Optional display of country/province names or acronyms within the map.
\end{compactenum}

\thinparagraph{Prompt and caption design:} We focus captions on regions that are clearly visible on the map to ensure clarity. Along with listing data values for key regions, we analyze the overall data distribution, such as trends by cardinal direction or differences between coastal and inland areas. We also include additional details like title and legend interpretation.

Figure~\ref{fig:map_prompt} shows the prompt used to generate captions. When preparing the data table in the prompt, we include each region’s data value, color name, and some geographic details:
\begin{compactenum}
\item\textit{Location.} Compass direction (e.g., north, southeast).
\item \textit{Type.} Whether the region is coastal or inland.
\item \textit{Area.} The size of the region.
\end{compactenum}

\subsubsection{Post-processing}
We filter and modify the generated responses from LLMs such that they mostly resembles caption of an image. We observe that the generated response describes the details without first identifying the type of the chart since it is provided in the context of the prompt. However, such information is not granted in real conversation. We rephrase the first sentence such that it always start with identifying the chart type presented in the image before actual captioning. For instance, ``{\textcolor{gray}{The line chart titled ``xxxx'' visually represents...}'' is rephrased into ``{\textcolor{gray}{The image shows a line chart titled ``xxxx'', which visually represents...}}''. We implement it by rule-based matching and replacement. Apart from first sentence rephrasing, we also reuse the processing strategies stated in \S~\ref{sec:collage-caption-detail} to enhance caption quality.

\begin{table}[t]
    \centering
    \captionsetup{justification=centering}
      \resizebox{\textwidth}{!}{
    \begin{tabular}{lccccc}
    \toprule
   & FigureQA & DVQA & PlotQA & MapQA& \ours{}-Chart\\\midrule
   Image type & Scatter/Bar/Line/Pie & Bar & Scatter/Bar/Line & Map & Bar/Line/Pie/Map \\
   Text type & Yes/No QA &Open-ended QA &Open-ended QA &Open-ended QA & Detailed Caption \\
   Text generation & Template-based & Template-based & Template-based & Template-based & LLM-generated\\
    \bottomrule
    \end{tabular}}
    \caption{Comparison of existing synthetic chart datasets.}
    \label{tab:chart_dataset_summary}
\end{table}

\subsubsection{Comparison with Existing Synthetic Datasets} Table~\ref{tab:chart_dataset_summary} compares existing synthetic chart-QA datasets (FigureQA~\citep{kahou2017figureqa}, DVQA~\citep{kafle2018dvqa}, PlotQA~\citep{methani2020plotqa}, and MapQA~\citep{chang2022mapqa}) with our chart-caption dataset. Unlike previous methods that generate templated question-answer pairs for charts, our pipeline emphasizes detailed captions.

Previous methods revoke a system to learn three abilities: structure understanding, data retrieval, and reasoning, often through carefully designed templates targeting a single ability. By providing precise instructions to the LLM, we enable it to generate captions that naturally integrate all three abilities. This approach not only eliminates the need for rigid templates but also encourage diversity in the generated captions.

\begin{figure}[!h]
    \centering
    \captionsetup{justification=centering}
        \includegraphics[width=\linewidth]{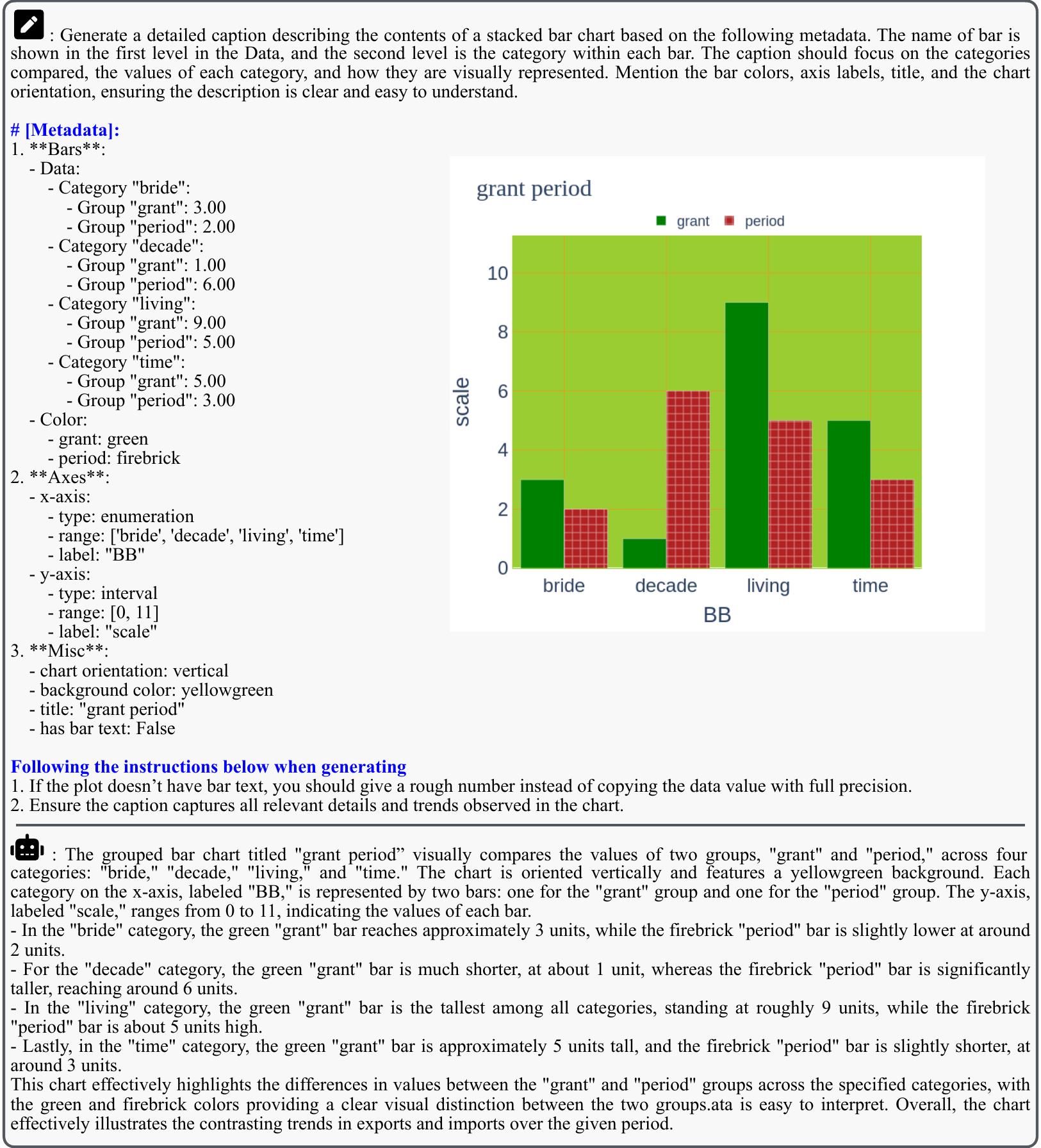}
        \caption{{Prompt design and response example for bar chart.}}
        \label{fig:bar_prompt}
\end{figure}
\clearpage

\begin{figure}[!h]
    \centering
    \captionsetup{justification=centering}
        \includegraphics[width=\linewidth]{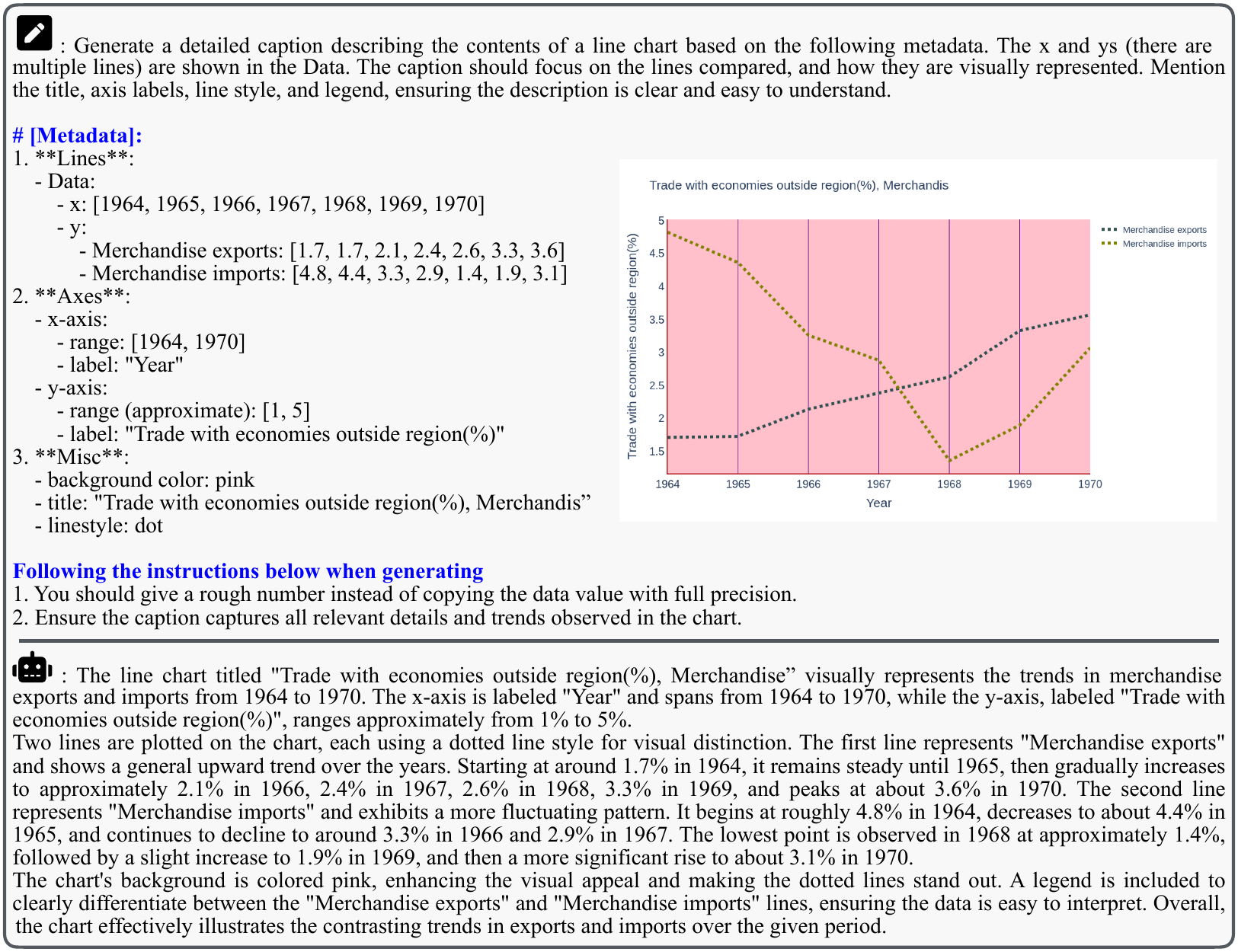}
        \caption{{Prompt design and response example for line plot.}}
        \label{fig:line_prompt}
\end{figure}
\clearpage

\begin{figure}[!h]
    \centering
    \captionsetup{justification=centering}
        \includegraphics[width=\linewidth]{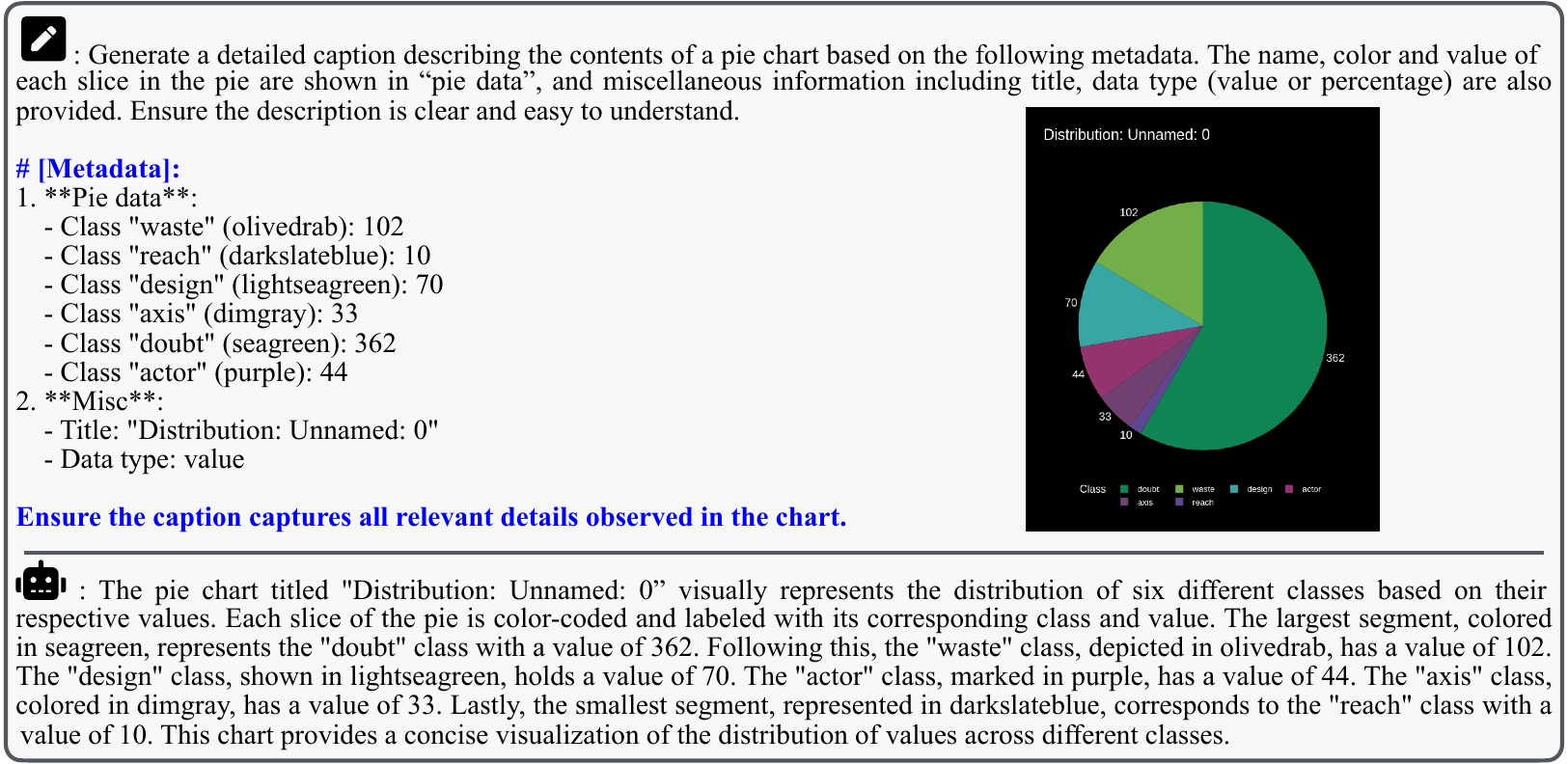}
        \caption{{Prompt design and response example for pie chart.}}
        \label{fig:pie_prompt}
\end{figure}
\clearpage

\begin{figure}[!h]
    \centering
    \captionsetup{justification=centering}
        \includegraphics[width=\linewidth]{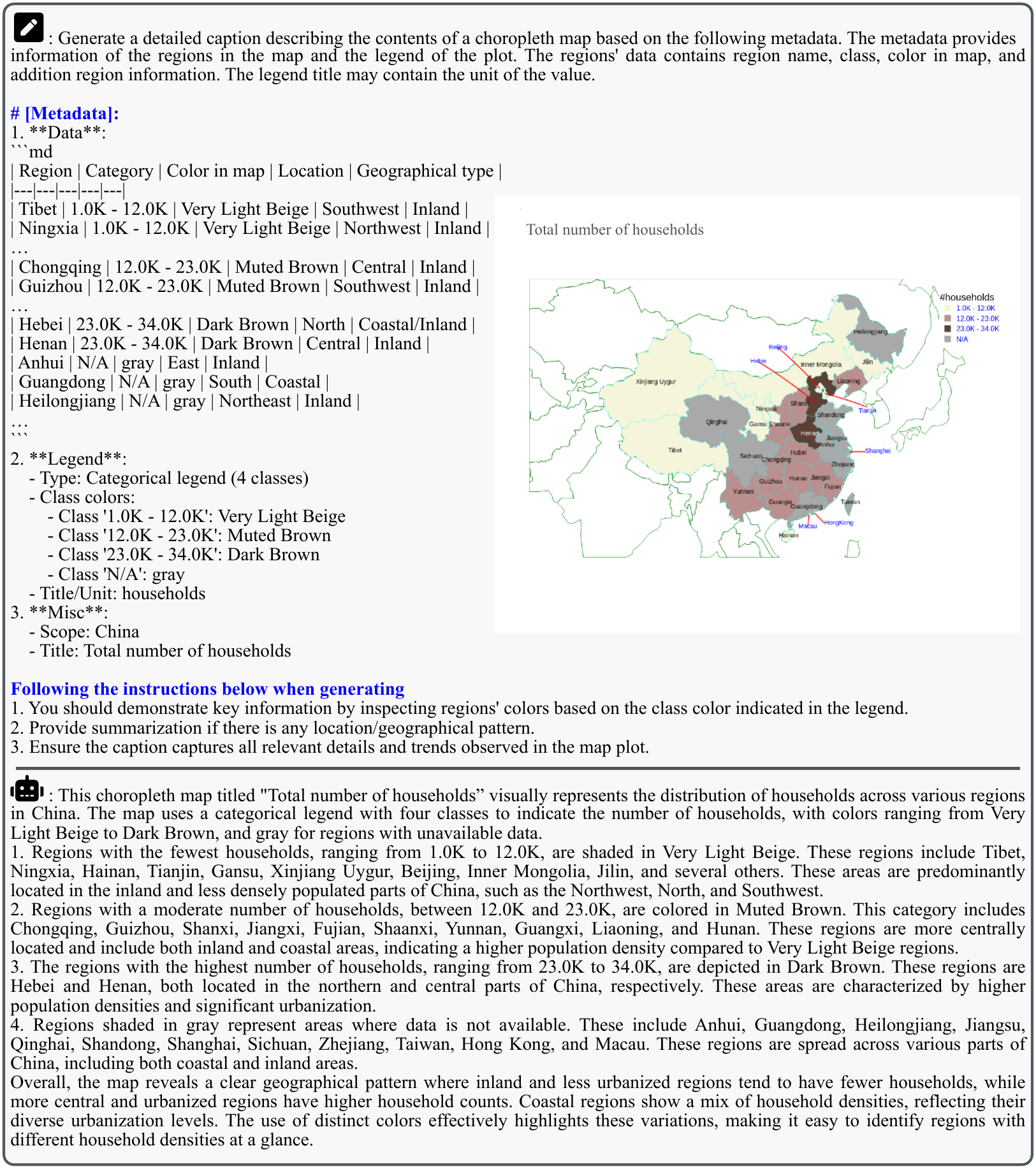}
        \caption{{Prompt design and response example for choropleth map.}}
        \label{fig:map_prompt}
\end{figure}
\clearpage

\subsection{\faSitemap~Diagram Implementation}\label{app:diagram}
We employ Mermaid~\citep{mermaid}, a JavaScript-based diagramming tool, to convert markdown text into diagram images. This tool allows us to transform text into visual representations seamlessly. Additionally, we prompt LLMs to analyze the markdown text, generating captions that not only describe each element in the diagram but also clarify relationships and provide potential insights. Figure~\ref{fig:diagram-pipe} demonstrates the implementation of the pipeline. 
\begin{figure*}
    \centering
    \captionsetup{justification=centering}
    \includegraphics[width=\linewidth]{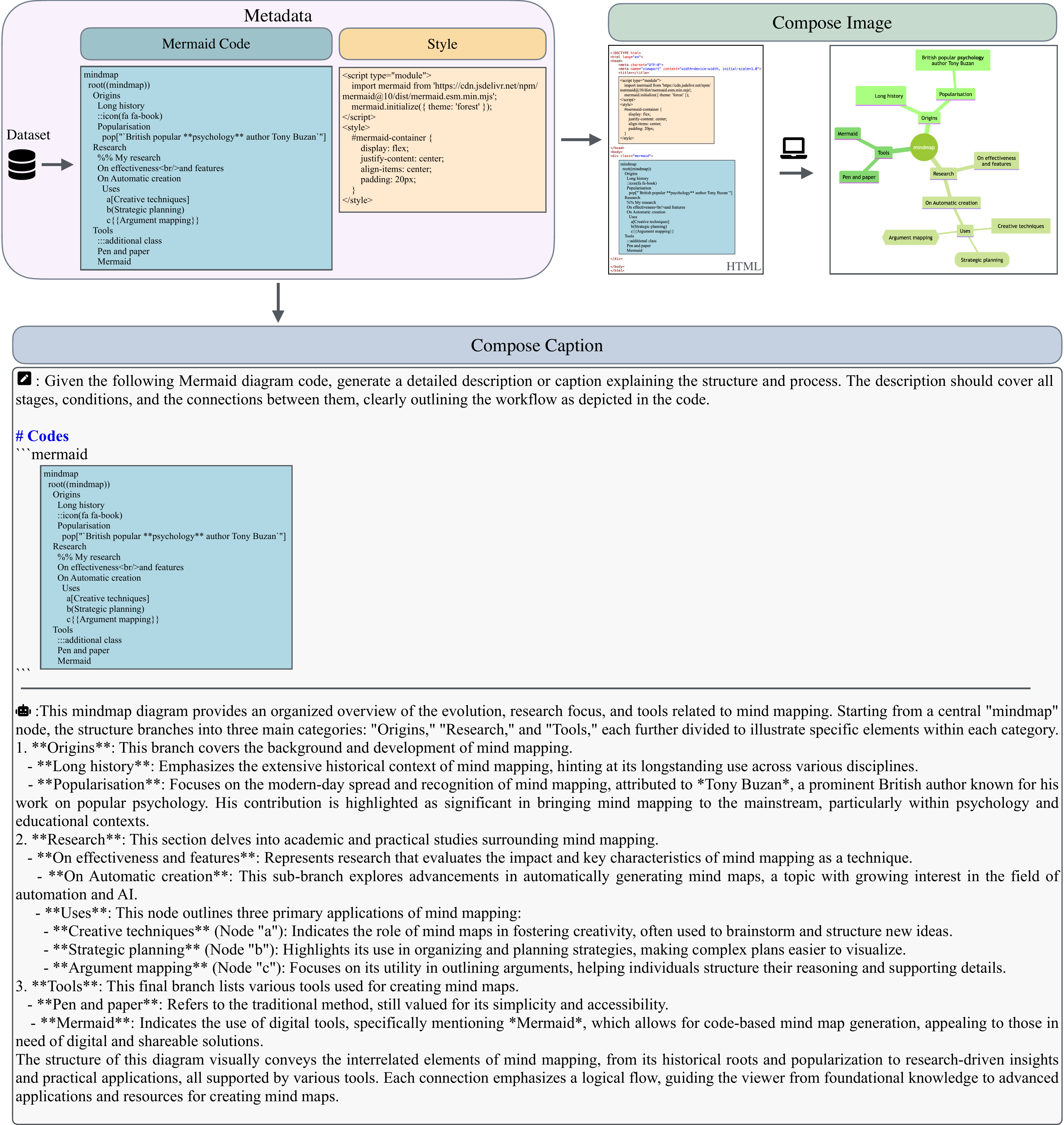}
    \caption{{The Diagram implementation.}}
    \label{fig:diagram-pipe}
\end{figure*}

\subsubsection{Diagram Visualization}
This section elaborates on the data source of the Mermaid codes, how they are rendered into diagram images, and customizations of the diagram style.
\thinparagraph{Data source:} We acquire Mermaid code through two primary methods:
\begin{compactenum}
    \item \textit{GitHub Crawling.} We collect text files containing ``mermaid'' as a keyword from licensed repositories on GitHub.
    \item \textit{LLM-Generated Code.} We prompt LLMs to generate Mermaid code for specific diagram types such as class diagrams, ER diagrams, and flowcharts.
\end{compactenum}
We filter the collected codes by running a rendering test, yielding approximately 3K valid diagram codes: 2K from GitHub and 1K generated by LLMs.

\thinparagraph{Rendering:} Mermaid’s advantage lies in its automatic optimization of spatial arrangements, ensuring diagrams display well in HTML. By simply declaring the required packages and placing the Mermaid code within the HTML body, the browser renders the diagram seamlessly. In our process, we generate an HTML file for each Mermaid code, open it in Chrome, and capture a screenshot of the rendered diagram. We use Selenium to automate this process of browsing and saving images.

\thinparagraph{Diagram style customizations:} Mermaid also offers styling parameters to customize the theme and visual appearance of rendered diagrams. These parameters can be included in the HTML header and thus separated from the main diagram content. We prompt LLMs to generate 53 styling specifications, creating a candidate set. For each HTML file, we randomly select one styling option from this set to increase visual diversity. In cases where a styling option is incompatible with a specific diagram type, the default styling is automatically applied. We retain all successfully rendered HTML files.

\subsubsection{Prompt and Caption Design}
Understanding diagrams is more challenging because they contain numerous objects and emphasize the relationships among them. Specifically, object relationships in diagrams are often more complex compared to other CI types, as they frequently use arrows or nesting to indicate directions or hierarchies. Therefore, our designed captions focus on extracting elements and relationships, placing less emphasis on the diagram's appearance details.

To generate captions that provide a detailed breakdown of the diagrams, we prompt Llama-3.1-405B to read the diagram code and translate it into natural language. To ensure the generated captions are as invariant as possible to the diagram's appearance, we include only the Mermaid code in the prompt, excluding any styling-related codes. We find that minimal instruction is sufficient for the LLM to accurately analyze the code.

In post-processing the generated captions, we first modify the opening sentence to include an identification of the diagram type, similar to our approach with chart captions. Some Mermaid code retrieved from GitHub contains style arguments like hex color codes or stroke widths for text boxes. Since the LLM interprets code, these styling details sometimes appear in their responses. For example, a box labeled ``\textcolor{gray}{Customer A}'' might be described in the caption as ``\textcolor{gray}{Customer A (\#a1320f, stroke width 2)}''. This pattern also occurs when the diagram code assigns a shorter variable name (e.g., ``A'') to an object like ``Customer A''. To enhance caption quality, we refine the LLM-generated responses by removing parentheses that contain styling arguments or variable names.
 
\newpage
\subsection{\faCode~Code Implementation}\label{app:code}
We use Carbon~\citep{carbon_now} to create code screenshots with a customized syntax theme and font style. Specifically, it provides 29 themes and 14 font families, as well as other style parameters such as font color, presence or absence of line numbers, line space, window size, etc. We randomize those options to enhance the diversity of the generated screenshots. Next, we demonstrate what code data are used for screenshot rendering and the caption design.

\thinparagraph{Data source:} While numerous code generation datasets provide a variety of sources for code snippets, we find these datasets often contain overly complex, lengthy examples with extensive comments. This complexity results in code snippets that are too long to fit within a suitably sized screenshot. Additionally, the detailed comments provide explicit explanations of functionality, whereas we aim for MLLMs to learn inference directly from the code itself, without relying on predefined explanations. Therefore, we seek to use LLMs to generate simpler code snippets in different languages.

We consider 9 programming languages: C, C++, Ruby, R, Python, Java, JavaScript, CSS, and SQL. For each language, we ask Llama-3.1-70B to generate 200 functions/topics that can be implemented by the selected language. We then prompt the LLM to generate the code given the topic and the language. We focus on relatively simple functions such as mathematical implementations, textbook algorithms and use case of data structures. 

\thinparagraph{Caption design:} Our goal is for MLLMs to first extract the code text from the screenshot, interpret it, and then provide an explanation of its functionality The code explanation can be obtained by LLMs. Building on the previous code generation step, we further have the Llama-3.1-70B to generate the corresponding explanation for its generated code snippet. The code snippet and its corresponding explanation are then concatenated together to compose the caption.  Figure~\ref{fig:code-caption} illustrates an example code screenshot and its composed caption.


\begin{figure}[t]
    \centering
    \captionsetup{justification=centering}
    \includegraphics[width=\linewidth]{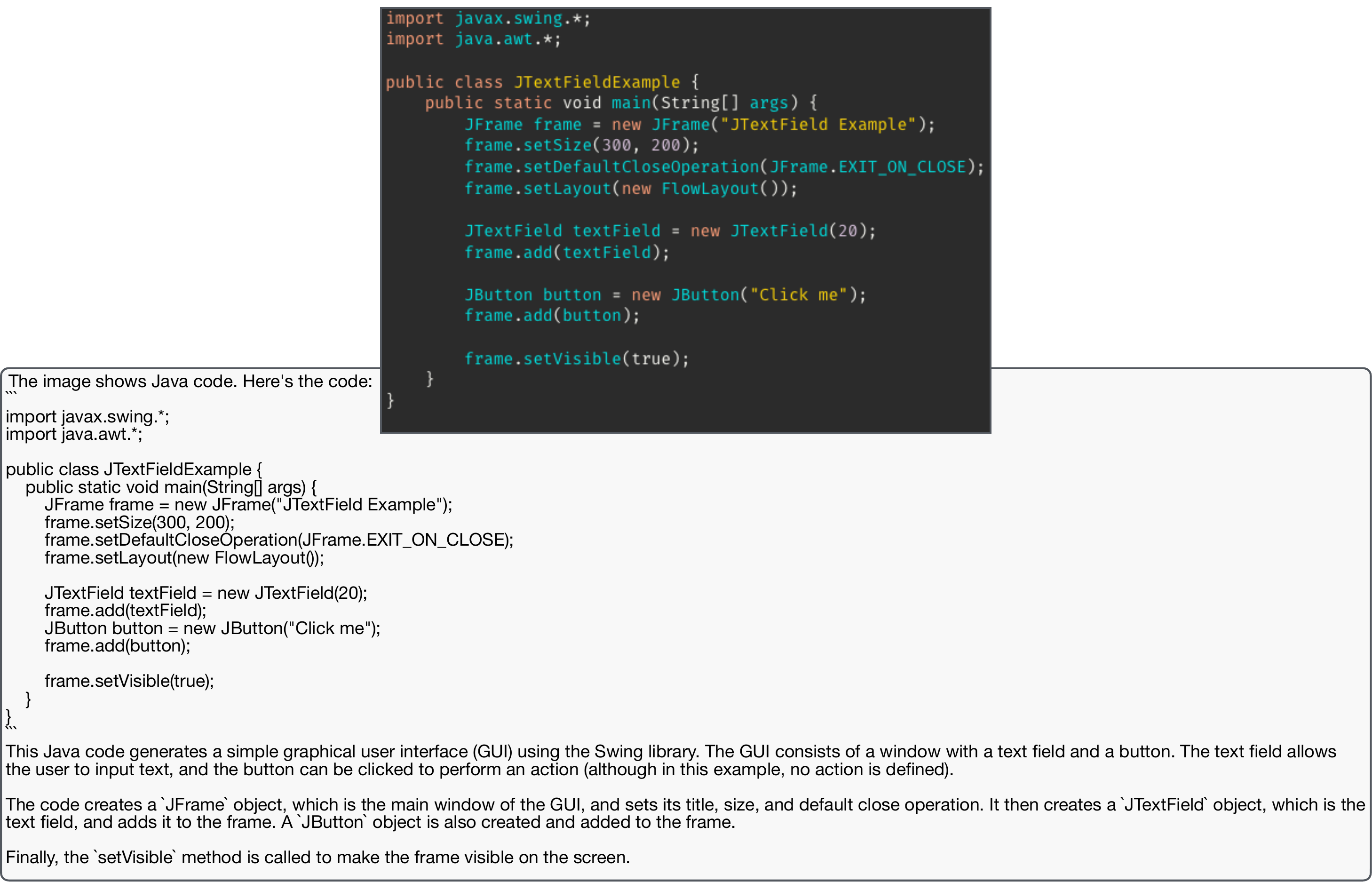}
    \caption{{Caption design for code screenshot.}}
    \label{fig:code-caption}
\end{figure}
\begin{figure}[h]
    \centering
    \captionsetup{justification=centering}
    \includegraphics[width=0.8\linewidth]{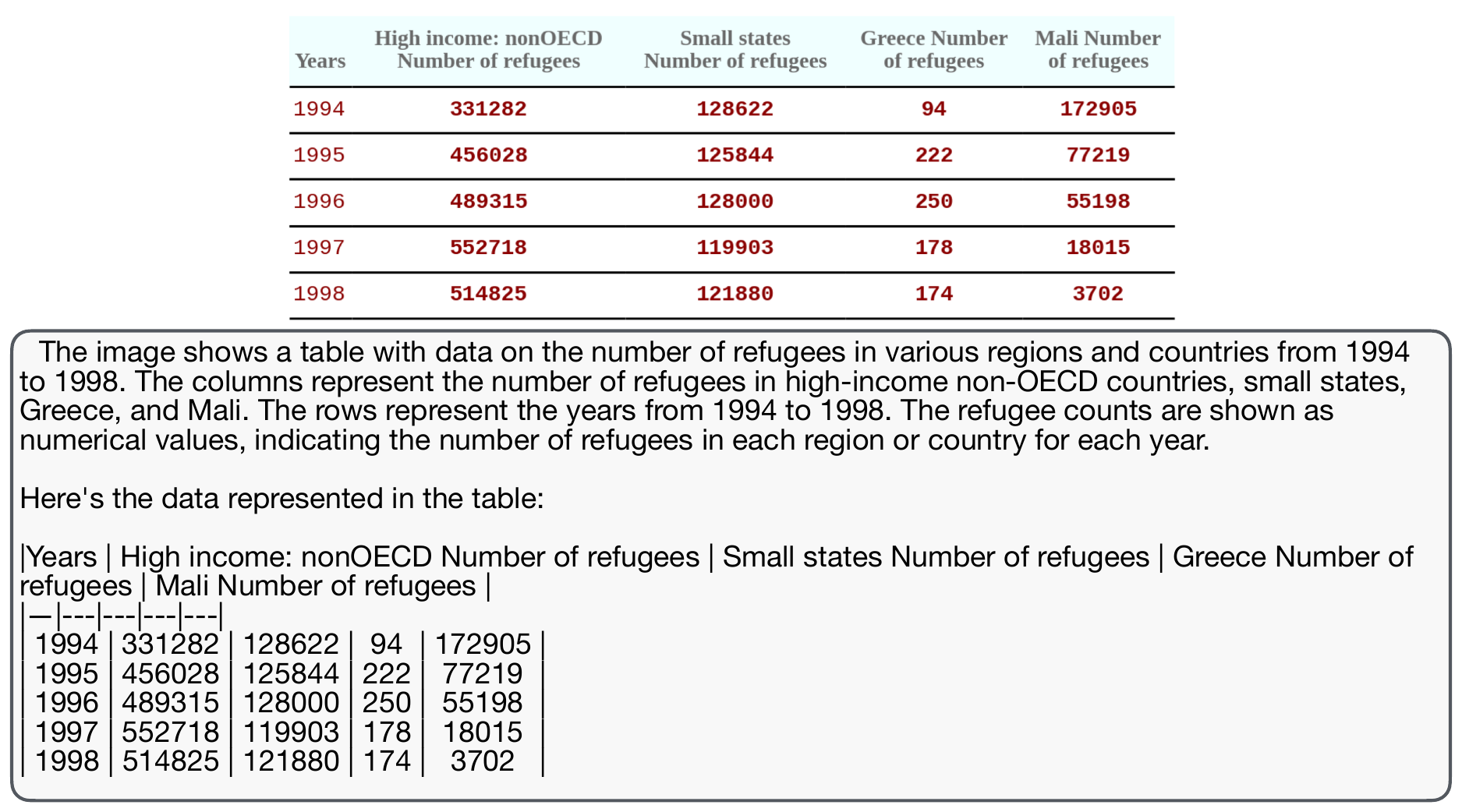}
    \caption{{Caption design for table image.}}
    \label{fig:table-caption}
\end{figure}

\subsection{\faTable~Table Implementation}\label{app:table}

We use Matplotlib~\citep{matplotlib} to generate table images from tabular data. Similar to our approach for chart visualization, the tabular data for these table images is sourced from the UniChart dataset. Below, we outline the different types of customizations applied to the table images, followed by a discussion of the caption design.

\thinparagraph{Visualization:} The following table customizations were applied:
\begin{compactenum} 
\item \textit{Table size}: Varying the table's width and height. 
\item \textit{Cell style}: Adjusting the width, height, and color of individual rows or columns. 
\item \textit{Border style}: Modifying the style, thickness, and color of table borders. 
\item \textit{Font style}: Varying font color, type, and size. 
\item \textit{Alignment}: Applying different alignments (left, center, right) to individual columns. 
\end{compactenum}
We ensure that the data remains clearly visible despite these customizations. Specifically, we maintain strong contrast between foreground and background colors, and adjust font and table sizes appropriately for readability. 

\thinparagraph{Caption design.} 
The caption design follows similar principles to those used in code image captions, focusing on extracting the table from the image and analyzing the presented data. To obtain the analysis text, we convert the table into markdown format and prompt the Llama-3.1-70B to generate a response that provides a summary and detailed breakdowns. The final caption for the table image is composed by combining the markdown table with the analysis text. Figure~\ref{fig:table-caption} shows an example table image and its composed caption.

\section{Experiment Details}\label{app:exp_setting}
\subsection{Demonstrative Experiment}\label{app:exp_demo}
\thinparagraph{Collecting VQA data for NIs and CIs:} We randomly select 1K samples from the VQAv2~\citep{balanced_vqa_v2} validation set to create VQA pairs for NI.  However, to our knowledge, no existing benchmark comprehensively covers the wide variety of CI types. To address this, we curate a toy benchmark comprising 1K VQA pairs for CIs by sampling from datasets ChartQA~\citep{chartqa}, DocVQA~\citep{docvqa}, InfoVQA~\citep{infovqa}, MapQA~\citep{chang2022mapqa}, MME~\citep{yin2023survey}, OCRBench~\citep{liu2023hidden}, MMVet~\citep{yu2023mm}, and MMBench~\citep{liu2025mmbench}. This curated benchmark includes various CI types collages, charts, tables, code, documents, diagrams, infographics, etc, to offer a broad evaluation of MLLMs’ CI comprehension abilities.  The composition of this curated benchmark is summarized in Table~\ref{tab:demo-dataset-composition}
\input{tables/composition_demo_benchmark}

\thinparagraph{Inference:} We evaluate the MLLMs’ caption and VQA accuracy using the provided QA pairs.  In the caption-conditioned QA pipeline, the captions generated by MLLMs may sometimes lack sufficient context for an LLM to answer the visual question accurately. We instruct the LLM to respond with ``I don’t know'' (IDK) when it identifies insufficient information in the caption. We count the IDK percentages for NIs and CIs. Table~\ref{tab:demo-idk} shows a higher IDK rate for CIs than for NIs, indicating that CI captions tend to capture less information or information of lesser relevance.



\begin{minipage}{0.48\textwidth}
\begin{center}
\centering
\captionsetup{type=table}
\captionsetup{justification=centering}
\begin{tabular}{lcc}\toprule
& CI  & NI \\\midrule
IDK percentage  &  58\% & 12\%\\\bottomrule
\end{tabular}
\caption{{IDK percentage:} Percentage of an LLM answering IDK given the visual question and caption context.}
\label{tab:demo-idk}
\end{center}
\end{minipage}
\hfill
\begin{minipage}{0.48\textwidth}
\begin{center}
\centering
\captionsetup{type=figure}
\captionsetup{justification=centering}
\includegraphics[width=\linewidth]{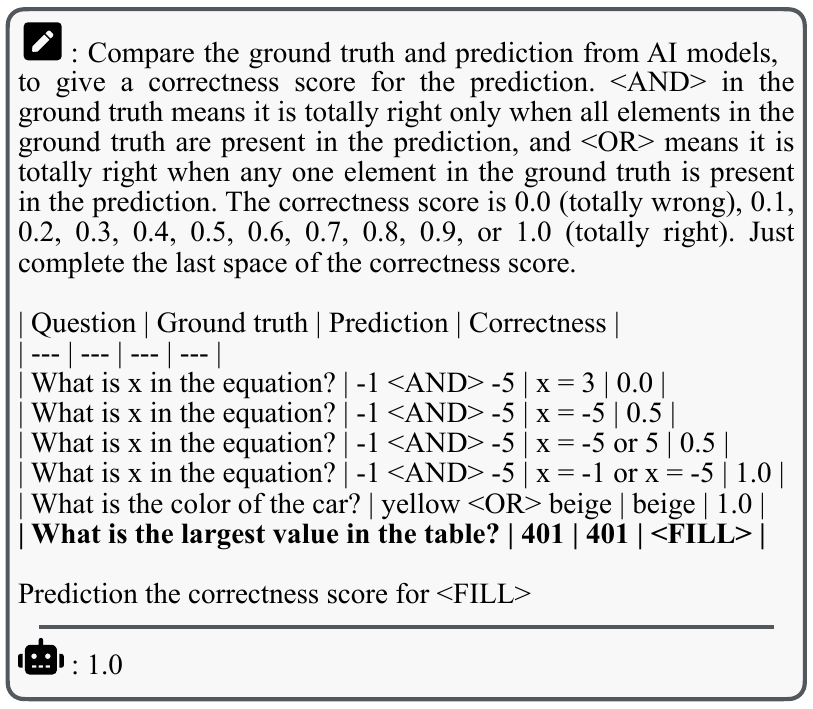}
\caption{{Prompt for evaluating of MLLMs' predictions.}}
\label{fig:mmvet-prompt}
\end{center}
\end{minipage}
\thinparagraph{Evaluation:} We only evaluate MLLMs' caption accuracy on VQA pairs that are answerable by LLMs. We discard original benchmark evaluation guidelines, as each has its own rubric, and instead adopt MMVet’s evaluation approach by prompting GPT-4 to score the predicted answers. We demonstrate the prompt used for predicting answer correctness in Figure~\ref{fig:mmvet-prompt}. Figure~\ref{fig:cap-vqa-demo}b reports the average prediction score as measures of caption accuracy and VQA accuracy for both CI and NI. For agreement percentage in Figure~\ref{fig:cap-vqa-demo}c, we evaluated agreement by checking the correctness scores for each VQA pair; we considered VQA and caption predictions to be in agreement if the absolute difference between their scores was less than 0.2.
\subsection{MLLMs Training}
\thinparagraph{Pre-training:} We do not pre-train the MLLMs in our experiment. Instead, we directly use the public pre-trained MLLMs checkpoint to initialize the weights and focus on the SFT stage. For xGen-MM, we use the v1.5 checkpoint to initialize the model weights.
\thinparagraph{SFT data recipe for xGen-MM-inst:} Since the dataset used to instruction fine-tune xGen-MM is not released, we curate a similar SFT dataset according to the data mixture mentioned in their paper~\citep{xue2024xgen}. Specifically, we include 781K image-text instruction data from various domains~\citep{Liu2023ImprovedBW,chartqa,kafle2018dvqa,docvqa,lin2014microsoft,yan2024list,lindstrom2022clevr,ainslie2023gqa,mishra2019ocr,krishna2017visual,chen2023sharegpt4v,textvqa,lu2022learn,kembhavi2016diagram}, and 211K pure text instruction following data~\citep{mukherjee2023orca,cobbe2021training,zhou2024lima}. We retrain xGen-MM-inst-4B using our own curated dataset for a fair comparison.
\input{tables/training_hyperparams}

\thinparagraph{Hyperparameters:} We show the training hyperparameters for the \ours{} series in Table~\ref{tab:hyperparams}. The reproduction of xGen-MM-inst. follows the same hyperparameters as in \ours{}-4B. We use 8 Nvidia A100 GPUs to train the 4B MLLMs and 32 Nvidia A100 GPUs for the 7B and 13B MLLMs.

\subsection{ChartQA Image Captioning}
We employ an superior MLLM to generate captions for chart images in the ChartQA training set. This process produces a total of 18,317 chart-caption pairs. As there are multiple instruction data corresponding to one chart image, we replace data in an image-level in the caption-instruction ablation study. Figure~\ref{fig:chart-gpt-caption-example} shows some examples of chart caption.



\begin{figure}[h]
    \centering\captionsetup{justification=centering}

    \includegraphics[width=\linewidth]{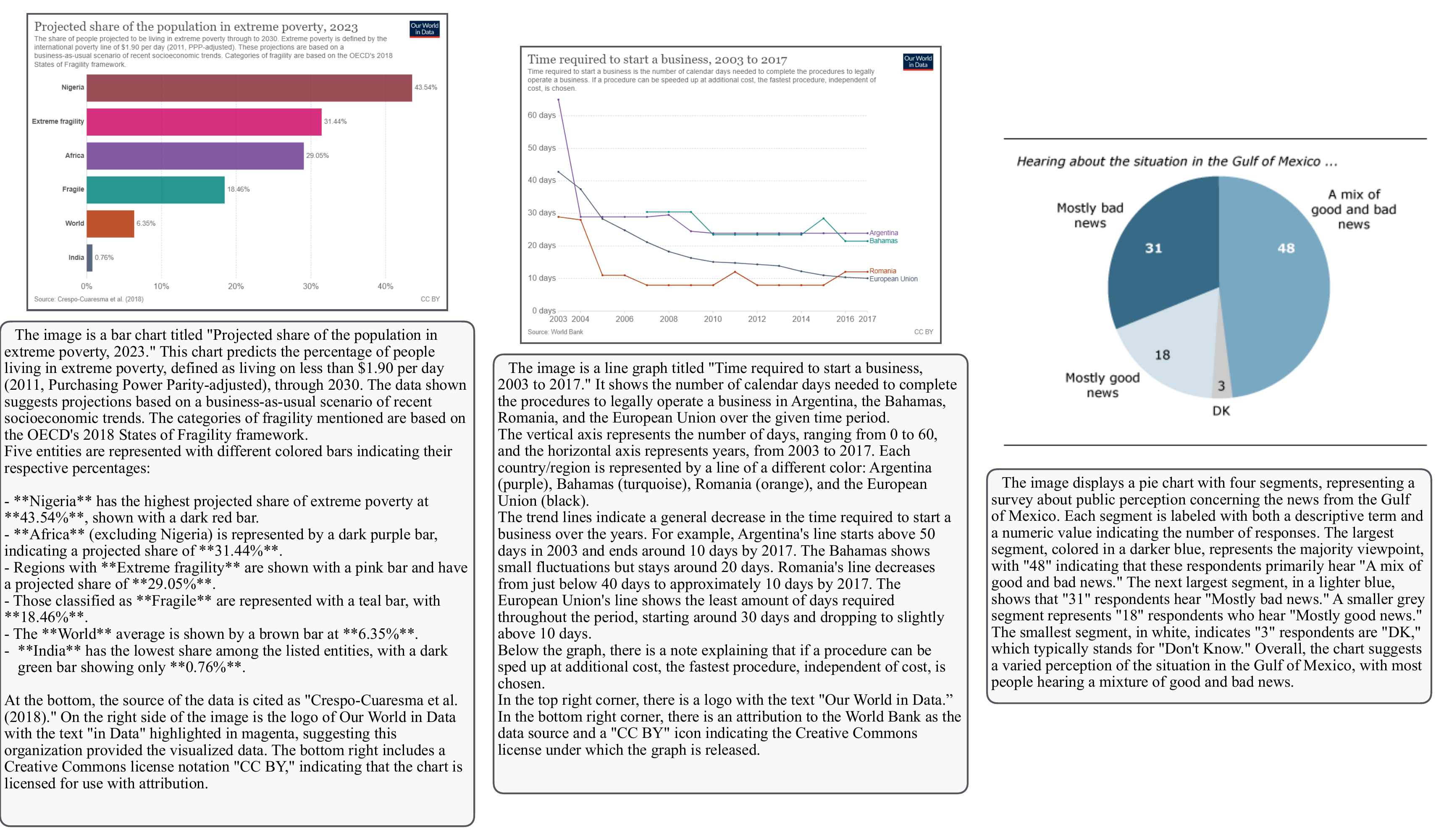}
    \caption{{Examples of ChartQA captions}}
    \label{fig:chart-gpt-caption-example}
\end{figure}
\clearpage

\newpage
\input{tables/main_results}

\input{tables/ablation_full}
\section{Additional Results}
\subsection{Extended Comparison}
We provide the full comparison result against SoTA MLLMs in Table~\ref{tab:main_result}. We additionally include the results from  MLLMs such as GPT-4o~\citep{openai20234v}, Qwen-VL-Max~\citep{wang2024qwen2}, and InternVL-76B~\citep{chen2024far}.

\subsection{Full result on the ablation study of CI category}
Table~\ref{tab:full_ablation} provides the scores of all benchmark for the CI component ablation study. 

\input{tables/training_schedule_ablation}

\subsection{Ablation on training data sampler}
During the SFT stage, MLLMs are typically trained to follow instructions and enhance conversational capabilities. Unlike the usual instruction data, which is often in QA format, \ours{}-118K primarily emphasizes caption data, which aims at facilitating vision-language alignment. In this experiment, we explore various strategies for combining our caption data with instructional data during training. Specifically, we set a proportion of caption data from \ours{}-118K in each training batch, and adjust this ratio at every training step.

We hypothesize that increasing caption data early in training will strengthen alignment, while focusing more on instructional data in later stages will maintain the model's instruction-following proficiency. To explore this, we experiment with four training data samplers:

\begin{compactenum}
    \item \textit{Truncated.} This sampler first samples from \ours{}-118K until all data is used, then shifts to the original downsampled SFT dataset.
    \item \textit{Cosine}. For the $t$-th training step, this sampler returns a batch where $\alpha \text{cosine}(t/T)$ percent is drawn from \ours{}-118K, with $T$ representing the total training steps and $\alpha$ adjusted so all \ours{}-118K data is covered by training’s end.
    \item \textit{Linear.} Similar to the Cosine sampler, but the schedule changes linearly to $1-t/T$
    \item \textit{Uniform.} This sampler uniformly mixes data from \ours{}-118K and the original downsampled SFT dataset in each batch.
\end{compactenum}
We evaluate the impact of these samplers using \ours{}-7B/13B. We train them on the SFT dataset with each sampler and present the results in Table~\ref{tab:alblation_train_schedule}. We find that the uniform sampler consistently outperforms others for both 7B and 13B MLLMs. Unless specified otherwise, all models in our experiments use the uniform sampler for training.

\subsection{Diversity Analysis}
We demonstrate the diversity of the caption for each CI category in Figure~\ref{fig:diversity_visualization}. The inner circle of the plot displays the root verbs of the captions, while the outer circle represents the corresponding direct nouns. We display the top 20 root verbs for each CI class, along with the top 5 nouns associated with each root verb.

\begin{figure*}
    \begin{tabular}{cc}\\
 \includegraphics[width=0.46\linewidth]{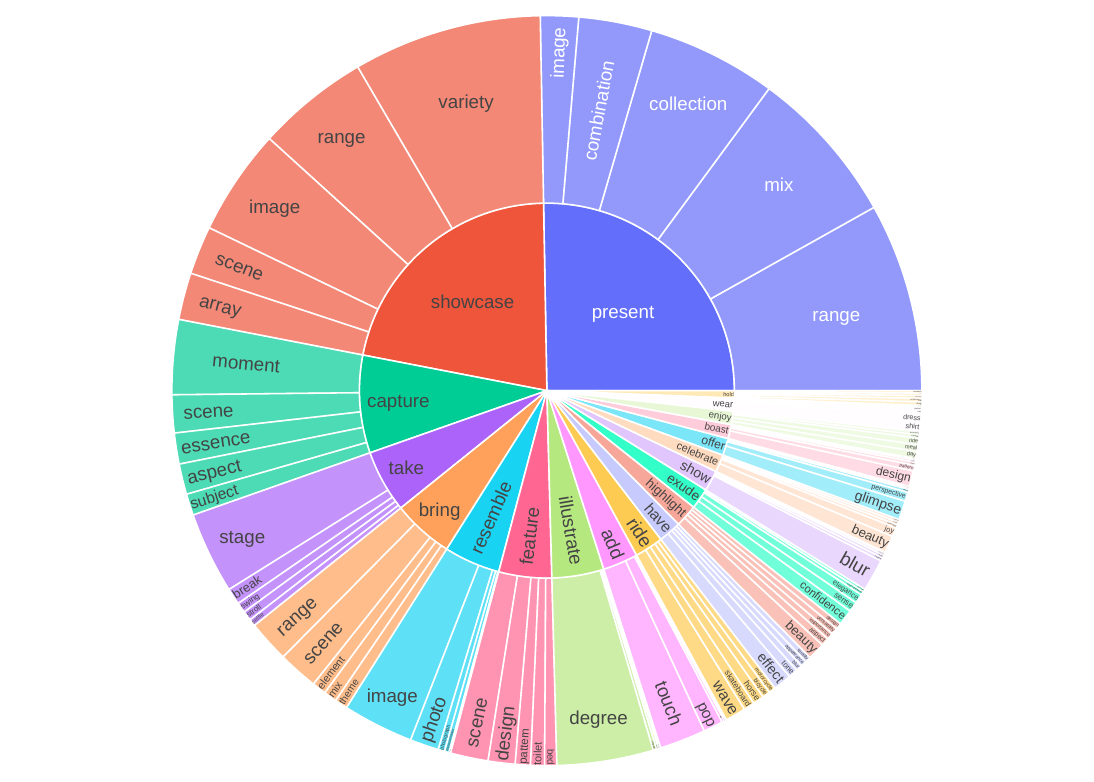} &
 \includegraphics[width=0.46\linewidth]{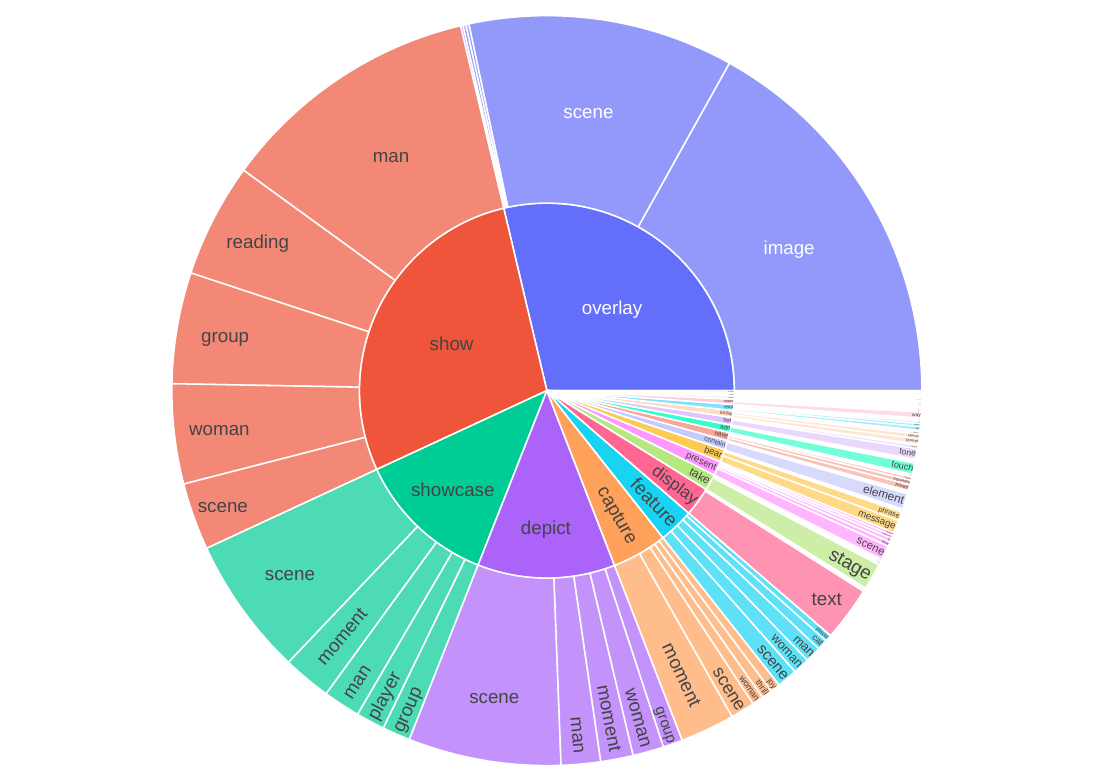} \\
 Collage & Image-Text \\\\
 \includegraphics[width=0.46\linewidth]{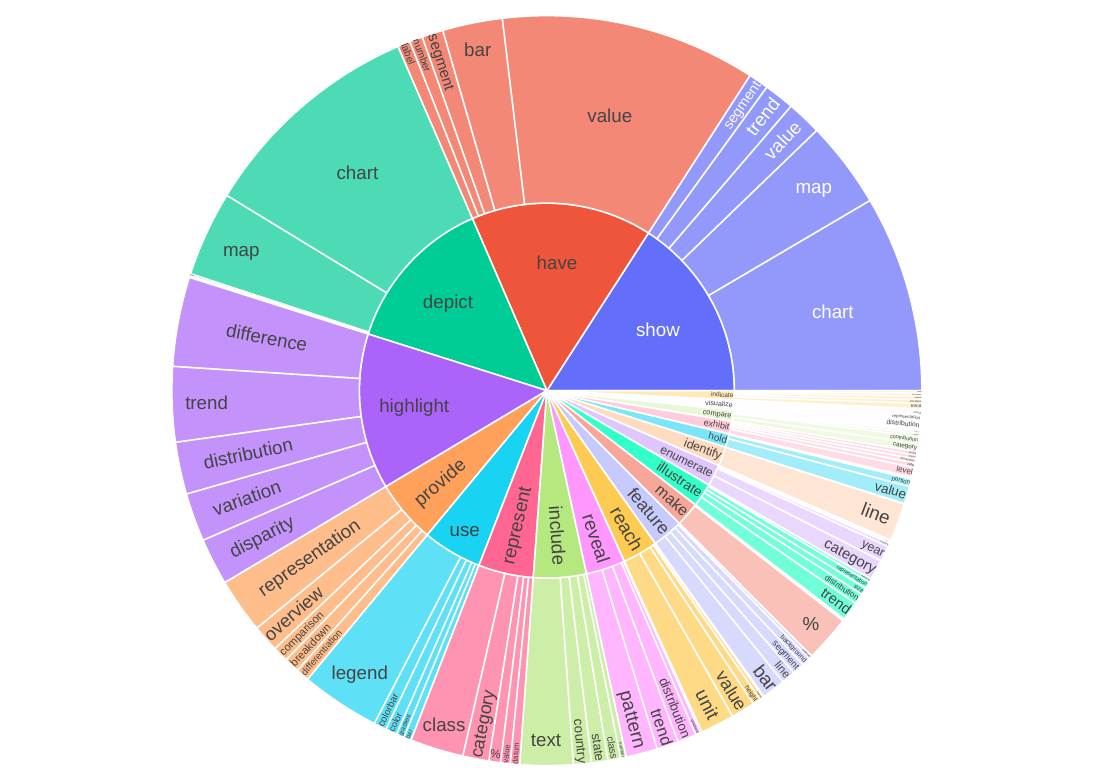} &
 \includegraphics[width=0.46\linewidth]{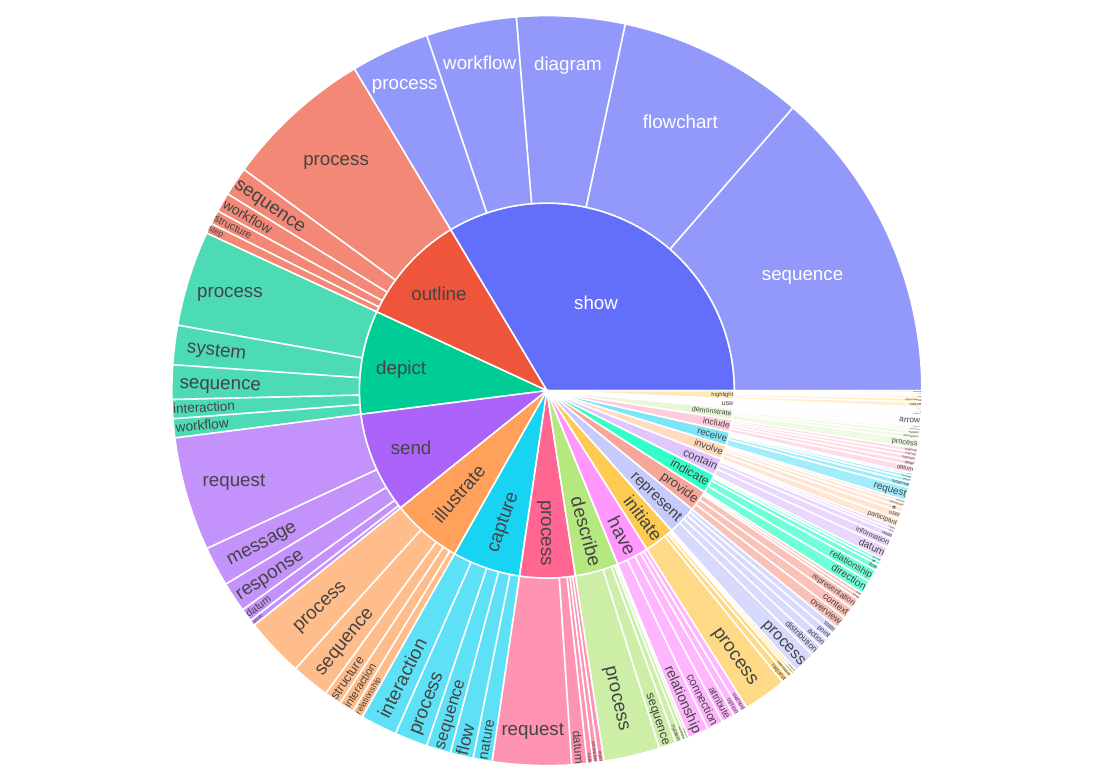} \\
 Chart & Diagram \\\\
 \includegraphics[width=0.46\linewidth]{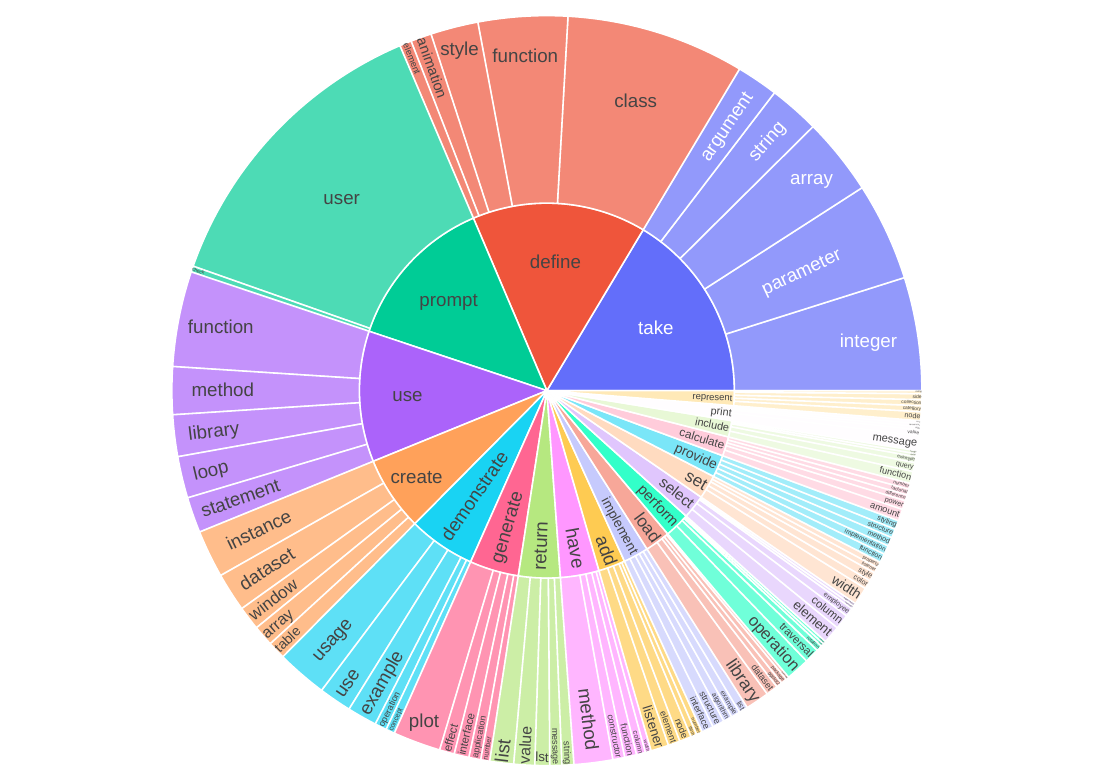} &
 \includegraphics[width=0.46\linewidth]{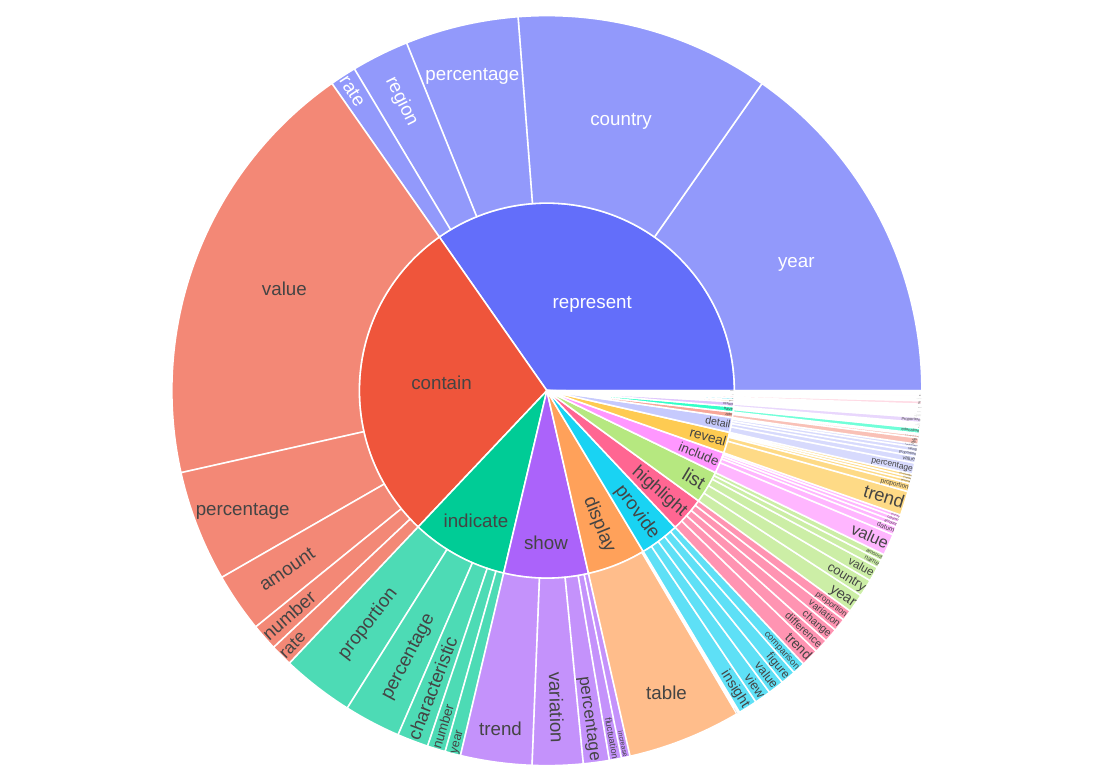} \\
 Code & Table\\
    \end{tabular}
    \caption{\textbf{Diversity analysis of captions for different CI types.}}
    \label{fig:diversity_visualization}
\end{figure*}

\subsection{Examples of Synthesized CI-caption Pairs}
We show examples of the curated image-caption pairs for each CI type in Figures~\ref{fig:collage-image-example}, \ref{fig:image-text-image-example}, \ref{fig:chart-image-example}, \ref{fig:diagram-image-example}, \ref{fig:table-image-example}, \ref{fig:code-image-example}.

\begin{figure*}
     \centering
    \includegraphics[width=0.95\linewidth]{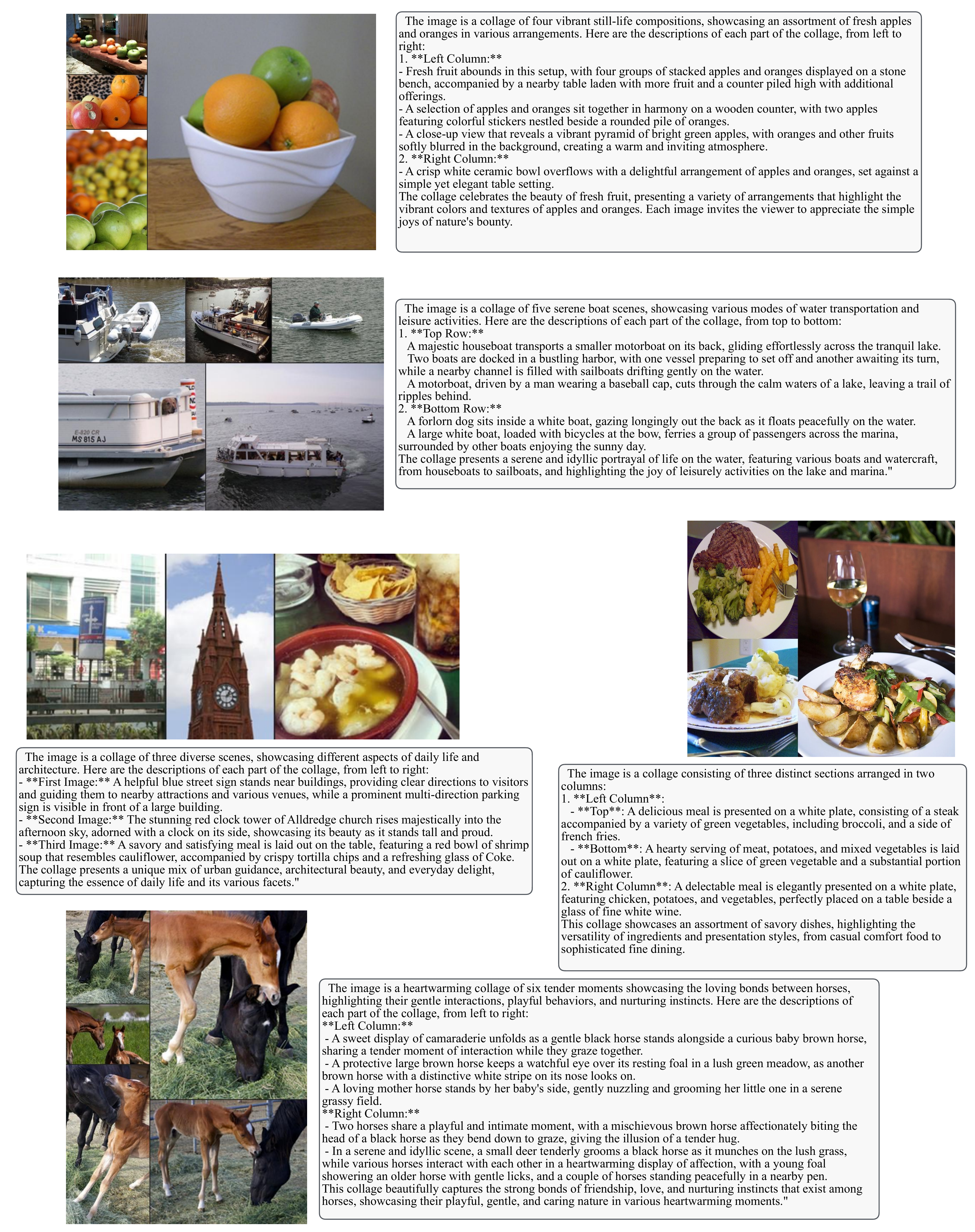}
    \caption{\textbf{Image and caption examples of collage in \ours{}-118K}}
    \label{fig:collage-image-example}
\end{figure*}
\begin{figure*}
    \vspace{16em}
    \centering
    \includegraphics[width=0.95\linewidth]{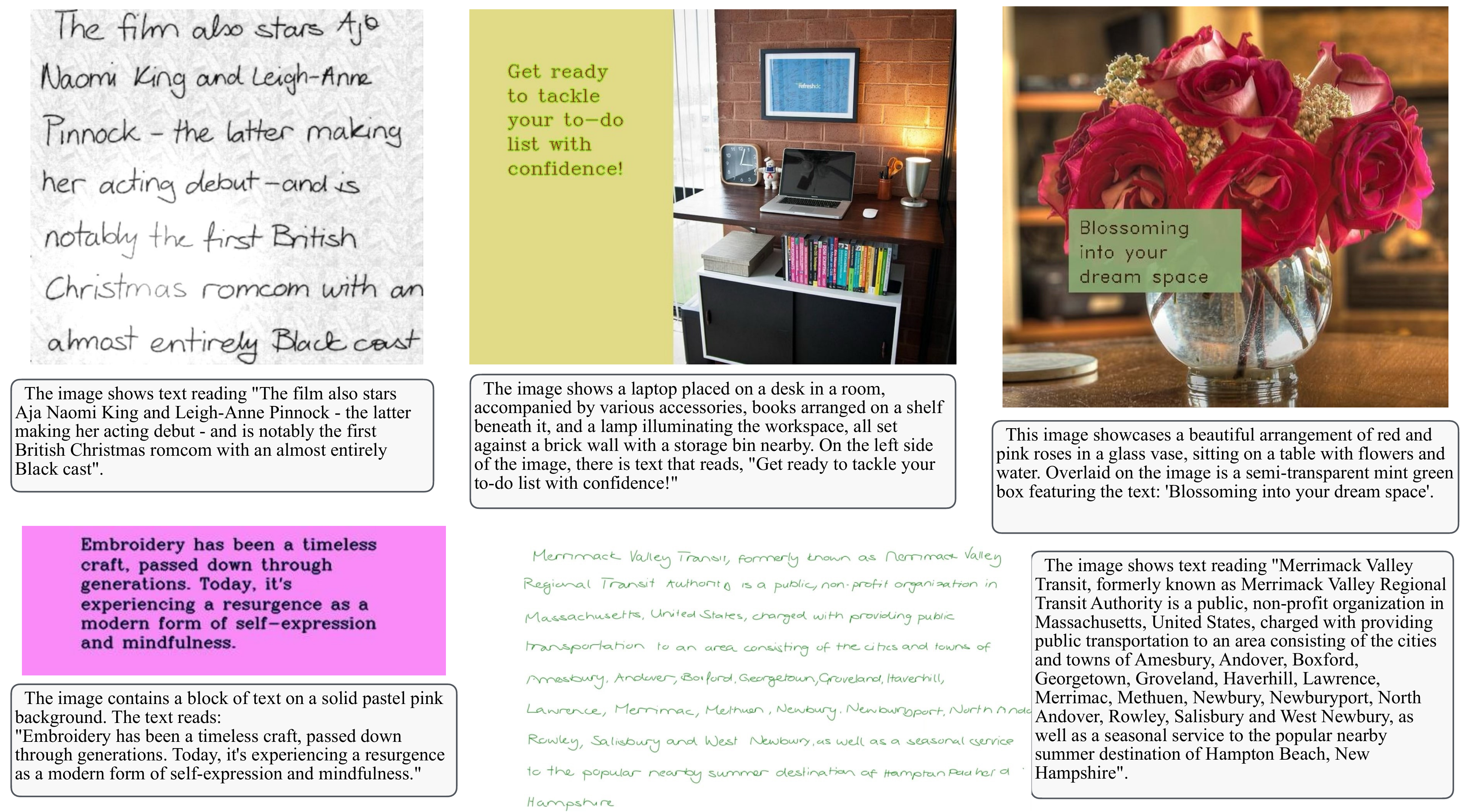}
    \caption{\textbf{Image and caption examples of image-text in \ours{}-118K}}
    \vspace{16em}
    \label{fig:image-text-image-example}
\end{figure*}
\begin{figure*}
     \centering
    \includegraphics[width=0.95\linewidth]{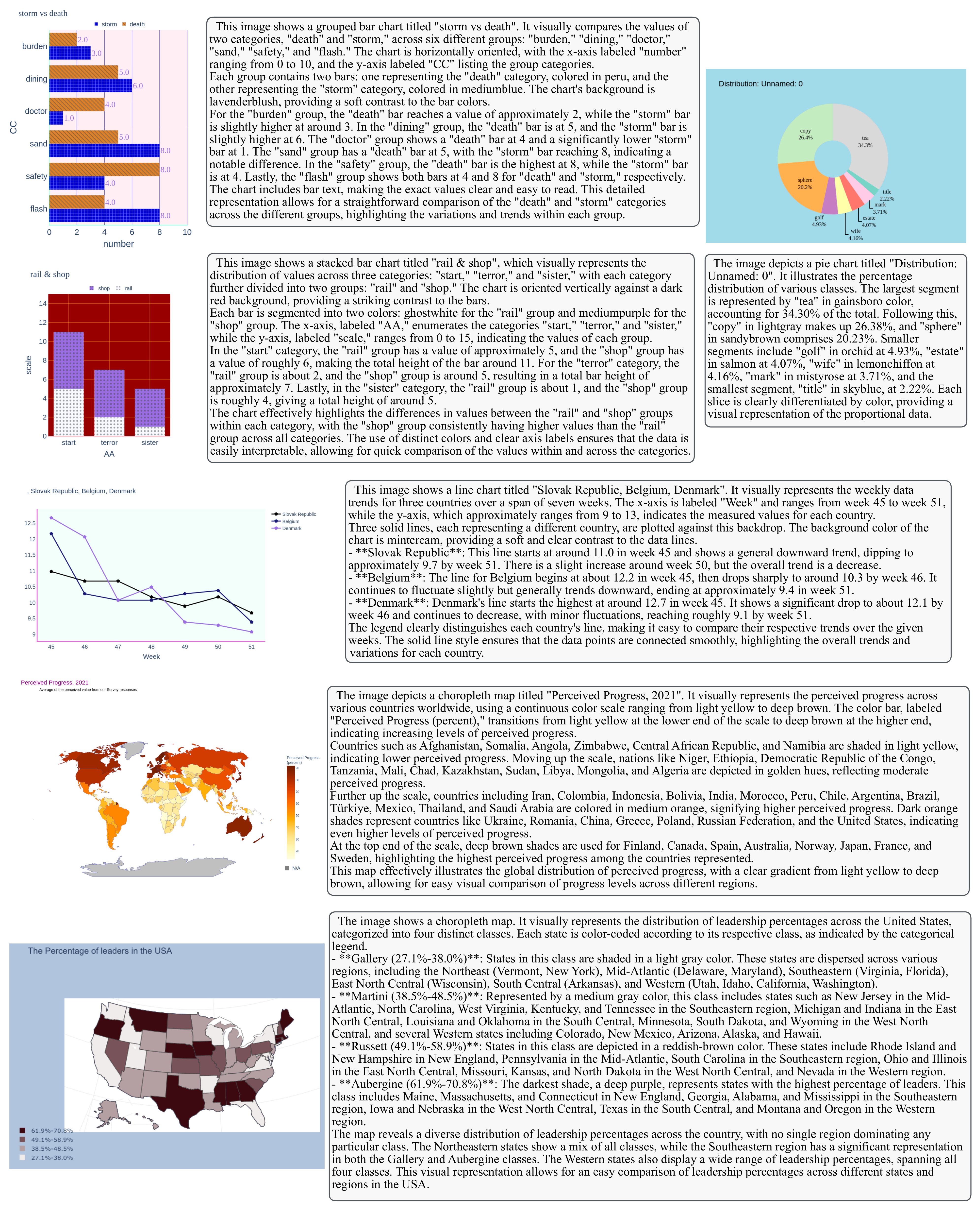}
    \caption{\textbf{Image and caption examples of chart in \ours{}-118K}}
    \label{fig:chart-image-example}
\end{figure*}
\begin{figure*}
    \vspace{4em}
    \centering
    \includegraphics[width=0.95\linewidth]{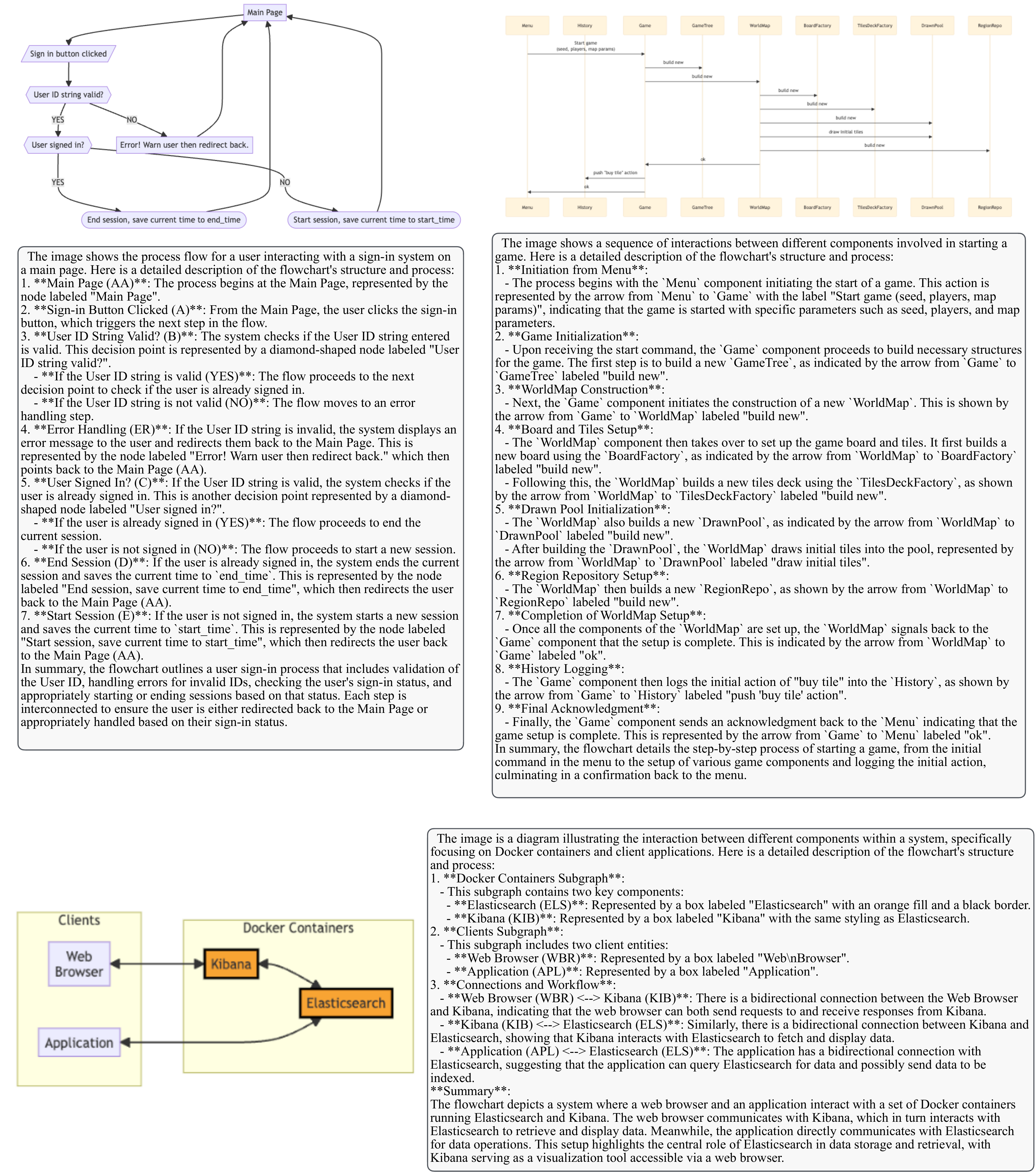}
    \caption{\textbf{Image and caption examples of diagram in \ours{}-118K}}
    \label{fig:diagram-image-example}
    \vspace{4em}
\end{figure*}
\begin{figure*}
    \vspace{4em}
     \centering
    \includegraphics[width=0.95\linewidth]{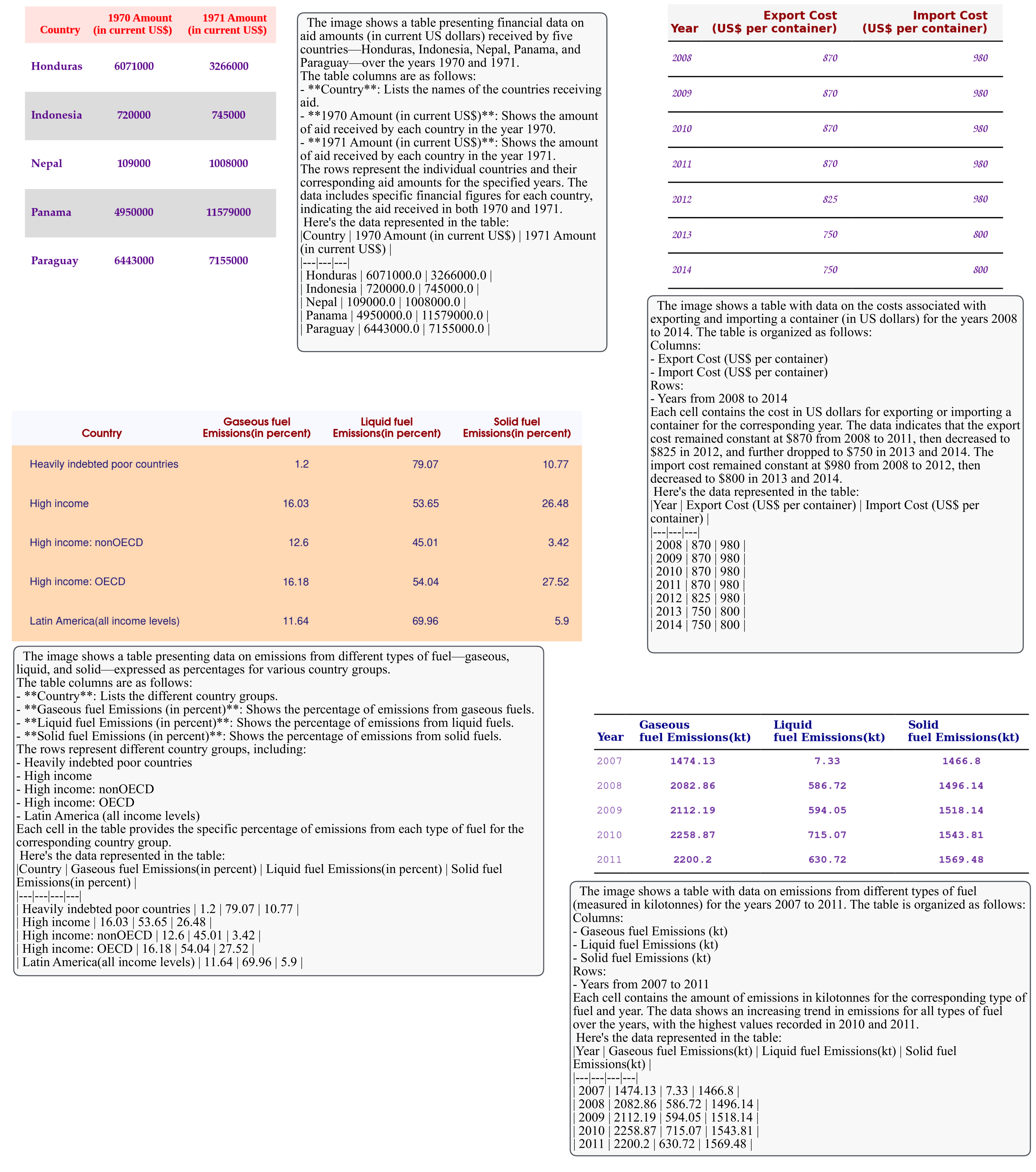}
    \caption{\textbf{Image and caption examples of table in \ours{}-118K}}
    \label{fig:table-image-example}
    \vspace{4em}
\end{figure*}

\begin{figure*}
     \centering
    \includegraphics[width=0.95\linewidth]{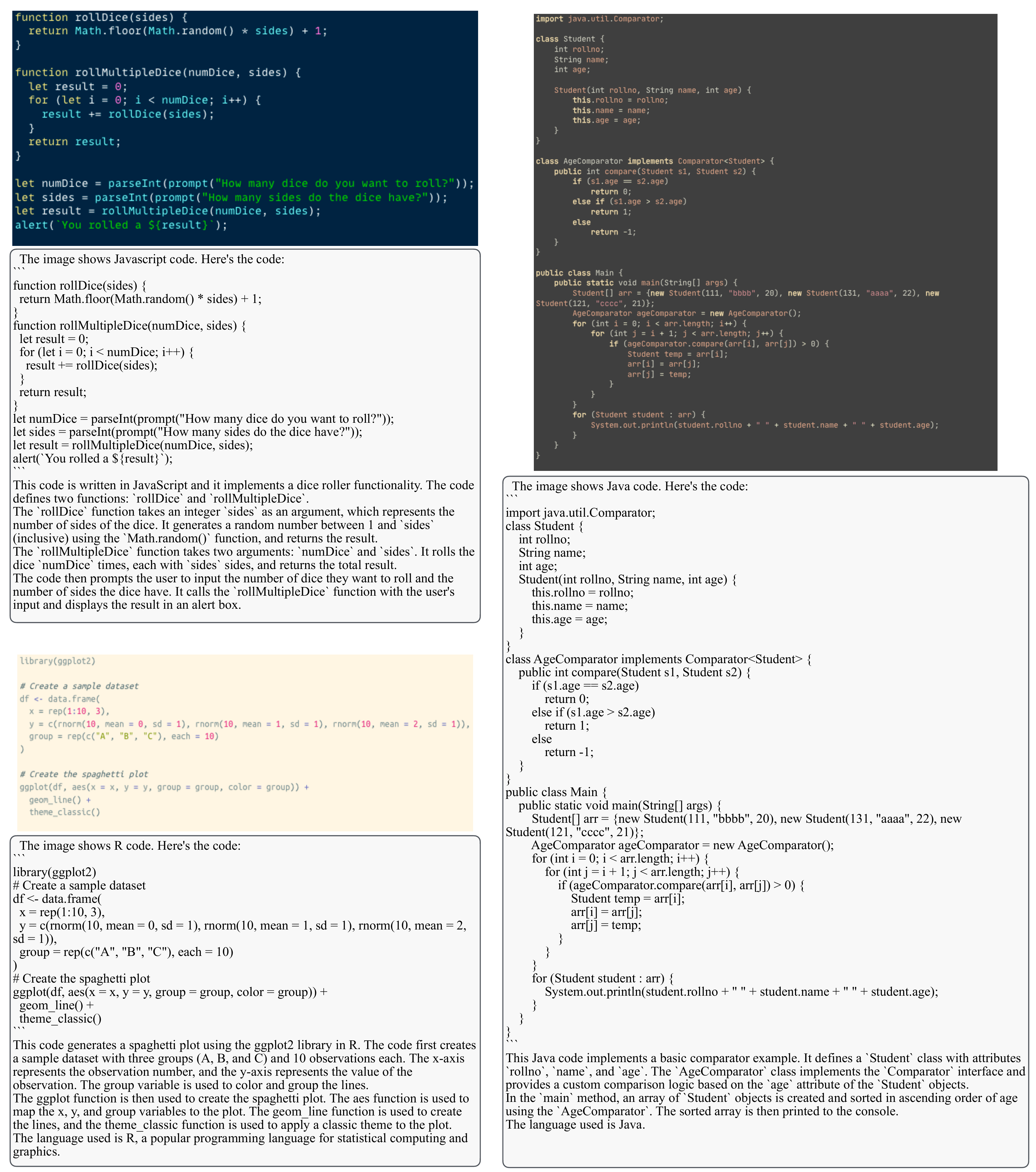}
    \caption{\textbf{Image and caption examples of code in \ours{}-118K}}
    \label{fig:code-image-example}
\end{figure*}

\clearpage
\section{Qualitative Examples}
In this section, we show more qualitative captioning results (Figures~\ref{fig:example-infer-2}, \ref{fig:example-infer-3}, \ref{fig:example-infer-4}, \ref{fig:example-infer-5}) of \ours{}-13B, in comparison with LLaVA-NeXT-Vicuna-13B.
\begin{figure*}[h]
    \centering
    \includegraphics[width=\linewidth]{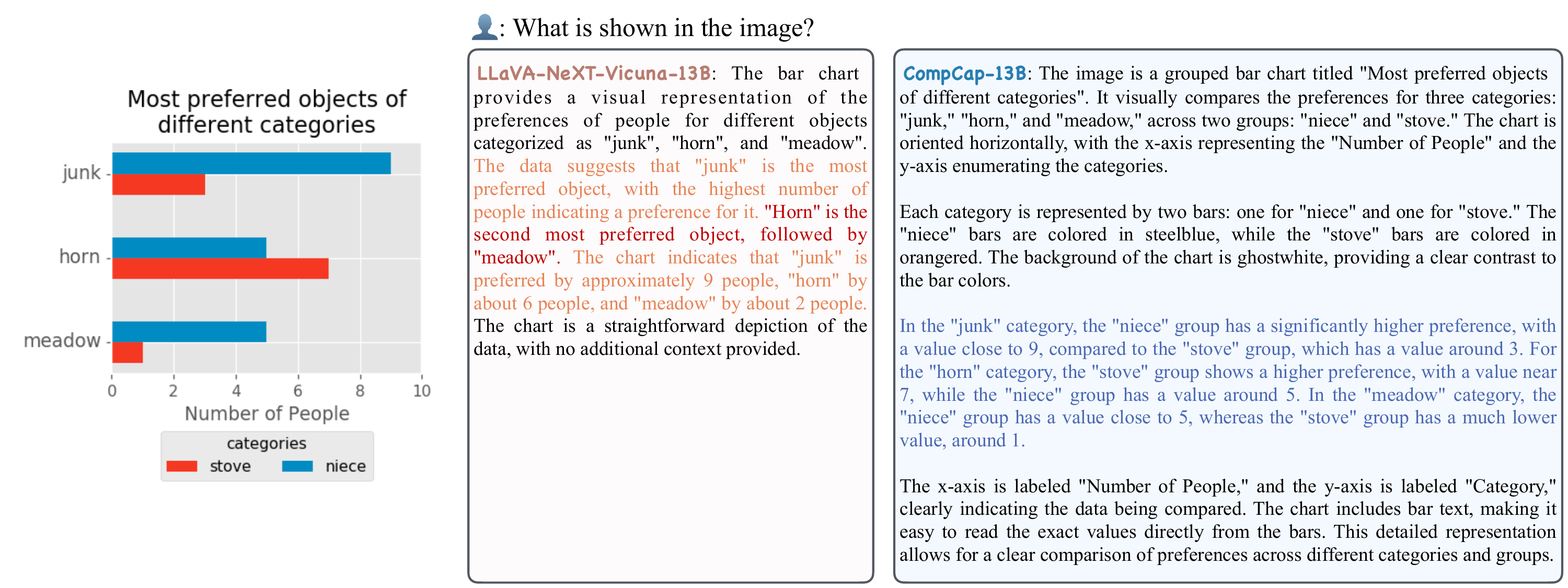}
    \caption{\textbf{More examples of MLLMs on CI captioning (part 1).}}
    \label{fig:example-infer-2}
\end{figure*}
\begin{figure*}[h]
    \centering
    \includegraphics[width=\linewidth]{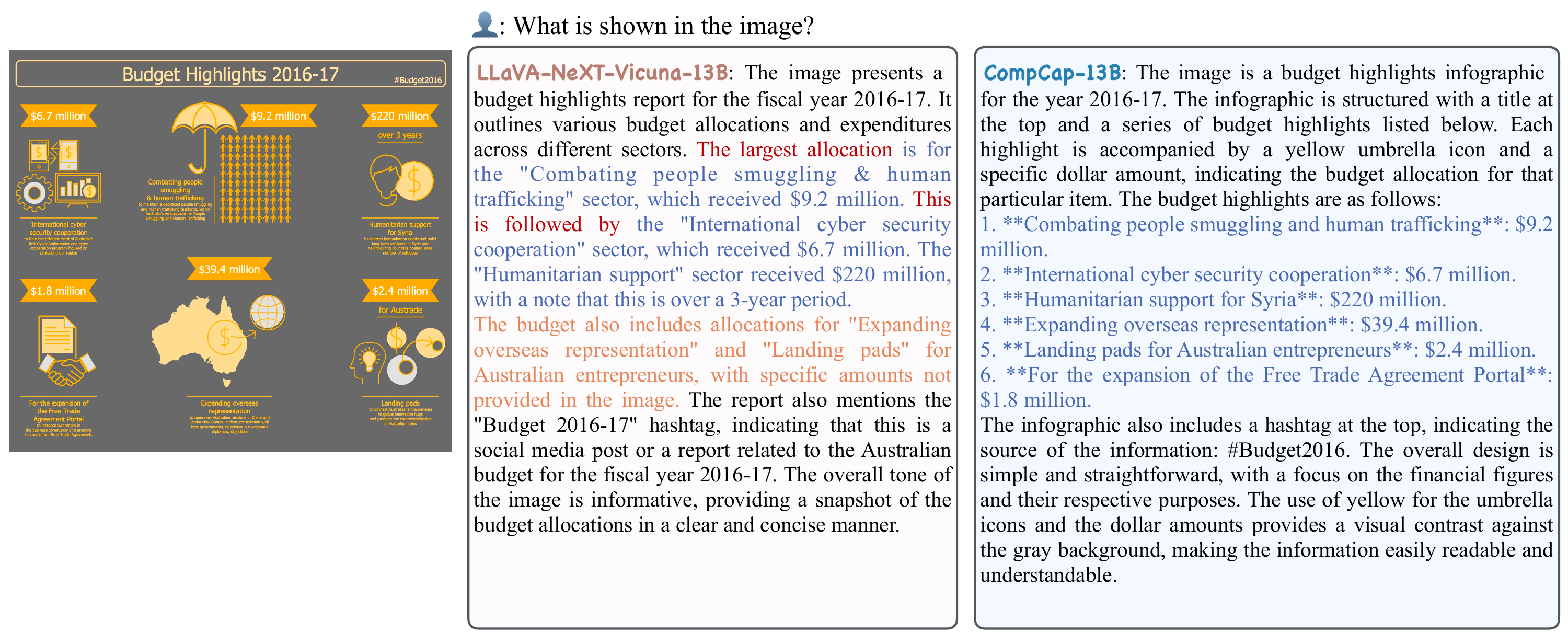}
    \caption{\textbf{More examples of MLLMs on CI captioning (part 2).}}
    \label{fig:example-infer-3}
\end{figure*}
\begin{figure*}[h]
    \centering
    \includegraphics[width=\linewidth]{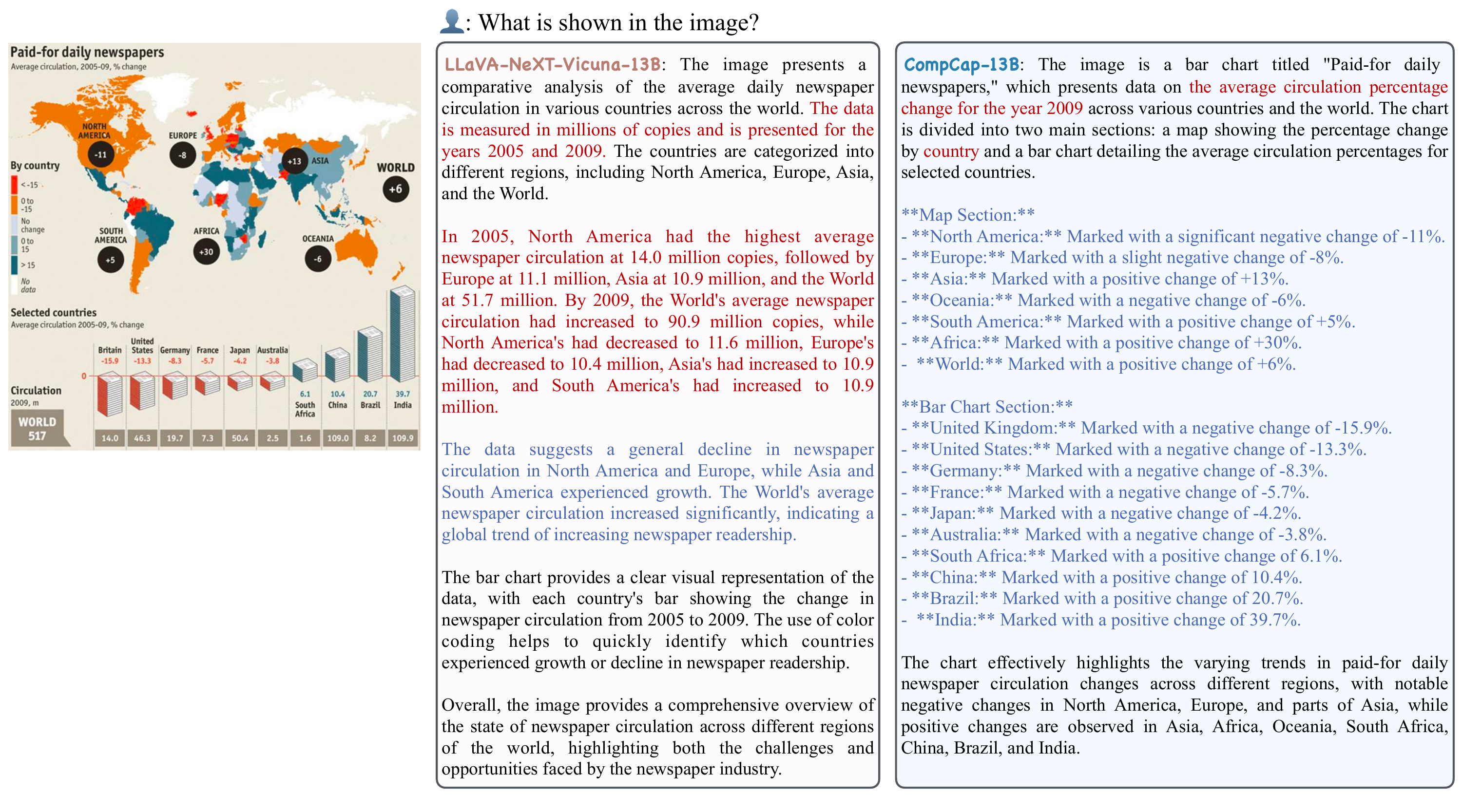}
    \caption{\textbf{More examples of MLLMs on CI captioning (part 3).}}
    \label{fig:example-infer-4}
\end{figure*}
\begin{figure*}[h]
    \centering
    \includegraphics[width=\linewidth]{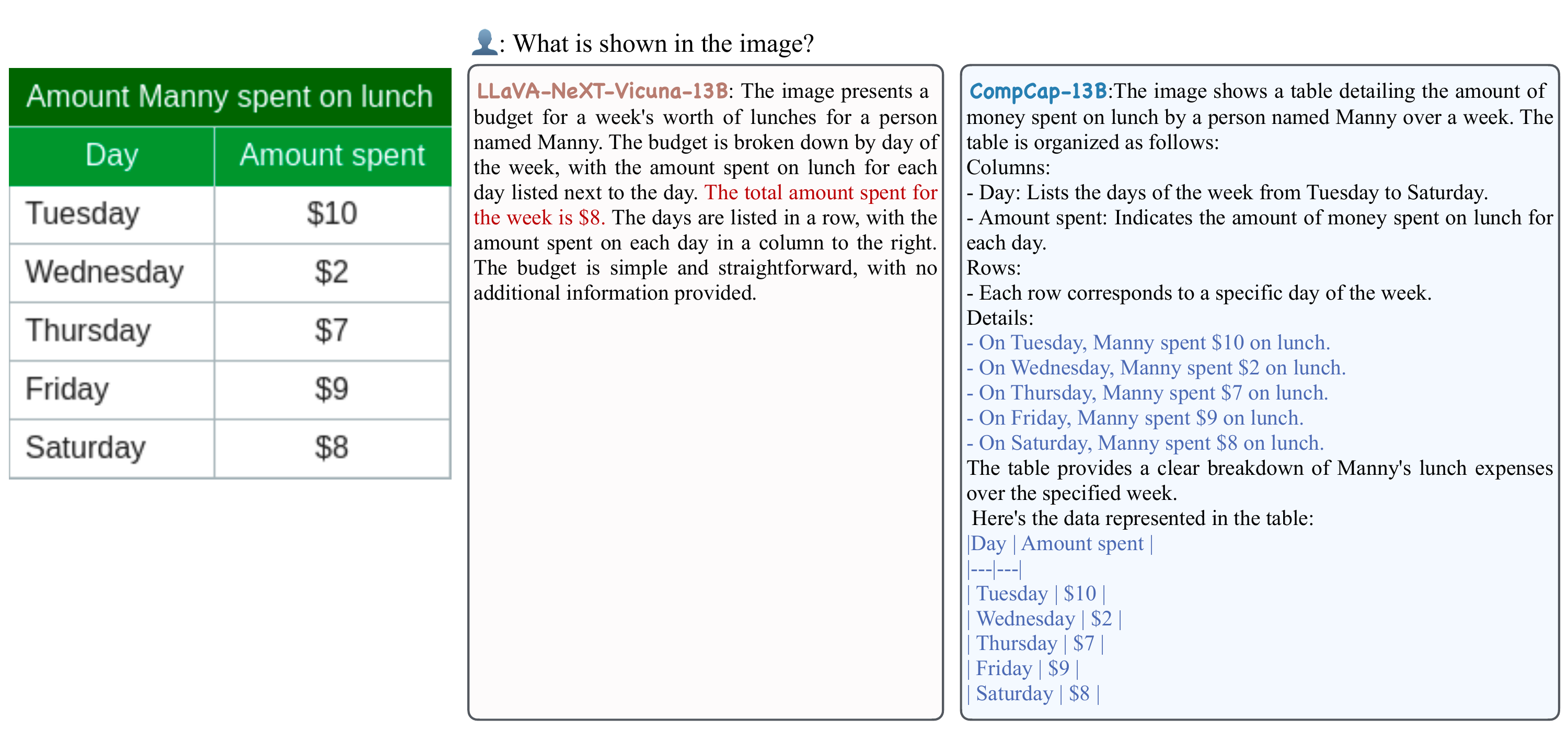}
    \caption{\textbf{More examples of MLLMs on CI captioning (part 4).}}
    \label{fig:example-infer-5}
\end{figure*}

%% file: tables/hw_effect.tex
\begin{table*}[t]
\centering
\captionsetup{justification=centering}
\resizebox{\linewidth}{!}{
\begin{tabular}{ll}
\toprule
\textbf{Augmentation} & \textbf{Applied Effect} \\
\midrule
\emph{Ink Phase}\\
\midrule
InkBleed & Applies random noise along text edges them to mimic fuzzy, bleeding ink when blurred. \\
BleedThrough & Combines ink bleed with Gaussian blur to recreate an effect where ink seeps through the paper. \\
\midrule
\emph{Paper Phase}\\
\midrule
PaperFactory & Replaces the background with a randomly chosen texture, resized or tiled to cover the entire image. \\

Tessellation & Applies a repeating geometric pattern that interlocks seamlessly, giving a structured, mosaic-like texture. \\

NoiseTexturize & Adds a random noise pattern in varying scales to create a realistic paper texture. \\

BrightnessTexturize & Introduces random brightness variations to mimic subtle textural differences in paper. \\
\midrule
\multicolumn{2}{l}{\emph{Combined Phase}}\\
\midrule
DirtyDrum & Simulates a dirty drum effect by adding vertical and horizontal noise lines across the image. \\

DirtyRoller & Recreates the streaking effect of old or dirty document rollers in scanners. \\

SubtleNoise & Adds slight, uneven noise to replicate minor lighting inconsistencies seen in scans of solid colors. \\

BadPhotoCopy & Adds a grainy, noisy overlay to mimic the quality of a worn-out photocopier. \\

ShadowCast & Casts shadows on the paper to simulate natural shadows from scanning or photocopying. \\

ReflectedLight & Draws bright ellipses on the paper to recreate the effect of light reflecting on the surface. \\
\bottomrule
\end{tabular}
}
\caption{Augraphy augmentations for pure text. Effects are applied on the image sequentially according to the row order.}
\label{table:augmentors}
\end{table*}

%% file: tables/composition_demo_benchmark.tex
\begin{table}[h]
\centering
    \captionsetup{justification=centering}
\begin{tabular}{lcc}\toprule
Benchmark & Covered CI Types &\#Samples\\\midrule
ChartQA~\citep{chartqa} & Chart & 200\\
DocVQA~\citep{docvqa} & Document & 100\\
InfoVQA~\citep{infovqa} & Infographic & 200\\
MapQA~\citep{chang2022mapqa} & Chart (Map) & 50\\
MME~\citep{yin2023survey} & Collage/Code & 60\\
OCRBench~\citep{liu2023hidden} & Document/Infographic &100\\
MMVet~\citep{yu2023mm} & Collage/Chart/Diagram/Table & 40\\
MMBench~\citep{liu2025mmbench} & Collage/Chart/Diagram/Table & 250\\
\bottomrule
    \end{tabular}
    \caption{{Dataset sources for the curated CI benchmark.}}
    \label{tab:demo-dataset-composition}
\end{table}

%% file: tables/training_hyperparams.tex
\begin{table}[h]
\centering
\captionsetup{justification=centering}
    \begin{tabular}{lccc}\toprule
 & \ours{}-4B & \ours{}-7B & \ours{}-13B \\\midrule
\#vision tokens  & 128 & 2880 & 2880\\
vision encoder & SigLip~\citep{zhai2023sigmoid} & CLIP & CLIP \\
image aspect ratio & anyres & anyres & anyres \\
batch size & 64 & 128 & 128 \\
lr & 2e-5 & 2e-5 & 2e-5\\
lr schedule & cosine & cosine & cosine \\
weight decay & 0 & 0 & 0 \\
optimizer & AdamW & AdamW & AdamW \\
\#epochs & 1 & 1 & 1 \\\bottomrule
    \end{tabular}
    \caption{{Hyperparameters for training \ours{} series}}
    \label{tab:hyperparams}
\end{table}

%% file: tables/main_results.tex
\begin{table*}[t]
    \centering\captionsetup{justification=centering}
    \resizebox{\linewidth}{!}{
    \begin{tabular}{lccccccccccccc}
\toprule
Model &
    PT/SFT \#Data&
    \rotatebox{90}{\small{SEEDBench} 
    {$\mathbin{\Diamond}$}} &
    \rotatebox{90}{\small{TextVQA} 
    {$\mathbin{\Diamond}$}} &
    \rotatebox{90}{\small{MMBench} $\mathbin{\Diamond}$\tiny{$\mathbin{\blacklozenge}$}} &
    \rotatebox{90}{\small{MME (norm)} $\mathbin{\Diamond}$\tiny{$\mathbin{\blacklozenge}$}} &
    \rotatebox{90}{\small{LLaVABench} $\mathbin{\Diamond}$\tiny{$\mathbin{\blacklozenge}$}} &
    \rotatebox{90}{\small{MathVista}
    $\mathbin{\blacklozenge}$\tiny{$\mathbin{\Diamond}$}} &
    \rotatebox{90}{\small{OCRBench}
    $\mathbin{\blacklozenge}$\tiny{$\mathbin{\Diamond}$}} &
    \rotatebox{90}{\small{ChartQA} 
    {$\mathbin{\blacklozenge}$}} &
    \rotatebox{90}{\small{DocVQA} 
    {$\mathbin{\blacklozenge}$}} &
    \rotatebox{90}{\small{InfoVQA} 
    {$\mathbin{\blacklozenge}$}} &
    \rotatebox{90}{\small{WebSRC} 
    {$\mathbin{\blacklozenge}$}} &
    \rotatebox{90}{\small{Avg.}} \\
\midrule
\rowcolor{lightgray}
\multicolumn{14}{l}{\emph{SoTA MLLMs}}\\
\midrule
GPT-4o~\citep{openai20234v} & UNK./UNK. & 77.1 & - & 81.8 & 83.2 & 102.0 & 63.8 & 73.6 & 85.7 & 92.8 & - & - & -\\
Qwen-VL-Max~\citep{wang2024qwen2} & UNK./UNK. & 77.9 & 85.5 & - & 88.7 & 74.9 & 70.5 & 85.5 & 88.3 & 96.5 & 84.5 & - & -\\
InternVL-76B~\citep{chen2024far} & UNK./UNK. & 77.6 & 84.4 & - & 86.3 & 96.3 & 65.6 & 84.2 & 88.4 & 94.1 & 82.0 & - & -\\
\midrule
\rowcolor{lightgray}
\multicolumn{14}{l}{\emph{3B - 4B MLLMs}}\\
\midrule
MM1-3B~\citep{mckinzie2024mm1} & 2.9B/1.45M & 68.8 & \underline{71.9} & 67.8 & 62.9 & 72.1 & {32.0} & - & - & - & - & - & -\\
VILA-1.5-3B~\citep{lin2024vila} & 32.8M/5.9M & 68.0 & 55.6 & 62.4 & 58.2 & 65.5 & 30.6 & 43.7 & 52.9 & - & - & - & -\\
Phi-3-vision~\citep{abdin2024phi} & 5B/$>$8.3M & 70.9 & 63.6 & 74.2 & 55.2 & \textbf{82.2} & \textbf{45.1} & \textbf{63.7} & \textbf{81.8} & \textbf{84.3} & \textbf{50.0} & \textbf{65.2} & \textbf{66.9}\\
xGen-MM-inst.-4B~\citep{xue2024xgen} & $>$25M/UNK. & \textbf{71.8} & \textbf{72.0} & \underline{76.0} & \underline{64.1} & 75.7 & \underline{39.5} & \underline{54.8} & \underline{59.5} & \underline{61.1} & \underline{31.3} & \underline{55.8} & \underline{60.2}\\
\rowcolor{isabelline}
xGen-MM-inst.-4B$^*$~\citep{xue2024xgen} & \cellcolor{lightblue}$>$25M/1M & 71.3 & 67.7 & 75.5 & 64.0 & 78.2 & 32.6 & 51.6 & 54.8 & 55.2 & 27.6 & 50.6 & 57.2\\
\rowcolor{isabelline}
CompCap-4B & \cellcolor{lightblue}$>$25M/1M & \underline{71.6} & 67.9 & \textbf{76.2} & \textbf{64.7} & \underline{81.0} & {35.0} & {52.7} & 57.4 & 58.1 & 27.9 & \underline{55.8} & {58.9}\\
\midrule
\rowcolor{lightgray}
\multicolumn{14}{l}{\emph{7B - 8B MLLMs}}\\
\midrule
VILA-1.5-8B & 32.8M/5.9M & 65.0 & 60.2 & 68.6 & 60.7 & 71.7 & 37.3 & 43.8 & 50.9 & - & - & - & -\\
ShareGPT4V-7B~\citep{chen2023sharegpt4v} & 1.2M/665K & 69.3 & 58.3 & 68.8 & \textbf{68.4} & 66.9 & 26.5 & 37.1 & 21.3 & 14.4 & 14.7 & 36.4 & 43.8\\
Qwen-VL-chat-7B~\citep{wang2024qwen2} & UNK./UNK. & 64.8 & 60.7 & 60.6 & 66.4 & 67.7 & 34.9 & 48.8 & 49.8 & 62.6 & 29.7 & 53.6 & 54.5\\
Cambrian-8B~\cite{tong2024cambrian} & 1.2M/7M & \textbf{73.3} & \textbf{72.6} & \textbf{75.9} & 64.4 & 71.0 & \textbf{47.0} & \textbf{61.4} & \textbf{72.6} & \textbf{77.8} & \underline{40.1} & \underline{68.9} & \textbf{65.9}\\
\rowcolor{isabelline}
LLaVA-NeXT-Vicuna-7B~\cite{liu2024llavanext} & \cellcolor{lightblue}558K/779K & \underline{71.2} & 65.2 & 67.6 & 66.3 & \underline{72.4} & 39.6 & 55.1 & 63.5 & 76.5 & 39.2 & 70.4 & 62.5\\
\rowcolor{isabelline}
CompCap-7B & \cellcolor{lightblue}558K/779K & 70.5 & \underline{65.6} & \underline{68.9} & \underline{67.5} & \textbf{75.5} & \underline{41.7} & \underline{58.5} & \underline{68.9} & \underline{77.6} & \textbf{40.8} & \textbf{73.7} & \underline{64.5}\\
\midrule
\rowcolor{lightgray}
\multicolumn{14}{l}{\emph{13B MLLMs}}\\
\midrule
VILA-1.5-13B~\citep{lin2024vila} & 32.8M/5.9M & 72.7 & 61.2 & 74.3 & 61.4 & 73.4 & 42.5 & 46.0 & \underline{74.6} & - & - & - & -\\
ShareGPT4V-13B~\citep{chen2023sharegpt4v} & 1.2M/665K & 70.6 & 52.7 & 69.0 & 66.2 & 69.1 & 29.3 & 39.8 & 24.6 & 14.5 & 17.2 & 39.4 & 44.8\\
OmChat-v2.0-13B~\citep{zhao2024omchat} & $>$6.5B/20M & \textbf{75.2} & \textbf{79.8} & \textbf{82.1} & \textbf{76.1} & 66.1 & \textbf{57.1} & \textbf{72.8} & 79.9 & \textbf{88.7} & \textbf{58.8} & \textbf{88.2} & \textbf{75.0}\\
Cambrian-13B~\citep{tong2024cambrian} & 1.2M/7M & \underline{73.2} & \underline{72.8} & \underline{75.7} & 66.8 & 76.1 & \underline{47.4} & 61.0 & 73.8 & 76.8 & 44.6 & 70.7 & 67.2\\
\rowcolor{isabelline}
LLaVA-NeXT-Vicuna-13B~\citep{liu2024llavanext} & \cellcolor{lightblue}558K/779K & 71.9 & 67.6 & 68.9 & 68.8 & \underline{77.1} & 42.4 & 57.7 & 68.5 & 79.9 & 43.8 & 75.3 & 65.6\\
\rowcolor{isabelline}
CompCap-13B & \cellcolor{lightblue}558K/779K & 72.2 & 67.8 & 70.8 & \underline{71.4} & \textbf{83.4} & 45.0 & \underline{61.4} & 73.9 & \underline{81.1} & \underline{47.0} & \underline{79.3} & \underline{68.5} \\
\bottomrule
\end{tabular}
}
\caption{{Full comparison on MLLM benchmarks.}}
\vspace{-1em}
\label{tab:main_result}
\end{table*}

%% file: tables/ablation_full.tex
\begin{table}[!h]
    \centering\captionsetup{justification=centering}
    \small
    \begin{tabular}{lcccccccccccc}
\toprule
Component &
    \rotatebox{90}{\small{SEEDBench} 
    {$\mathbin{\Diamond}$}} &
    \rotatebox{90}{\small{TextVQA} 
    {$\mathbin{\Diamond}$}} &
    \rotatebox{90}{\small{MMBench} $\mathbin{\Diamond}$\tiny{$\mathbin{\blacklozenge}$}} &
    \rotatebox{90}{\small{MME (norm)} $\mathbin{\Diamond}$\tiny{$\mathbin{\blacklozenge}$}} &
    \rotatebox{90}{\small{LLaVABench} $\mathbin{\Diamond}$\tiny{$\mathbin{\blacklozenge}$}} &
    \rotatebox{90}{\small{MathVista}
    $\mathbin{\blacklozenge}$\tiny{$\mathbin{\Diamond}$}} &
    \rotatebox{90}{\small{OCRBench}
    $\mathbin{\blacklozenge}$\tiny{$\mathbin{\Diamond}$}} &
    \rotatebox{90}{\small{ChartQA} 
    {$\mathbin{\blacklozenge}$}} &
    \rotatebox{90}{\small{DocVQA} 
    {$\mathbin{\blacklozenge}$}} &
    \rotatebox{90}{\small{InfoVQA} 
    {$\mathbin{\blacklozenge}$}} &
    \rotatebox{90}{\small{WebSRC} 
    {$\mathbin{\blacklozenge}$}} &
    \rotatebox{90}{\small{Avg.}} \\
\midrule
Baseline & 71.9 & 67.7 & 68.9 & 68.8 & 77.1 & 42.4 & 57.7 & 68.5 & 79.9 & 43.8 & 75.3 & 65.6 \\
\midrule
+ \faThLarge~Collage & 72.1 & 67.3 & 69.9 & 69.1 & 78.4 & 43.2 & 58.8 & 70.9 & 80.4 & 45.3 & 75.5 & 66.4 \\
+ \faCode~Code  & 72.3 & 67.8 & 70.2 & 69.3 & 76.4 & 43.6 & 58.7 & 71.1 & 80.6 & 46.1 & 76.1 & 66.6 \\
+ \faTable~Table & 72.3 & 67.8 & 70.3 & 69.4 & 78.7 & 43.6 & 59.1 & 72.0 & 80.5 & 46.1 & 76.6 & 67.0\\
+ \faSitemap~Diagram & 72.2 & 67.6 & 70.7 & 69.9 & 80.4 & 43.6 & 58.3 & 72.9 & 81.2 & 46.9 & 77.6 & 67.4 \\
+ \faChartBar~Chart & 72.4 & 67.5 & 70.5 & 70.0 & 84.0 & 46.8 & 58.6 & 73.1 & 81.1 & 46.1 & 77.6 & 68.0 \\
+ \faNewspaper~Image-Text & 72.2 & 67.8 & 70.8 & 71.4 & 83.4 & 45.0 & 61.4 & 73.9 & 81.1 & 47.0 & 79.3 & 68.5 \\
\bottomrule
    \end{tabular}
    \caption{{Full benchmark result of the ablation study on each CI category.}}
    \vspace{-2em}
    \label{tab:full_ablation}
\end{table}

%% file: tables/training_schedule_ablation.tex
\begin{table}[h]
    \centering
    \captionsetup{justification=centering}
    \small
    \begin{tabular}{cccccccccccccc}
\toprule
Training schedule &
    \rotatebox{90}{\small{SEEDBench} 
    {$\mathbin{\Diamond}$}} &
    \rotatebox{90}{\small{TextVQA} 
    {$\mathbin{\Diamond}$}} &
    \rotatebox{90}{\small{MMBench} $\mathbin{\Diamond}$\tiny{$\mathbin{\blacklozenge}$}} &
    \rotatebox{90}{\small{MME (norm)} $\mathbin{\Diamond}$\tiny{$\mathbin{\blacklozenge}$}} &
    \rotatebox{90}{\small{LLaVABench} $\mathbin{\Diamond}$\tiny{$\mathbin{\blacklozenge}$}} &
    \rotatebox{90}{\small{MathVista}
    $\mathbin{\blacklozenge}$\tiny{$\mathbin{\Diamond}$}} &
    \rotatebox{90}{\small{OCRBench}
    $\mathbin{\blacklozenge}$\tiny{$\mathbin{\Diamond}$}} &
    \rotatebox{90}{\small{ChartQA} 
    {$\mathbin{\blacklozenge}$}} &
    \rotatebox{90}{\small{DocVQA} 
    {$\mathbin{\blacklozenge}$}} &
    \rotatebox{90}{\small{InfoVQA} 
    {$\mathbin{\blacklozenge}$}} &
    \rotatebox{90}{\small{WebSRC} 
    {$\mathbin{\blacklozenge}$}} &
    \rotatebox{90}{\small{Avg.}} \\   
\midrule

\rowcolor{lightgray}
\multicolumn{13}{l}{\emph{\ours{}-7B}}\\\midrule
\includegraphics[width=0.055\textwidth]{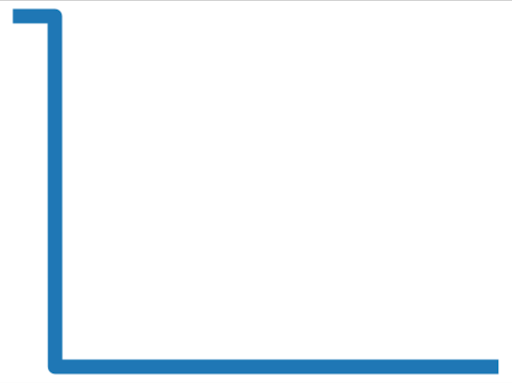}
& \textbf{70.8} & 65.4 & 67.0 & 65.5 & 73.2 & 40.4 & 54.4 & 67.0 & \textbf{78.0} & 39.2 & 72.1 & 63.0 \\ (truncated)\\\midrule
\includegraphics[width=0.055\textwidth]{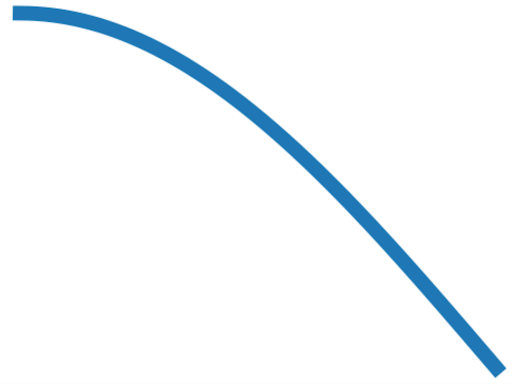}
& \textbf{70.8} & 65.3 & 67.9 & 65.8 & 71.1 & 40.2 & 55.2 & 67.4 & 77.3 & \textbf{40.8} & 73.0 & 63.2 \\ (cosine)\\\midrule
\includegraphics[width=0.055\textwidth]{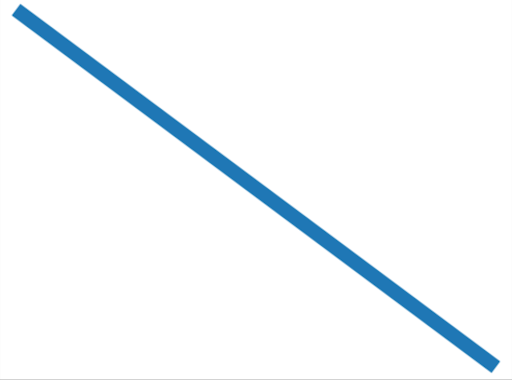}
& 70.6 & \textbf{65.8} & 67.6 & 67.4 & 73.9 & {40.8} & 54.7 & 67.4 & 77.1 & 39.9 & 71.8 & 63.4 \\ (linear)\\\midrule
\includegraphics[width=0.055\textwidth]{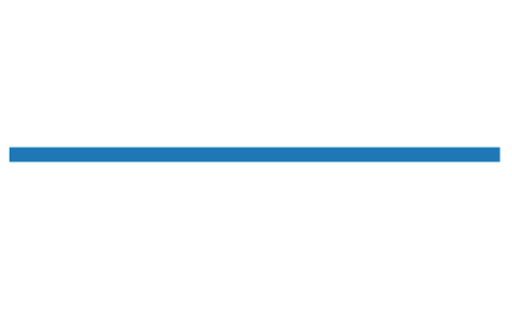}
& 70.5 & 65.6 & \textbf{68.9} & \textbf{67.5} & \textbf{75.5} & \textbf{41.7} & \textbf{58.5} & \textbf{68.9} & {77.6} & \textbf{40.8} & \textbf{73.7} & \textbf{64.5} \\ (uniform)\\\midrule

\rowcolor{lightgray}
\multicolumn{13}{l}{\emph{\ours{}-13B}}\\\midrule
\includegraphics[width=0.055\textwidth]{images/s0.png}
& 72.2 & \textbf{67.8} & 70.2 & 69.8 & 78.7 & 44.6 & 58.9 & 73.2 & \textbf{82.0} & 45.5 & 76.6 & 67.2 \\ (truncated)\\\midrule
\includegraphics[width=0.055\textwidth]{images/s1.png}
& 72.1 & 67.4 & 69.9 & 70.6 & 76.6 & 44.8 & 60.4 & 73.1 & 80.7 & 45.8 & 78.2 & 67.2 \\ (cosine)\\\midrule
\includegraphics[width=0.055\textwidth]{images/s5.png}
& \textbf{72.5} & 67.5 & \textbf{71.2} & 69.3 & 79.6 & \textbf{45.3} & 58.6 & 72.3 & 80.8 & 45.3 & 78.3 & 67.3 \\ (linear)\\\midrule
\includegraphics[width=0.055\textwidth]{images/unif.png}
& 72.2 & \textbf{67.8} & 70.8 & \textbf{71.4} & \textbf{83.4} & 45.0 & \textbf{61.4} & \textbf{73.9} & {81.1} & \textbf{47.0} & \textbf{79.3} & \textbf{68.5} \\ (uniform)\\
\bottomrule
    \end{tabular}
    \vspace{-0.5em}
    \caption{{Ablation study on train data samplers.}}
    \vspace{-1em}
    \label{tab:alblation_train_schedule}
\end{table}